\documentclass[a4paper,10pt]{article}

\usepackage[utf8]{inputenc}
\usepackage[backend=biber, style=numeric, sorting=none]{biblatex}
\usepackage{geometry}
\usepackage{amsmath}
\usepackage{mathptmx}
\usepackage{graphicx}
\usepackage{caption}
\usepackage{hyperref}
\usepackage{CJKutf8}
\usepackage{array}
\usepackage{float}
\usepackage{multirow} 
\usepackage{csquotes}
\usepackage{enumitem}
\usepackage{multicol}
\usepackage{pdfpages}
\usepackage{tabularray}
\usepackage{array}
\usepackage{rotating}
\usepackage{lscape}
\usepackage{setspace}
\usepackage{tabularray}
\usepackage[normalem]{ulem}
\usepackage{fancyhdr}
\usepackage{array, multirow} 
\UseTblrLibrary{booktabs}
\usepackage{ifxetex,ifluatex}
\usepackage{etoolbox}
\usepackage{tikz}
\usepackage{framed}
\usepackage{tabularray}
\usepackage{adjustbox}
\usepackage[normalem]{ulem}
\usepackage{makecell}
\usepackage{hyperref}
\newgeometry{left=0.45in,right=0.45in,top=0.45in,bottom=0.45in}
\title{A comprehensive GeoAI review: Progress, Challenges and Outlooks}
\author{Anasse Boutayeb \textsuperscript{*} \and Iyad Lahsen-cherif \textsuperscript{*} \and Ahmed El Khadimi \textsuperscript{*}}
\date{\today}
\fancypagestyle{firstpage}{ 
  \fancyhf{} 
  
  \fancyfoot[L]{\raisebox{12pt}{\textsuperscript{*} Institut National des Postes et Télécommunications (INPT)}} 
}
\usepackage{lipsum}
\usepackage{tikz}
\usetikzlibrary{backgrounds}
\makeatletter

\tikzset{%
  fancy quotes/.style={
    text width=\fq@width pt,
    align=justify, 
    inner sep=1em,
    inner xsep=2em, 
    anchor=north west,
    minimum width=\linewidth - 4em, 
  },
  fancy quotes width/.initial={.8\linewidth},
  fancy quotes marks/.style={
    scale=4, 
    text=black, 
    inner sep=0pt,
  },
  fancy quotes opening/.style={
    fancy quotes marks,
  },
  fancy quotes closing/.style={
    fancy quotes marks,
  },
  fancy quotes background/.style={
    show background rectangle,
    inner frame xsep=0pt,
    background rectangle/.style={
      fill=white, 
      rounded corners,
    },
  }
}
\newenvironment{fancyquotes}[1][]{%
  \noindent
  \begin{center} 
  \begin{tikzpicture}[fancy quotes background]
  \node[fancy quotes opening,anchor=north west] (fq@ul) at (0,0) {``};
  \tikz@scan@one@point\pgfutil@firstofone(fq@ul.east)
  \pgfmathsetmacro{\fq@width}{\linewidth - 2*\pgf@x - 4em} 
  \node[fancy quotes,#1] (fq@txt) at (fq@ul.north west) \bgroup}
  {\egroup;
  \node[overlay,fancy quotes closing,anchor=east] at (fq@txt.south east) {''};
  \end{tikzpicture}
  \end{center}
}
\makeatother
\begin{document}
\maketitle
\thispagestyle{firstpage}
\setlength{\columnsep}{12pt}
\noindent\textbf{Abstract-} In recent years, Geospatial Artificial Intelligence (GeoAI) has gained traction in the most relevant research works and industrial applications, while also becoming involved in various fields of use. This paper offers a comprehensive review of GeoAI as a synergistic concept applying Artificial Intelligence (AI) methods and models to geospatial data. A preliminary study is carried out, identifying the methodology of the work, the research motivations, the issues and the directions to be tracked, followed by exploring how GeoAI can be used in various interesting fields of application, such as precision agriculture, environmental monitoring, disaster management and urban planning. Next, a statistical and semantic analysis is carried out, followed by a clear and precise presentation of the challenges facing GeoAI. Then, a concrete exploration of the future prospects is provided, based on several informations gathered during the census. To sum up, this paper provides a complete overview of the correlation between AI and the geospatial domain, while mentioning the researches conducted in this context, and emphasizing the close relationship linking GeoAI with other advanced concepts such as geographic information systems (GIS) and large-scale geospatial data, known as big geodata. This will enable researchers and scientific community to assess the state of progress in this promising field, and will help other interested parties to gain a better understanding of the issues involved.
\\
\\
\\
\begin{multicols}{2}
\raggedcolumns
\section{Introduction}
Artificial intelligence (AI) has become one of the most influential technological advances in human history \cite{zhang2021study}, involving radical metamorphoses in numerous active sectors, such as healthcare, finance, commerce, transport, etc. AI relies on several algorithms and methods to emulate human cognition, enabling data analysis in any quantity or complexity, achieving advanced automation of the most complicated processes, and helping to make the most crucial decisions. It can also perform various tasks with unprecedented precision. For example, AI facilitates the exploitation of medical images while enabling the automatic and rapid detection of tumours; it also facilitates the analysis of the most complex genetic data \cite{jiang2017artificial}. On the other hand, AI algorithms optimise production and exploitation of renewable energies, predict consumer demand and the optimum location for a large surface area \cite{entezari2023artificial}. In addition, AI can make an effective contribution to improving teaching methods, by adapting the pace and content of courses to students' levels, while analysing consumer purchasing behaviour \cite{chen2020artificial}, improving sales techniques and boosting yields \cite{bawack2022artificial}. In short, AI is currently a promising vector for efficiency and development, guaranteeing exceptional performance.
\\
Furthermore, geospatial information has a particular impact on people's daily lives, making it easy to navigate in urban and rural areas, locating schools, parking lots, stores, hospitals and medical centers with great precision, and finding directions to these places very quickly. In addition, geospatial technologies are at the heart of cutting-edge fields such as agriculture \cite{mitran2021geospatial, raju2019geospatial}, environment \cite{rai2022geospatial, butenko2020geospatial}, defense \cite{sui2008geospatial}, urban planning \cite{shen2012geospatial, reddy2018geospatial, jiang2010geospatial} and healthcare \cite{bhunia2019geospatial, faruque2022geospatial}, while offering a new dimension of analysis and reflection, by combining information from various acquisition systems, specifically satellites, drones, GPS receivers, etc. This enables a high-level community of researchers and industrialists to address today's global challenges in every field, to participate in the era of digital transformation, and to ensure sustainability and global development.
\\
More specifically, AI is a key tool allowing the management and exploitation of geospatial data, paving the way for a new era of analysis and interpretation, particularly when it comes to massive and complex data, or when it comes to sophisticated tasks. As a result, AI can perform these tasks while guaranteeing high accuracy, including the processing of remote sensing images to produce land use maps \cite{alqadhi_applying_2024}, the segmentation of agricultural crops \cite{gallo_-season_2023}, the detection of cars on road networks \cite{moranduzzo2012sift}, the analysis of changes in spatio-temporal time series \cite{kalinicheva2020unsupervised}, and the use of precipitation to make weather forecasts \cite{sengoz_machine_2023}. Additionally, powerful AI models are designed to exploit temperature and vegetation informations in order to quantify the degree of drought over a vast geographical area, massive trajectory data can also be exploited to model high-performance route calculation algorithms. AI can also be used to predict the weather of floods by analysing remote sensing images and rainfall level measurements \cite{lammers2021prediction}. It should be noted that this multitude of applications of artificial intelligence in relation to geospatiality has given rise to a new concept, namely \textbf{GeoAI}, which is carefully explored in this paper, indicating the issues involved, the level of progress achieved and futurs trends.
\\
The remainder of this paper is organized as follows: section 2 presents the issues at stake in the subject, the Research Questions (RQs) to be answered and the research directions that have followed, section 3 introduces the theme of artificial intelligence, covering the main concepts and models, as well as new emerging approaches, section 4 develops the research directions defined above, leaving the final section to provide exhaustive answers to the research questions posed, and to introduce the related challenges and the future prospects, presented as results of the exploration carried out.
\section{Issues, key questions and research directions}
\subsection{History and definitions}
Looking at history of the synergy between AI and geospatial area, it is noted that this coupling began as early as the 80s and 90s, through the works of Smith \cite{smith1984artificial}, Estes et al. \cite{estes1986applications}, Couclelis \cite{couclelis1986artificial}, Openshaw and Openshaw \cite{openshaw1997artificial}. A major milestone in the history of GeoAI is the "Association for Computing Machinery (ACM) SIGSPATIAL conference" in 2017, where a specific workshop is dedicated to the subject, defining GeoAI explicitly as the coupling of Artificial Intelligence techniques to geographic information or geospatial data \cite{mao_geoai_2018, hu2019geoai}. Moreover, several definitions of GeoAI have been provided. According to Gao \cite{gao_geospatial_2021}, GeoAI can be regarded as :
\begin{fancyquotes}
a branch of AI whose objective is to implement intelligent computer programs aiming to imitate human perception, particularly with regard to spatial reasoning and geographical dynamics.
\end{fancyquotes}
Li and Hsu \cite{li_geoai_2022} also define GeoAI as :
\begin{fancyquotes}
a new field of interdisciplinary research that uses geospatial Big Data while leveraging and developing AI for location-based analysis.
\end{fancyquotes}
Furthermore, emphasizing the applicative aspect of the geospatial field, Kausica et al. \cite{kausika_geoai_2021} assert that : 
\begin{fancyquotes}
GeoAI combines AI algorithms with geospatial methods to extract the most significant information, with the purpose of performing sophisticated tasks in the same field.
\end{fancyquotes}
\begin{figure*}[ht]
    \centering
    \includegraphics[width=0.8\textwidth]{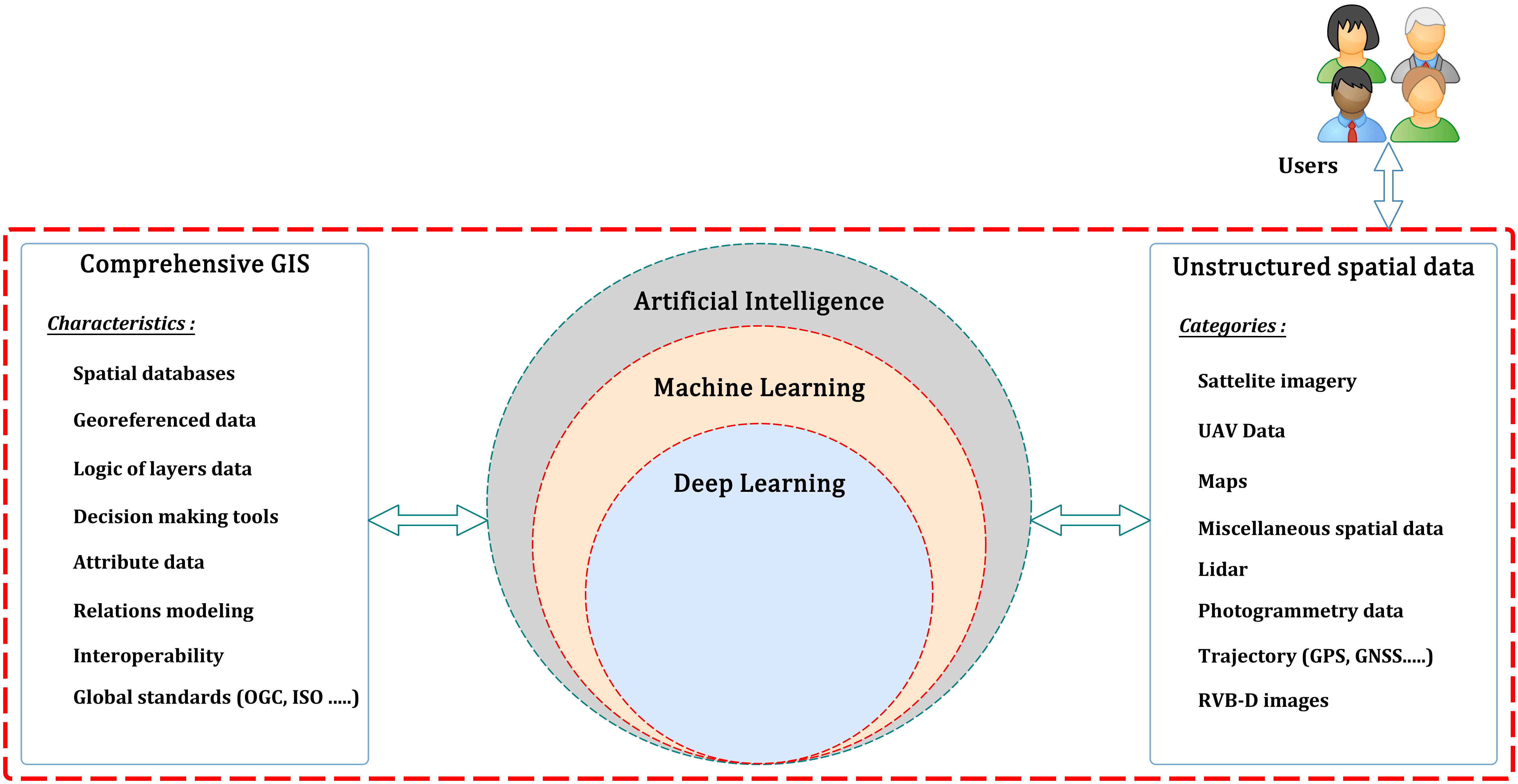}
    \captionsetup{font=scriptsize}
    \caption{GeoAI and GIS.}
    \label{fig:label-de-votre-figure2}
\end{figure*}
\subsection{Related topics, problematic and census approach}
As presented above, GeoAI highlights a fundamental aspect that will be exploited in this paper, namely the synergy between AI and its most advanced techniques, on the one hand, and geospatial data, on the other. At a deeper level, the importance of linking GeoAI to geographic information systems (GIS) should be noted, as this enables geospatial data to be catalogued for appropriate manipulation, analysis and presentation \cite{noauthor_understanding_2022}. Figure 1 illustrates the contribution of GIS as an ideal collaborator with AI, due to its efficiency and standardization.
\\
Another fundamental aspect to highlight is the use of big geodata, a concept referring to large-scale geospatial data, which emphasises the use of AI models to manage, process and use this type of data, taking into account its variety of sources, natures and scales \cite{zhou2019geoai}.  
\\
To address the subject of GeoAI comprehensively, it is essential to emphasize the importance of citing the various applications of GeoAI, including urban planning, logistics and transportation, agriculture, water resource management, and many others. In fact, the importance of listing GeoAI research works by application domains allows to follow a semantically consistent approach, i.e. each sub-domain to be developed is completely distinct, in terms of research objects, from the others. Furthermore, and to situate the aim of this work in relation to previous works, a great deal of research is carried out on this subject, as shown in Table 1, while touching on different sub-branches of geospatial field, but without giving a detailed overview, nor focusing on a wide range of application areas.
\\
The problem at hand, is to provide a \textbf{comprehensive} review of the GeoAI's state of the art in terms of application domains, differing from previous work both in terms of the completeness of the inventory and in terms of research objectives.
\\
To address this issue, a scoping and design phase is carried out, which consists of choosing the working methodology and clearly identifying the preliminary issues. Then, a definition of research questions is fundamental to orientate the census to be carried out. Next, a determination of research axes is worked out to identify key elements of the concerned investigation, using an advanced keyword search on Scopus database \cite{scopus2024} and Google Scholar engine \cite{google_scholar}, on the one hand, and refining the choice of fields of application to be discussed, on the other. As illustrated in Figure 2, the papers are selected on the basis of a set of criterias: 
\begin{itemize}[leftmargin=*]
    \item Innovative contribution: given that the final objective of this work is to explore the level of progress achieved.
    \item Diversification of selection themes: the aim is to cover all areas in question without duplicating the subjects tackled.
    \item Quality of papers: those mainly published by renowned publishers such as Elsevier \cite{sciencedirect}, the Institute of Electrical and Electronics Engineers (IEEE) \cite{ieeeXplore}, Springer \cite{springer} and Multidisciplinary Digital Publishing Institute (MDPI) \cite{mdpi_platform} are selected. 
\end{itemize}
\subsection{Research questions and directions}
A crucial step in this research is to first formulate the research questions in order to clarify the objectives to be achieved and to guide the work to be carried out. Three questions are important to ask:
\begin{itemize}[leftmargin=*]
    \item \textbf{RQ1:} How can GeoAI methods contribute to effective and accurate results in various fields of application? 
    \item \textbf{RQ2:} What are the challenges currently facing GeoAI?
    \item \textbf{RQ3:} What are the future prospects for GeoAI? More specifically, what are the promising themes to be explored in this context?    
\end{itemize}
Throughout the rest of this paper, relevant research focusing on the application of creative AI methods to geospatial domain are identified. To this end, the following research directions are chosen for exploration, according to their growing importance and their presence in the most relevant research works:

\begin{itemize}[leftmargin=*]
    \item Precision agriculture;
    \item Urban planning, logistics and transportation;
    \item Environmental management;
    \item Water resources management and precipitation forecasting; 
    \item Natural disaster monitoring;
    \item Healthcare.
\end{itemize}
The definition of these axes is intended to give an overview of the state of the art of GeoAI, not only in terms of the multitude of functionalities it offers, but also in terms of the diversity of its fields of application and its contribution to geospatial area, as well as the relevance of messages and services it provides.
\\
Before getting to the heart of the chosen research axes, it is needed to present the basic principles of artificial intelligence, its definitions, methods, approaches and evaluation metrics, which is conducted in the following subsections (from 3.1 to 3.4). Subsection 3.5 is devoted to the presentation of the commonly used datasets, afterwards, subsection 3.6 is dedicated to the hardwares used, its categories, architectures, specifications, etc. Section 4 then develops the above-mentioned research directions, with each subsection listing the various works presenting GeoAI models according to the relevant field of application, respecting the selection criteria indicated in 2.2 in order to provide a comprehensive answer to RQ1. Finally, section 5 is dedicated to the answers to RQ2 and RQ3, giving a detailed presentation of the challenges and the perspectives involved.
\section{Artificial Intelligence : definitions and models}
According to Xu et al. \cite{xu2021artificial} Artificial Intelligence (AI) corresponds to :
\begin{fancyquotes}
a discipline referring to the simulation of human intelligence by machines, with the aim of imitating human behavior in different situations.
\end{fancyquotes}
Moreover, Santosh et al. \cite{noauthor_deep_2022} define Machine Learning (ML) as :
\begin{fancyquotes}
a set of techniques for approximating a function that maps an input space to an output space, while extracting meaningful, non-redundant information from data samples.
\end{fancyquotes}
Furthermore, LeCun et al. \cite{lecun2015deep} presents Deep Learning (DL) as :
\begin{fancyquotes}
a branch of Machine Learning using multi-layer architectures to learn multiple representations of data, it is noted that deep learning uses the back-propagation technique to adjust the internal parameters of the model in order to recalculate these representations at each epoch.
\end{fancyquotes}
In this section, a comprehensive overview of all commonly used AI methods and models is given, along with the main concepts put forward in the context in question.
\subsection{Machine Learning (ML) models}
In order to carry out the various tasks commonly performed by AI, such as \textbf{regression} where the values of scalars are predicted, \textbf{classification} where a label is assigned to a specific piece of data, or \textbf{clustering} where a partitioning of data is proposed, a multitude of machine learning models are presented in the literature. In this subsection, different models are explored depending on the nature of the data used. In fact, two types of learning are distinguished: supervised learning, when the training data are labeled, and unsupervised learning, in the opposite case \cite{love2002comparing}.
\subsubsection{Supervised learning}
In this context, the data used for training is labeled, i.e. each data record is associated with a specific value. This value is then employed during the learning phases as a reference value, used to represent the links between inputs and outputs \cite{cunningham2008supervised}. In this subsection, the most notable regression and supervised classification models are explored.
\paragraph{\textbf{Linear Regression:}}is an approach for modelling the relationship between a scalar dependent variable and explanatory variables, the case of a single explanatory variable being called simple linear regression \cite{uyanik2013study}, the relative equation is as follows:
\begin{equation}
   \hat{y} = \beta_0 + \beta_1 x_1 + \beta_2 x_2 + \dots + \beta_n x_n + \epsilon,
\end{equation}
\end{multicols}
\newpage
\footnotesize
\newgeometry{margin=0.54in}
\begin{landscape}
\begin{longtblr}[
  label = {Previous research on GeoAI and GIS-AI synergy},
  entry = {Previous research on GeoAI and GIS-AI synergy},
  caption = {Previous research on GeoAI and GIS-AI synergy}
]{
  width = \linewidth,
  colspec = {Q[c,50]Q[210]Q[194]Q[433]}, 
  hlines,
  vlines,
  cell{1}{1} = {c},
  cell{1}{2} = {c},
  cell{1}{3} = {c},
  cell{1}{4} = {c},
}
\textbf{References}                                         & \textbf{Paper title}                                                                                                & \textbf{Journal/Conference}                                                                            & \textbf{Object}                                                                                                                                                                                                                                     \\
\cite{hu_geoai_2019}                           & GeoAI at ACM SIGSPATIAL: progress, challenges, and future directions                                                & SIGSPATIAL Special (2019)                                                                              & Explore the workshops from the 2017 SIGSPATIAL conference while covering several application areas, including transport and public health.                                                                                                              \\
\cite{hu_five-year_2024}                       & A five-year milestone: reflections on advances and limitations in GeoAI research                                    & Annals of GIS (2024)                                                                                   & Determining the level of progress and constraints of the GeoAI five years on after the first American Association of Geographers (AAG) symposium.                                                                                                          \\
\cite{p_s_chauhan_geoai_2021}       & GeoAI – Accelerating a Virtuous Cycle between AI and Geo                                                            & IC3-2021: Proceedings of the 2021 Thirteenth International Conference on Contemporary Computing (2021) & List the possibilities offered by AI in the geospatial field, identifying the areas of application of GeoAI, such as geocoding and change detection.                                                                                        \\
\cite{pierdicca_geoai_2022}       & GeoAI: a review of artificial intelligence approaches for the interpretation of complex geomatics data              & Geoscientific Instrumentation, Methods and Data Systems (2022)                                         & Use the Preferred Reporting Items for Systematic Reviews and Meta-Analyses (PRISMA) standard to systematically explore GeoAI techniques according to the type of geospatial data used, while developing a statistical analysis.                                                                          \\
\cite{mai_opportunities_2024}                 & On the Opportunities and Challenges of Foundation Models for GeoAI (Vision Paper)                                   & ACM Transactions on Spatial Algorithms and Systems (2024)                                              & Explore the prospects of possible GeoAI foundation models for different categories of geospatial data.                                                                                                                          \\
\cite{sieber2016geoai}                            & GeoAI and its implications                                                                                          & International Encyclopedia of Geography: People, the Earth, Environment and Technology (2016)          & Determine the degree of progress of GeoAI in many fields of application and explore its integration into the conceptualisation of geospatial methods.                                                                                         \\
\cite{li2020geoai}                               & GeoAI: Where machine learning and big data converge in GIScience                                                    & Journal of Spatial Information Science (2020)                                                          & Exploring the contribution of GeoAI to the processing and analysis of big geodata, while identifying its challenges and future promises.                                                                                                             \\
\cite{chen_exemplification_2023}                   & Exemplification on Potential Applications and Scenarios for GeoAI                                                   & 2023 Asia-Europe Conference on Electronics, Data Processing and Informatics (2023)                     & Determine the current advances in GeoAI research, while listing its latest practices in relation to geospatial methods.                                                                                                  \\
\cite{janowicz2020geoai}                 & GeoAI: spatially explicit artificial intelligence techniques for geographic knowledge discovery and beyond          & International Journal of Geographical Information Science (2020)                                       & List the various applications of AI in geospatial data processing, as well as the issues involved.                                                                                                   \\
\cite{gao2020review}                                & A review of recent researches and reflections on geospatial artificial intelligence                                 & Geomatics and Information Science of Wuhan University (2020)                                           & Present the relationship between GeoAI and the advancement of geospatial research, exploring challenges and outlooks.                                                                                                                        \\
\cite{gao2023special}                         & Special issue on geospatial artificial intelligence                                                                 & Geoinformatica (2023)                                                                                  & Explore all the papers presented at the 4th edition of ACM SIGSPATIAL related to the subject of GeoAI.                                                                                                                                                            \\
\cite{srivastava2023applications} & Applications of Artificial Intelligence and Machine Learning in Geospatial Data                                     & Emerging Trends, Techniques, and Applications in Geospatial Data Science (2023)                        & Discover the advances in the application of machine learning (ML) techniques, by presenting the contribution of GeoAI to the extraction and generation of geospatial information.                                                     \\
\cite{li2021geoai}                                   & GeoAI and deep learning                                                                                             & The International Encyclopedia of Geography (2021)                                                     & Present very quickly the close relationship between deep learning and the geospatial domain.                                                                                                                                                               \\
\cite{janowicz2023philosophical}               & Philosophical Foundations of GeoAI: Exploring Sustainability, Diversity, and Bias in GeoAI and Spatial Data Science & Handbook of Geospatial Artificial Intelligence (2023)                                                  & Discover the philosophical foundations of the GeoAI while highlighting fundamental issues such as the lack of ethical neutrality of the GeoAI and its sustainability.                                                                     \\
\cite{bordogna2022artificial}       & Artificial Intelligence for Multisource Geospatial Information                                                      & ISPRS International Journal of Geo-Information (2022)                                                  & Exploring the application of various artificial intelligence (AI) methods to geographic data from a multitude of sources.                                                                                    \\
\cite{liu2023geospatial}                      & Geospatial AI in Earth Observation, Remote Sensing, and GIScience                                                   & Applied Sciences (2023)                                                                                & Determine the progress of GeoAI in relation to GIS and remote sensing.                                                                            \\
\cite{mohan_brief_2022}              & A Brief Review of Recent Developments in the Integration of Deep Learning with GIS                                  & Geomatics and Environmental Engineering (2022)                                                         & Provide a partial overview summarising the synergy between Deep Learning (DL) and GIS, while discovering its impact on several fields of application such as hydrology and natural disasters. \\
\cite{ekeanyanwu_merging_2022}         & Merging GIS and Machine Learning Techniques: A Paper Review                                                         & Journal of Geoscience and Environment Protection (2022)                                                & Discover several research projects combining Machine Learning (ML) methods and GIS, as well as the applications of this synergy, such as health and erosion modelling.                            \\
\end{longtblr}

\end{landscape}
\restoregeometry
\normalsize
\begin{multicols}{2}
\raggedcolumns

\begin{figure}[H]
  \centering
  \includegraphics[width=0.8\linewidth]{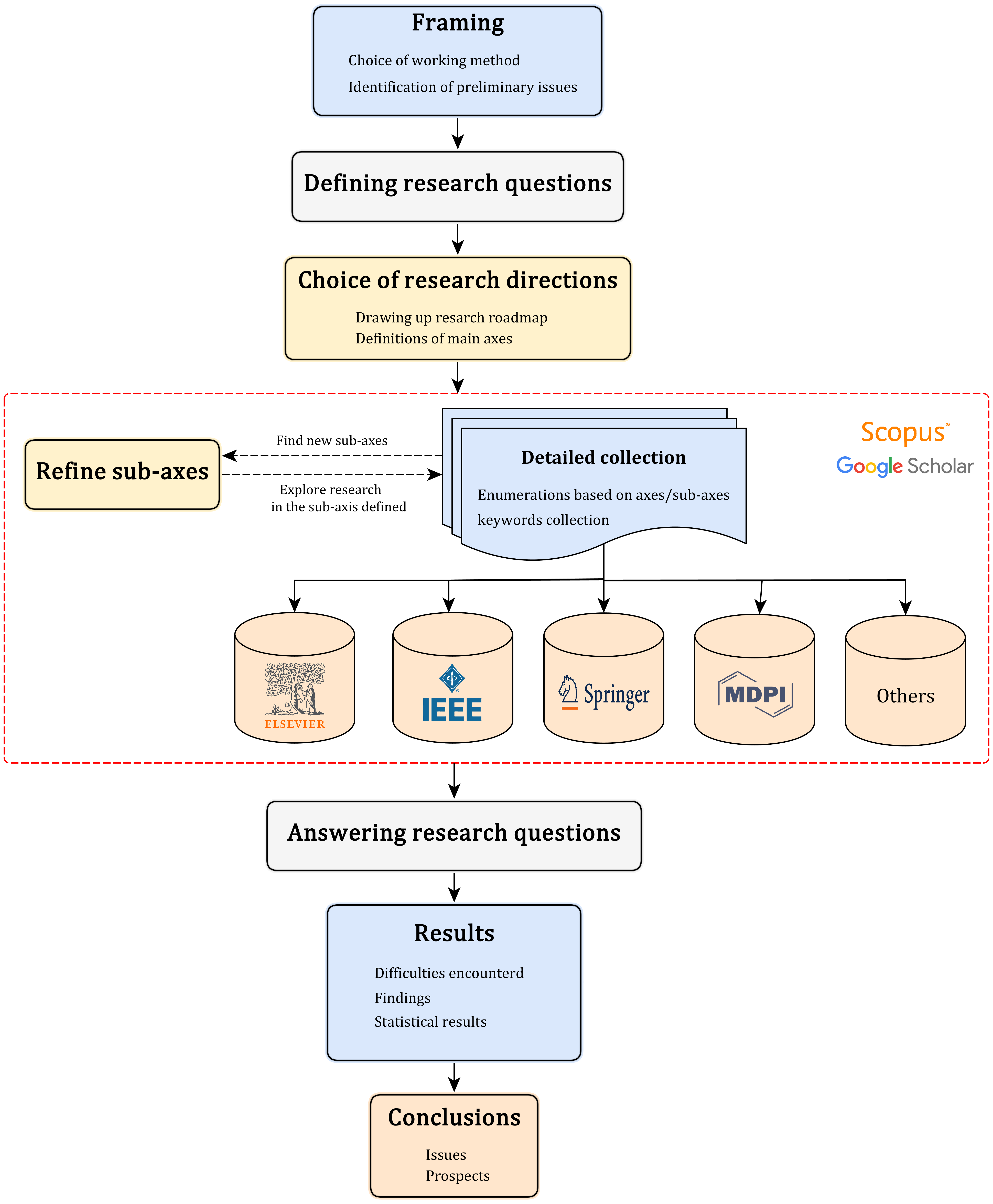} 
  \captionsetup{font=scriptsize}
  \caption{Review workflow.}
  \label{fig:label-de-votre-figure1}
\end{figure}

where: 
\begin{itemize}

\item y is the variable to be predicted,
\item $\hat{y}$ is the value of y predicted by the model, 
\item $x_1$, $x_2$ \dots $x_n$ are the independent variables used,
\item $\beta_0$, $\beta_1$ \dots $\beta_n$ are the coefficients of this regression,
\item $\epsilon$ is the error committed by the regression.
\end{itemize}
\noindent
The aim of this model is to learn the regression coefficients \(\beta_0, \beta_1, \ldots, \beta_n\) by minimizing the value $\sum_{i=1}^{m} (y_i - \hat{y_i})^2$, where m is the number of data points.
\\
In the case of a simple Linear Regression, Figure 3 shows the line that best fits a data point, $\beta_0$ is the y-intercept and $\beta_1$ is the slope of the blue line.

\begin{figure}[H]
  \centering
  \includegraphics[width=0.8\linewidth]{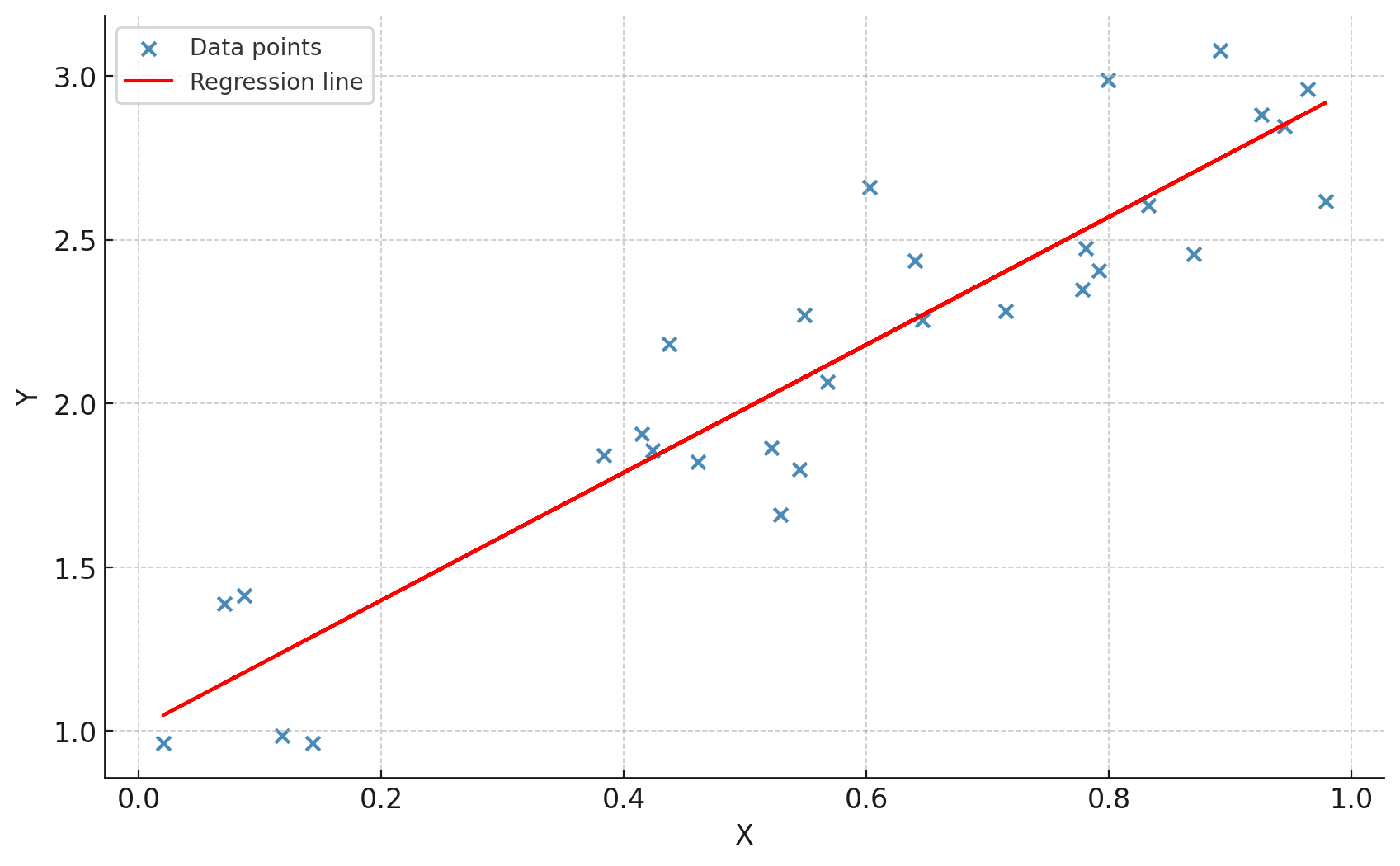}
  \captionsetup{font=scriptsize}
  \caption{Example of linear regression.}
  \label{fig:label-de-votre-figure1}
\end{figure}
\paragraph{\textbf{Logistic Regression:}}proposed by Cox \cite{cox1958regression}, logistic regression calculates the probability of a variable belonging to a given class by optimizing the involved coefficients. The associated probability function is a sigmoid function:

\begin{equation}
    P(y = 1 | x) = \frac{1}{1 + e^{-(\beta_0 + \beta_1 x_1 + \beta_2 x_2 + \dots + \beta_n x_n)}},
\end{equation}
\\
where: 
\begin{itemize}
\item y is the binary variable to be predicted (class membership),
\item $x_1$, $x_2$ \dots $x_n$ are the independent variables used,
\item $\beta_0$, $\beta_1$ \dots $\beta_1$ are the coefficients to be adjusted.
\end{itemize}

\paragraph{\textbf{Decision trees:}}is a model used for regression and classification tasks, the principle behind which is to divide the training data into sub-batches for each node of the tree, with a set of decision criteria defining the depth of the tree until the final predictions are reached \cite{quinlan1986induction}.

\paragraph{\textbf{Random Forest (RF):}}is a supervised model combining several decision trees to perform predictive tasks such as classification, regression, etc. Introduced by Breiman \cite{breiman2001random}, the principle of Random Forest consists in training each tree from a batch of data by random selection, the final result is then obtained from an average of results from all the trees, Figure 4 presents the principle of this model.

\begin{figure}[H]
  \centering
  \includegraphics[width=0.9\linewidth]{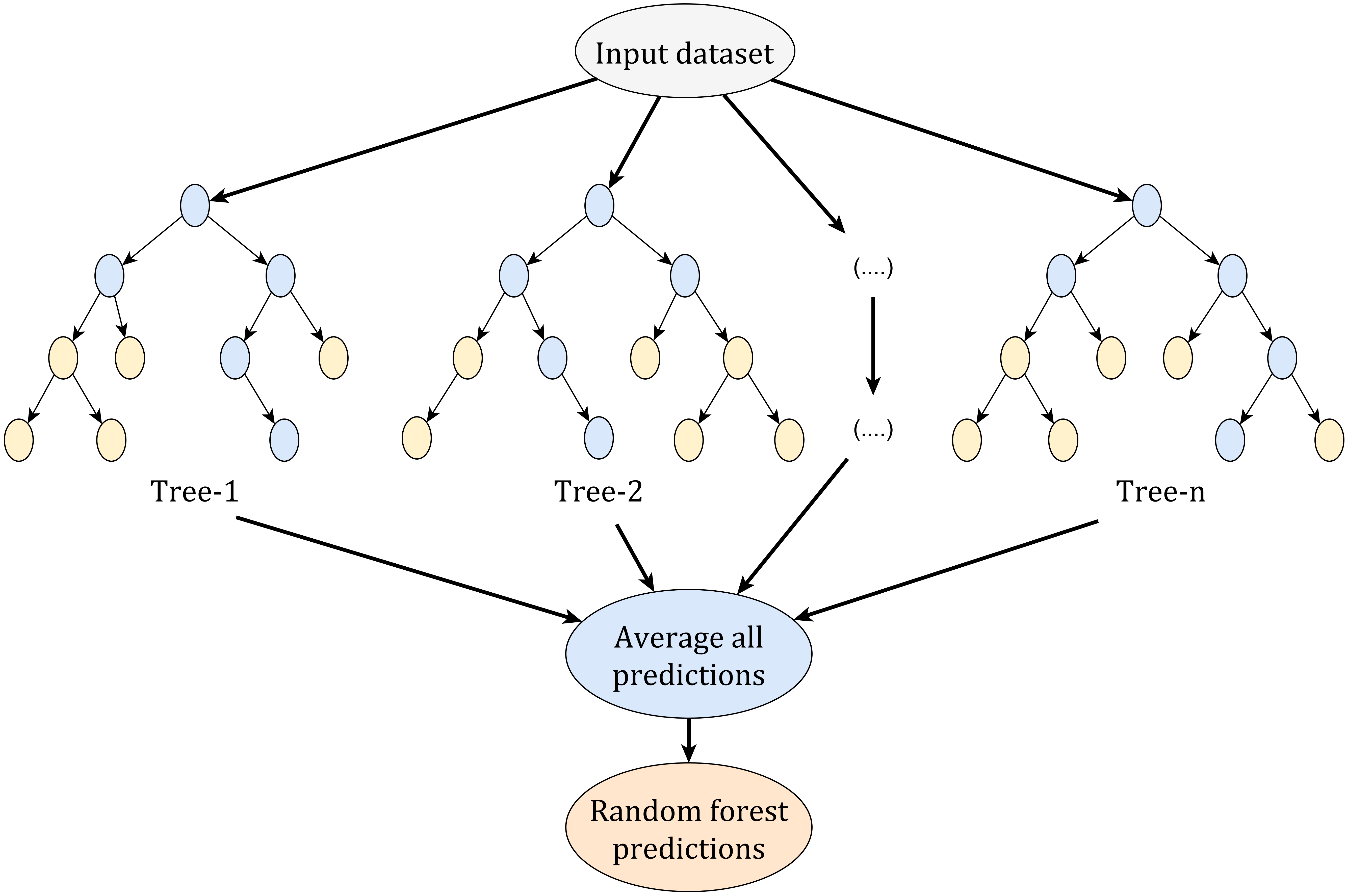}
  \captionsetup{font=scriptsize}
  \caption{Random Forest illustration (\cite{rf_ref}).}
  \label{fig:label-de-votre-figure1}
\end{figure}

\paragraph{\textbf{Support Vector Machine (SVM):}} proposed by Cortes and Vapnik \cite{cortes1995support}, it is a model used for classification. The mathematical principle is to search for the hyperplane (n-dimensional generalization of a plane) that best maximizes the distance between the hyperplane and the various data classes. 
\\
Figure 5 illustrates SVM model for a binary classification problem, the optimal hyperplane is constructed to separate the positive classes from the negative ones as best as possible, while maximizing the margin between its classes, two supporting hyperplanes (dotted lines) define the positive and negative boundaries.
\\
For the multi-class problem, several algorithms are used to transform it into several binary classification problems, such as One-vs-Rest and One-vs-One \cite{mayoraz1999support}.

\begin{figure}[H]
  \centering
  \includegraphics[width=0.9\linewidth]{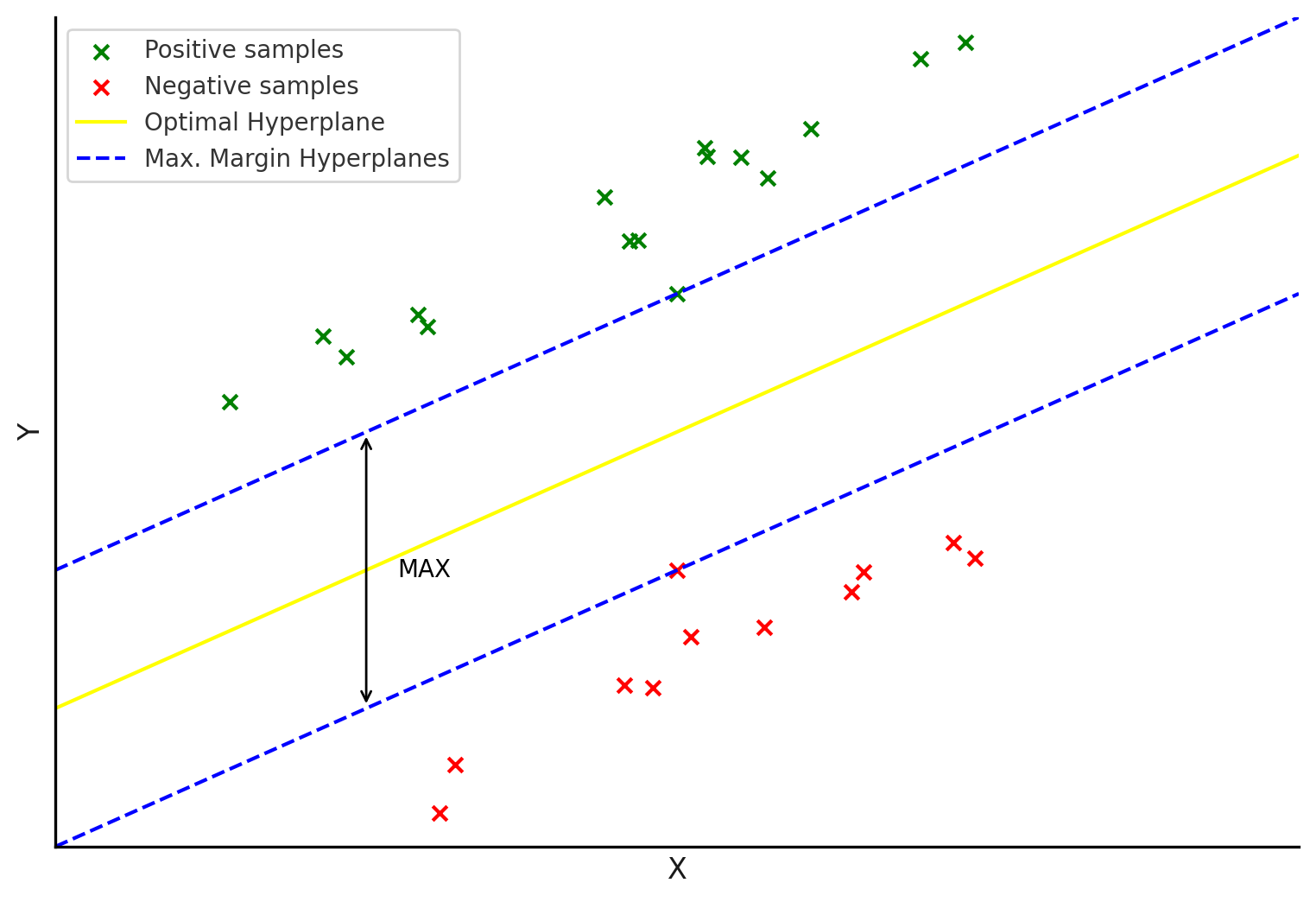}
  \captionsetup{font=scriptsize}
  \caption{SVM (\cite{svm_ref}).}
  \label{fig:label-de-votre-figure1}
\end{figure}
\paragraph{\textbf{K-Nearest Neighbors (KNN):}}is a model used for classification and regression, as shown in Figure 6, KNN consists in identifying, for a given measure, the K closest points in terms of distance \cite{fix1951discriminatory}. The value assigned to the candidate point is determined by the most frequent label of the K points, in the case of classification, and by averaging the K neighboring values, in the case of regression.
\begin{figure}[H]
  \centering
  \includegraphics[width=0.9\linewidth]{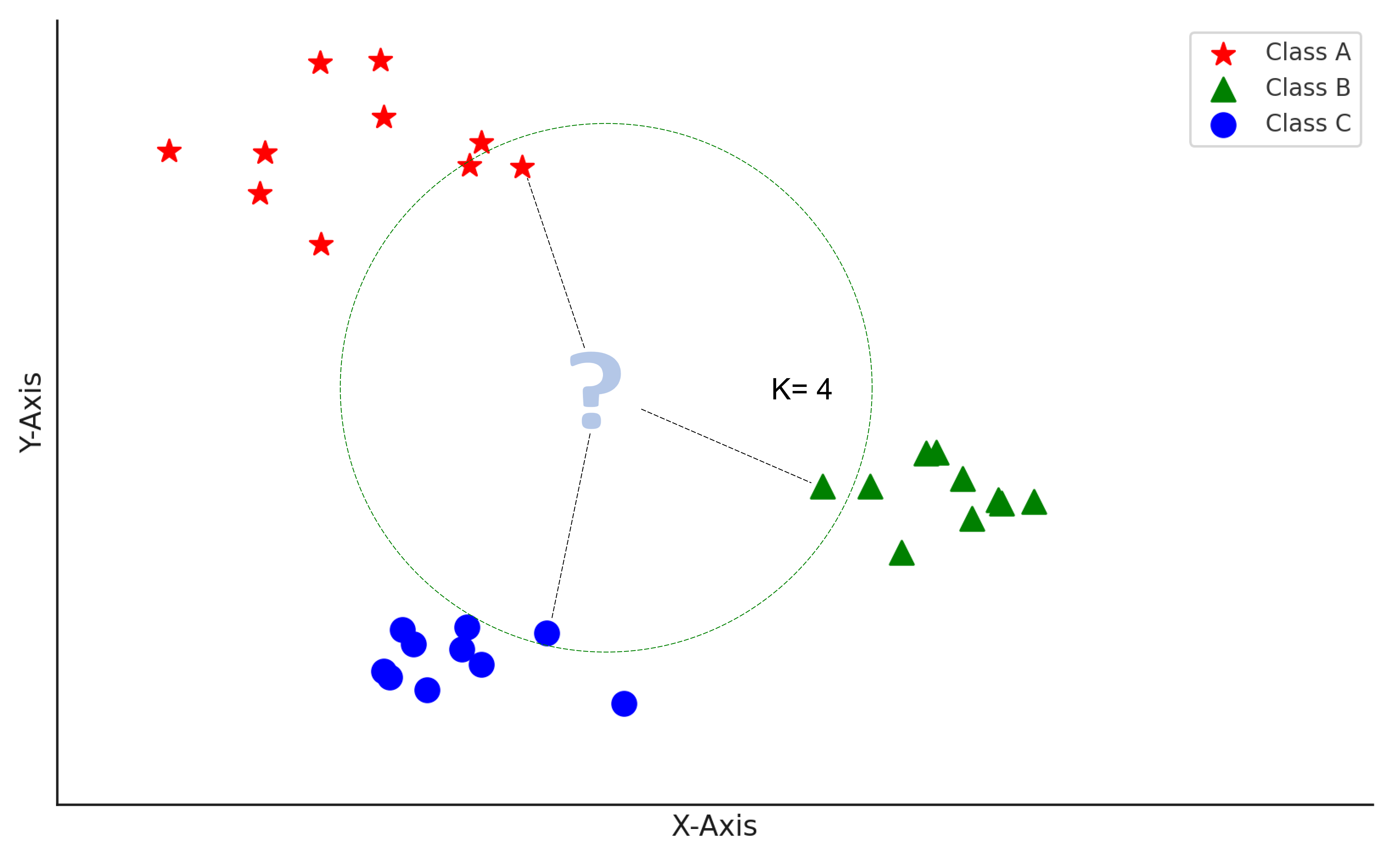}
  \captionsetup{font=scriptsize}
  \caption{KNN mechanism.}
  \label{fig:label-de-votre-figure1}
\end{figure}
\paragraph{\textbf{Gradient boosting:}}is a supervised technique used for classification and regression, which involves combining the prediction results of several decision trees. It is noted that Gradient boosting refines the prediction results of subsequent trees on the basis of the previous ones.
\\
Gradient boosting uses \textbf{ensemble learning}, the principle of which is to use several ML algorithms to obtain, via several voting mechanisms, results that are better than those obtained by each of the methods alone. Several methods are used in ensemble learning, such as Bagging \cite{breiman1996bagging}, Boosting \cite{freund1997decision} and Stacking \cite{wolpert1992stacked}.
\\
More specifically, Extreme Gradient Boosting (XGBoost) is an optimized implementation of the Gradient Boosting algorithm. Introduced by Chen and Guestrin \cite{chen2016xgboost}, XGBoost stands out for its computational efficiency and resource management.
\subsubsection{Unsupervised models}
As stated by Barlow and Horace \cite{barlow1989unsupervised}, unsupervised learning algorithms allows to learn the characteristics and representations of data without having any information about data labels. This seems very practical, given that, in many cases, the data are not annotated. In the following, unsupervised ML models are listed.
\paragraph{\textbf{K-means}} is a clustering method grouping a set of data points into K groups. Each measurement is taken and assigned to the nearest group, by measuring the distance between the point in question and the centroid of each group, then, this centroid is recalculated taking into account the new point added. This operation is repeated until the shape of groups becomes immutable \cite{macqueen1967some}. Mathematically, groups are formed by minimizing the sum of the squared distances between the measurements and the centroids of the groups, i.e: 
\begin{equation}
\sum_{i=1}^{K} \sum_{x \in S_i} \| x - \mu_i \|_{2}^{2},
\end{equation}
\noindent
\\
where: 
\begin{itemize}
\item x is the measurement in question,
\item $\mu_i$ is the centroid of the cluster,
\item $\|.\|_2$ is the L2 norm.
\end{itemize}
The K value is a key parameter in K-means algorithm, as it refers to the number of clusters, and therefore the clustering structure implemented, thus, it directly affects the model performance in terms of prediction quality and convergence.
\paragraph{\textbf{Density-Based Spatial Clustering of Applications with Noise (DBSCAN)}}is a clustering model proposed by Ester et al. \cite{ester1996density}, enabling a set of data points to be grouped together on the basis of their density. The principle of this algorithm consists in using a central point while defining a neighborhood radius, and a minimum number of points to be grouped together in each cluster. The strength of DBSCAN lies in the fact that it enables measurement points of various shapes and sizes to be grouped together, and does not require the number of resulting groupings to be specified.
\subsection{Deep Learning models}
These are model architectures mimicking the functioning of the human brain. Deep Learning (DL) uses Deep Neural Networks (DNN) as a basic approach, highlighting two key concepts: gradient descent to calculate the weights of each neuron, and back-propagation to calculate the gradients of the cost function by involving the propagation of error backwards \cite{rumelhart1986learning}. In this section, the main Deep Learning models are listed.
\subsubsection{Multi Layer Perceptron (MLP)}
Proposed by Rosenblatt \cite{rosenblatt1958perceptron}, MLP represents the most elementary Deep Neural Network architecture. The first layer transmits input data representation to the hidden layers, where each neuron is connected to all neurons in the next layer. This transmission of information between layers is based on non-linear activation functions applied to each neuron, enabling the model to better understand relevant dependencies between data. Other probabilistic activation functions are applied to output layer in the case of classification, or a linear activation function in the case of regression, in order to achieve meaningful results \cite{popescu2009multilayer}.
\\
In spite of its simplicity, MLP is used in particular for tasks such as games, text and voice recognition.

\subsubsection{Convolutional Neural Networks (CNN)}
Convolutional Neural Network is a specific Deep Neural Network (DNN). As shown in Figure 7, it is broken down into the following elements, convolutional layers for Feature Extraction (FE), pooling layers for dimension reduction and Fully Connected layers (FC) for result classification. It is noted that several researchers have introduced the concept of CNN, such as Zhang et al. \cite{zhang1988shift} and LeCun et al. \cite{lecun1989backpropagation}, but the first work to implement a base of CNN is Waibel et al, who in 1987 proposed a Time-Delay Neural Network (TDNN) \cite{waibel1989phoneme}, which can be considered as a one-dimensional convolutional neural network. 
\begin{figure*}[ht]
    \centering
    \includegraphics[width=\textwidth]{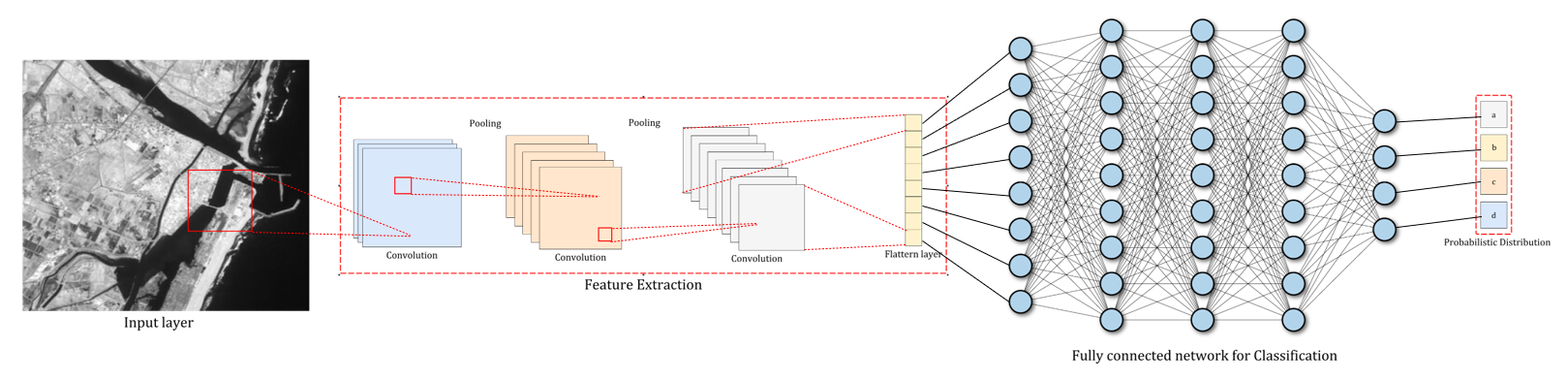}
    \captionsetup{font=scriptsize}
    \caption{CNN architecture.}
    \label{fig:label-de-votre-figure2}
\end{figure*}
\noindent CNN have demonstrated notable efficiency for several computer vision tasks, including classification, segmentation and object detection. In this context, a multitude of models have been developed, such as VGG \cite{simonyan2014very}, U-Net \cite{ronneberger2015u}, ResNet \cite{he2016deep} and YOLO \cite{redmon2016you}.
\subsubsection{Recurrent Neural Networks (RNN)}
RNN is a class of deep neural networks designed specifically to process sequential data including text, voice, time series, etc. Introduced in numerous works such as those by Hopefield \cite{hopfield1982neural} and Elman \cite{elman1990finding}, The special feature of the RNN is the recursiveness of its connections, enabling this type of network to remember previous entries.
\\
Despite the flexibility and adaptability offered by RNN, this architecture faces several problems, particularly when processing long sequences, specifically the problems of vanishing and exploding gradient \cite{hochreiter1998vanishing}, as well as the problem of retaining long-term dependencies \cite{bengio1994learning}, and the drain on computational resources due to long-term training \cite{appleyard2016optimizing}. These problems are overcome using the LSTM model \cite{hochreiter1997long}, a special RNN architecture made up of several blocks. Each block deals with a given stage of the sequence using four key elements:
\begin{itemize}
\item \textbf{Cell state}, transmitting dependencies throughout the sequence. 
\item \textbf{Forget state}, this component determines the information to be “forgotten” from cell state.
\item \textbf{Input gate}, deciding which information to add to cell state.
\item \textbf{Output gate}, sdetermining the information to be used for the final output.
\end{itemize}

\subsubsection{Auto-Encoders (AE)}
It is a particular neural network for unsupervised generative tasks. Presented by Hinton et al. \cite{hinton2006reducing}, the principle is to pass the data representation into a latent space in order to reduce dimensuality, and then reconstruct the data by transposed convolution. As illustrated in Figure 8, the encoder compresses the data while extracting the essential features captured, whereas the decoder aims to reconstruct the input structure through these features. 
\\
Auto-encoders are used for several tasks such as image compression \cite{jia2019layered}, anomaly detection \cite{finke2021autoencoders} and denoising \cite{ashfahani2020devdan}.
\begin{figure}[H]
  \centering
  \includegraphics[width=0.8\linewidth]{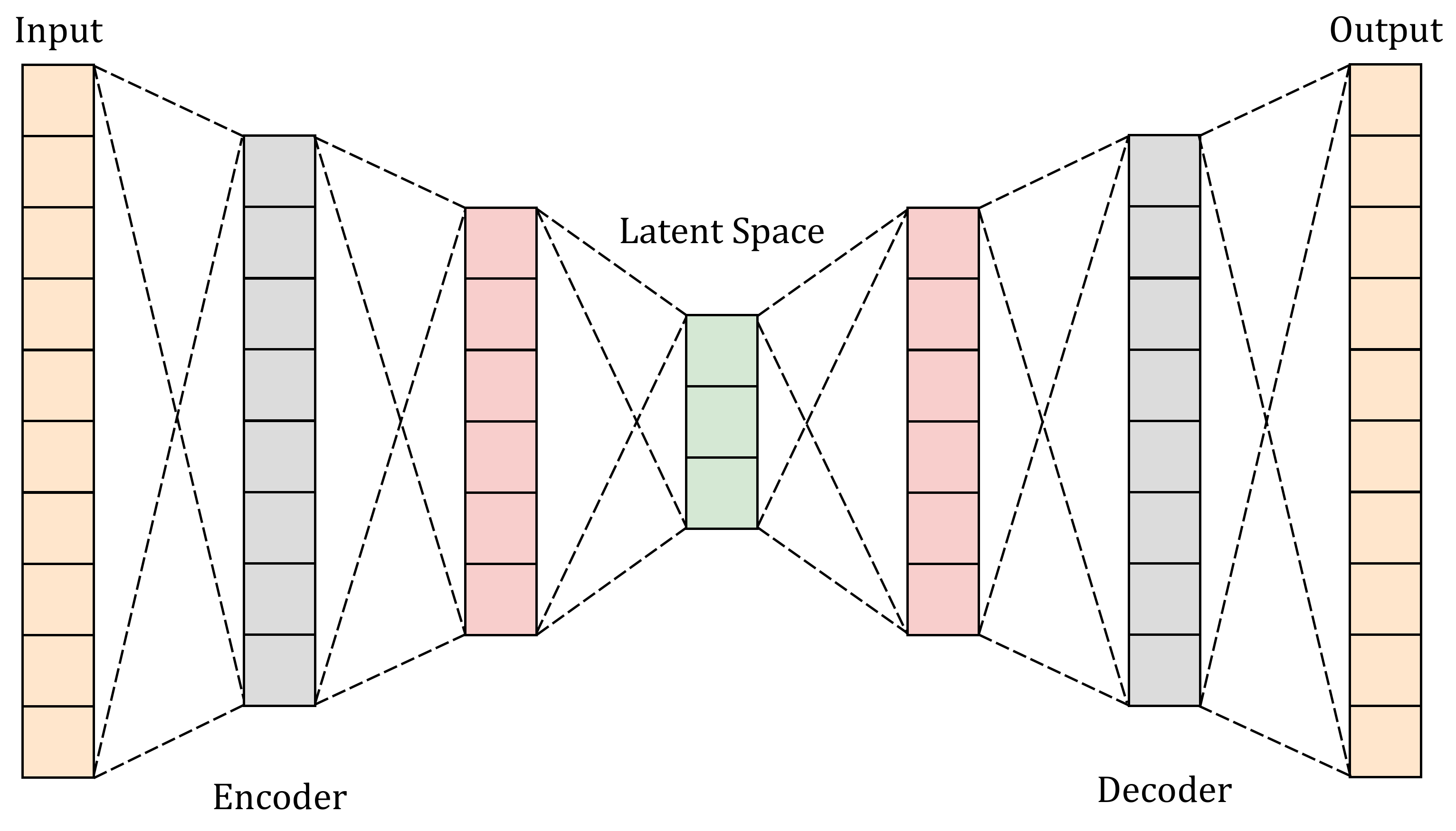}
  \captionsetup{font=scriptsize}
  \caption{AE architecture.}
  \label{fig:label-de-votre-figure1}
\end{figure}
\subsubsection{Generative Adversarial Networks (GAN)}
As their name suggests, Generative Adversarial Networks are used to generate data that resembles real ones as closely as possible. Proposed by Goodfellow et al. \cite{goodfellow2014generative}, a GAN consists of two networks, a generator creating data from a latent space, often a Gaussian or uniform distribution \cite{grinstead1997introduction}, and a descriminator that distinguishes between "false" and "real" data, enabling the generator to improve accuracy of the data it produces. Figure 9 describes this process.
\\
Several applications use the basic GAN architecture, such as art \cite{oh2022pose}, image enhancement \cite{ledig2017photo, wang2018esrgan}, Video synthesis \cite{vondrick2016generating} and image translation \cite{ma2018gan}.
\begin{figure}[H]
  \centering
  \includegraphics[width=1\linewidth]{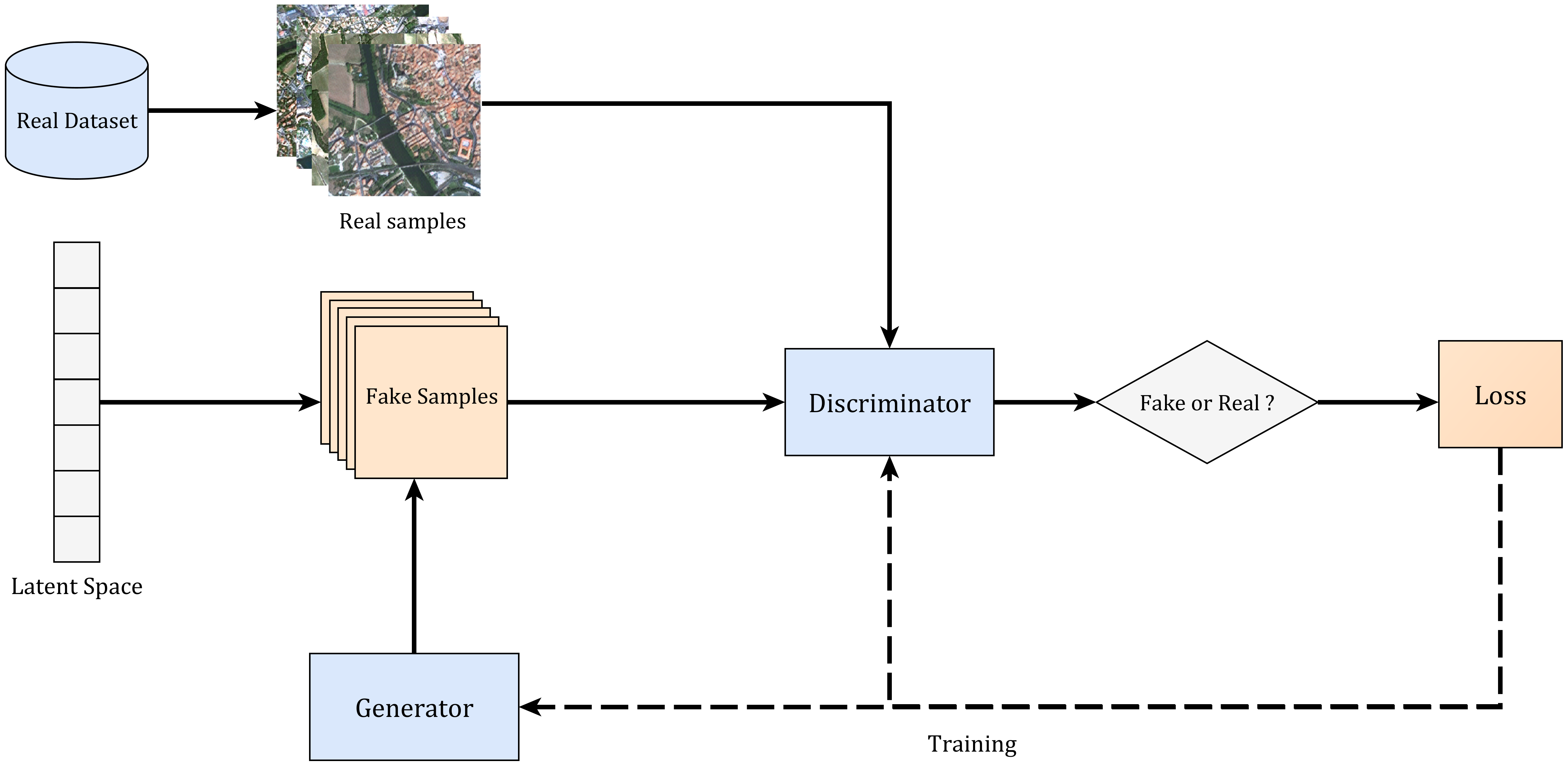}
  \captionsetup{font=scriptsize}
  \caption{GAN workflow.}
  \label{fig:label-de-votre-figure1}
\end{figure}

\subsection{Advanced notions of Artificial Intelligence}
In addition to Artificial Intelligence architectures, it is worth mentioning a number of advanced concepts linked to this subject. These concepts refer to mainly “theoretical” approaches, allowing traditional algorithms and methods to make unprecedented advances. Indeed, reinforcement, federated, ensemble, multimodal and transfer learning are presented, along with attention mechanisms.
\subsubsection{Reinforcement learning}
Reinforcement learning is an ML technique in which \textbf{the agent} improves iteratively its behavior on the basis of rewards or penalties considered as the results of its \textbf{actions}, representing the agent's interaction with a given environment, This \textbf{environment}, modeled by a set of \textbf{states}, stands for the external setting to which the agent tries to evolve its future state. To do this, the agent tries to improve its decision making during the learning process\cite{sutton1988learning}.
\\
Numerous variants of reinforcement learning are proposed, such as State-Action-Reward-State-Action (SARSA) \cite{andrew1998reinforcement}, Q-Learning \cite{watkins1992q} and  Proximal Policy optimization \cite{schulman2017proximal}.
\subsubsection{Federated learning}
Introduced by Mcmahan et al. \cite{mcmahan2017communication}, this innovative approach enables multiple entities (physical or logical), to contribute to the training of a model without exchanging directly users data. Each entity produces its own updated widgets based on this training, a central server combines then these updates to improve the model under consideration. Figure 10 summarizes this process in three steps.
\begin{figure}[H]
  \centering
  \includegraphics[width=0.8\linewidth]{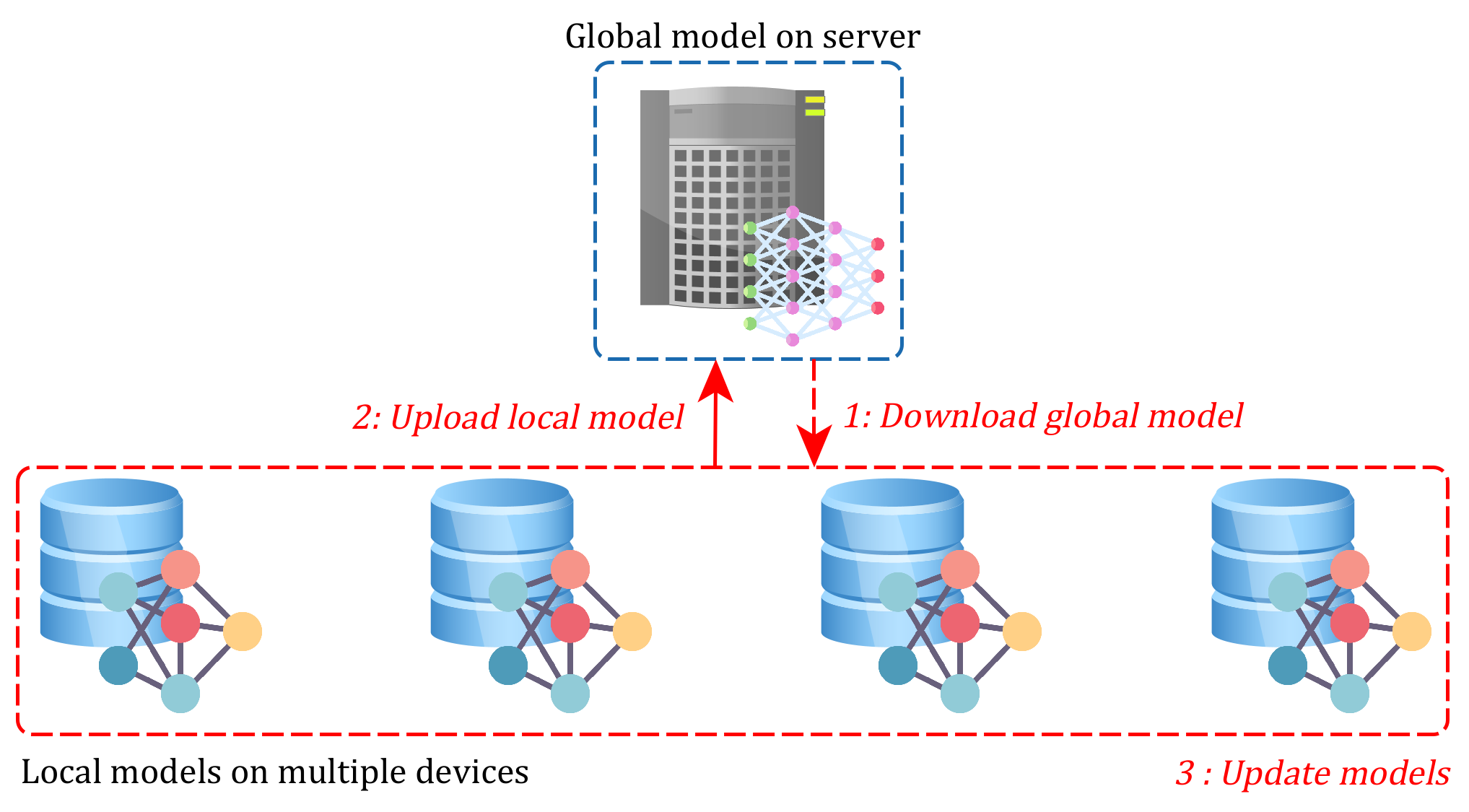}
  \captionsetup{font=scriptsize}
  \caption{Federated learning workflow.}
  \label{fig:label-de-votre-figure1}
\end{figure}
\noindent
One should note that federated learning architecture offers a multitude of remarkable attributes, including reduced communication costs, task scalability, security, confidentiality and data decentralization, but faces a multitude of problems such as data imbalance and system failures \cite{kairouz2021advances}.
\subsubsection{Multimodal learning}
According to Ngiam et al. \cite{ngiam2011multimodal}, multimodal learning involves the linking of information from multiple sources, for example, several categories of data including visual data, sound and text can be used to perform a specific task, such as recognising the exact representations of words or analysing feelings \cite{majumder2018multimodal}.
\\
Several approaches fit into the multimodal learning, including the \textbf{early fusion}, where multimodality is used from the start of model training, by combining features extracted from different sources of trained data, and the \textbf{late fusion}, where each modality is trained separately, the results of each prediction are then weighted towards the end of the process \cite{boulahia2021early}, as well as the \textbf{shared representation learning}, where all the modalities are projected into a single latent space. The aim of the latter approach is to capture the relationships between these different modalities in order to align and exploit them optimally \cite{borsa2016learning}.
\subsubsection{Transfer learning}
To overcome the problem of lack of data for a specific task, transfer learning exploits data relating to a source task or obtained in domains related to the main task \cite{pan2009survey, panait2005cooperative}. In practical terms, this involves adjusting the weights of an initial model, often trained on a large scale dataset, to a much smaller one.
\\
According to Weiss et al. \cite{weiss2016survey}, transfer learning improves learning while improving performance with respect to the configuration offered, by making better use of the training data, and by enormously reducing the cost of calculating weights. Despite this, transfer learning faces several challenges, in particular the negative transfer, referring to the case where the target domain is too different from the source one, as well as the non-availability of labelled data, whether for the source or target domain, a problem that is proving difficult to overcome.
\subsubsection{Attention mechanisms}
It is an advanced technique allowing a model to concentrate on the essential parts of data. The contribution of this concept is to be able to focus on relevant data by imitating the way humans pay attention to the interesting elements around them \cite{bahdanau2014neural}. Figure 11 describes an example of attention mechanism with respect to the constituents of a sequence, for instance, in this schema, remotely sensed image patches. Vector embedding is performed before calculating attention scores for each patch, followed by normalization using a probabilistic function for gradient stabilization and scaling, then,  weighted vector calculation and linear aggregation are implemented to combine the relevant extracted representations.
\begin{figure}[H]
  \centering
  \includegraphics[width=0.8\linewidth]{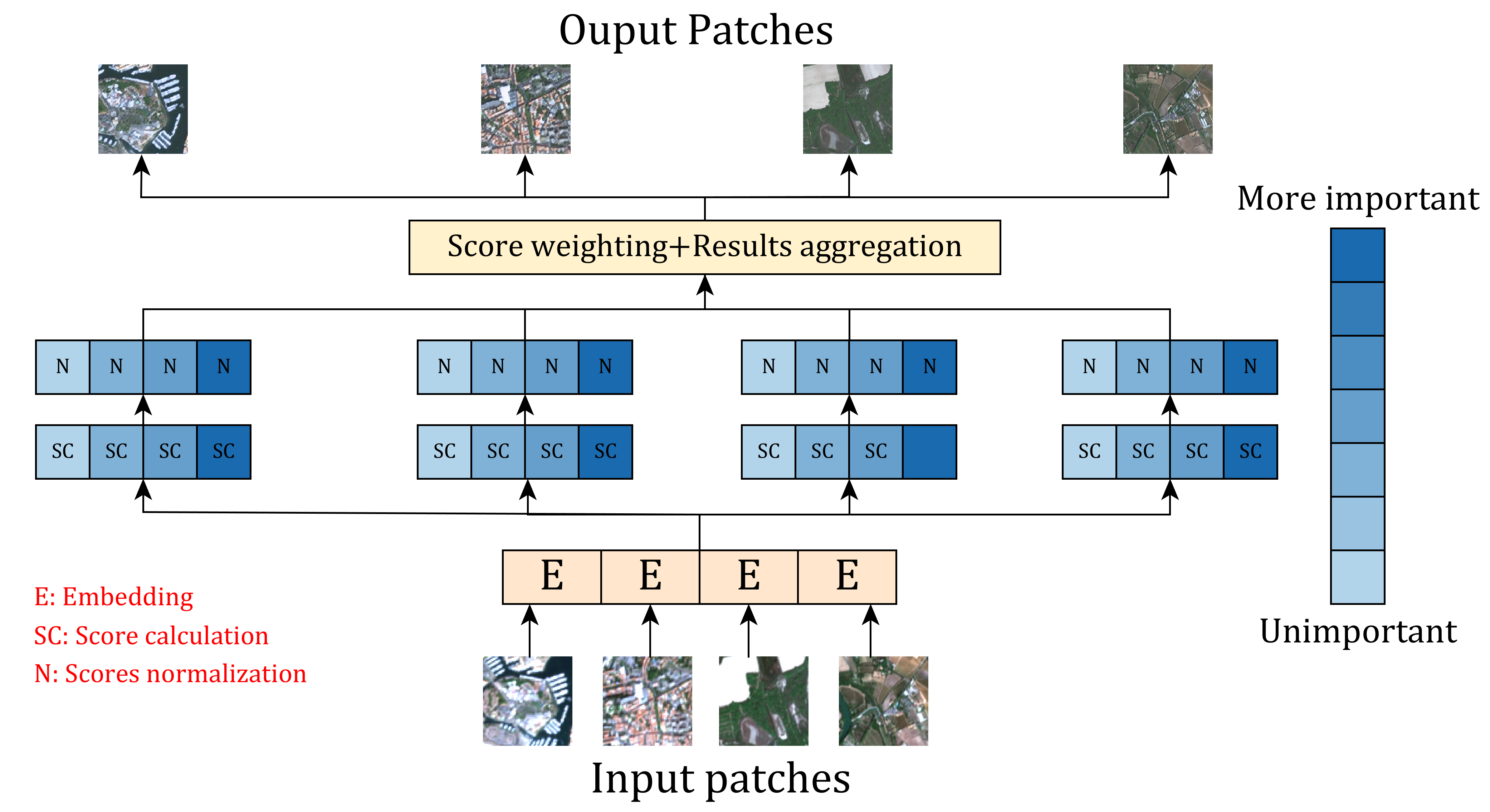}
  \captionsetup{font=scriptsize}
  \caption{Attention mechanisms}.
  \label{fig:bar_chart2}
\end{figure}
\noindent Many creative models are based on these mechanisms, for example: 
\paragraph{Transformers:} introduced by Vaswani et al. \cite{vaswani2017attention}, transformers are architechtures used mainly in Natural Language Processing (NLP), but adapted to multiple applications such as computer vision, via Vision transformers (ViTs) \cite{dosovitskiy2020image}.
\\
As indicated in Figure 12, a transformer consists of an encoder processing the inputs and a decoder generating the output. It is noted that each of these two components is made up of a single layer of feed forward neural networks, as well as residual addition and normalization to preserve the information transmitted by the model, along with positional encoding to manage sequence order. Each encoding-decoding layer is reinforced by a self-attention mechanism in order to extract dependencies between sequence components, including image patches, tokens, etc. In the attention mechanism in question, three vectors are calculated to compute attention scores, the \textbf{Query (Q)} representing the searched element, the \textbf{Key (K)} reflecting the input element, used by the query, and the \textbf{Value (V)} vector referring to the actual value related to the processed component of sequence.

\begin{figure}[H]
  \centering
  \includegraphics[width=0.75\linewidth]{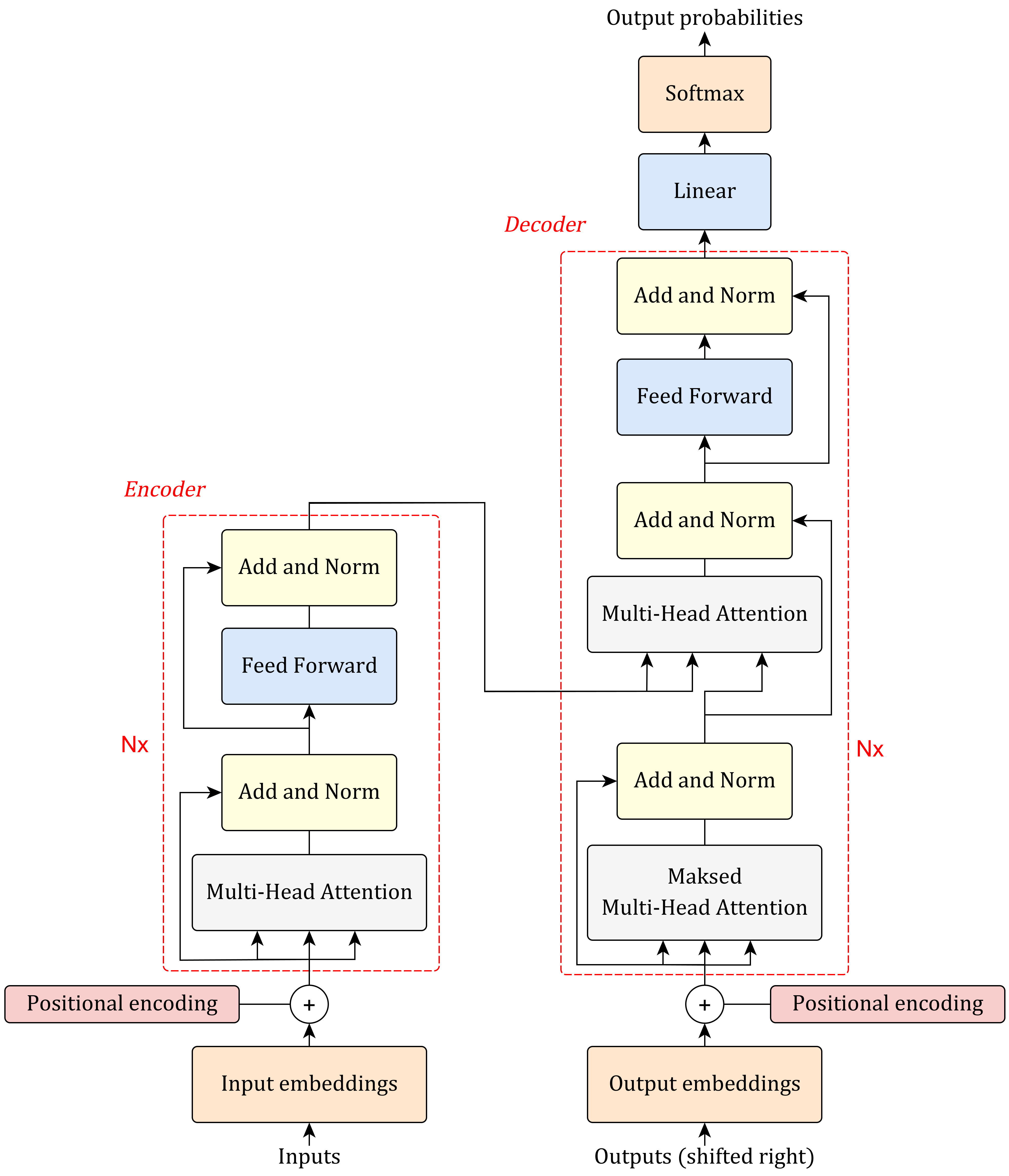}
  \captionsetup{font=scriptsize}
  \caption{Transformers architecture from \cite{vaswani2017attention}}.
  \label{fig:bar_chart2}
\end{figure}

\paragraph{Large Language Models (LLMs):} is a family of NLP models generating coherent, meaningful text from deep neural networks. The associated architecture is characterized by the use of attention mechanisms and a very large-scale text data. Several LLM models have revolutionized the field of NLP, such as Text-to-Text Transfer Transformer (T5) \cite{raffel2020exploring} and Generative Pre-trained Transformer (GPT) \cite{radford2018improving}.
\subsection{Evaluation metrics}
A variety of metrics are employed to evaluate the accuracy of models, depending on the nature of concerned tasks. In this subsection, the metrics mentioned are presented in order to give the reader a clear and concise overview.
\subsubsection{Mean Absolute Error (MAE)}
This is a simple metric for calculating the accuracy of a regression model, by measuring the magnitude of model's prediction errors \cite{hodson2022root}:
\begin{equation}
    \text{MAE} = \frac{1}{n} \sum_{i=1}^{n} |y_i - \hat{y}_i|,
\end{equation}
\noindent
where:
\begin{itemize}
\item $n$ is the number of observations,
\item for an observation  $i$:
\begin{itemize}
    \item $y_i$ and $\hat{y}_i$ are respectively the actual and the predicted value.
    \item $|.|$ is the absolute value.
\end{itemize}
\end{itemize}
\subsubsection{Mean Square Error (MSE)}
Used also to measure the quality of a regression model \cite{plevris2022investigation}, the formula involved is as follows: 
\begin{equation}
    MSE = \frac{1}{n} \sum_{i=1}^{n} (y_i - \hat{y}_i)^2,
\end{equation}

\noindent Root Mean Square Error (RMSE) is a metric estimating the magnitude of prediction errors for a given model \cite{plevris2022investigation}. The associated formula is:
\begin{equation}
    \text{RMSE} = \sqrt{MSE}.
\end{equation}
\subsubsection{Mean Average Precision Error (MAPE)}
MAPE is a metric used mainly in regression tasks in order to calculate the predictive accuracy of a given value \cite{khurram2011effect}. This metric gives a normalized measure of prediction errors and is calculated as follows:

\begin{equation}
\text{MAPE} = \frac{100}{n} \sum_{i=1}^{n} \left| \frac{y_i - \hat{y}_i}{y_i} \right|.
\end{equation}

\noindent Comparing the metrics presented so far, it is clear that MAE offers a more direct measure by giving equal weight to all errors, that MSE is more sensitive to errors than MAE and MAPE, since it amplifies the errors measured. Additionally, the RMSE is easier to interpret than MSE, since it is expressed in the same unit of observations, while MAPE allows relative error to be measured in percentage terms, offering the possibility of evaluating the efficiency of models driven by data at different scales.
\subsubsection{Determination coefficient R²}
R² is a statistical measure, introduced by Wright \cite{wright1921correlation}, to measure the degree to which a regression model truly describes the relationship between its variables, by calculating the proportion of variance of the dependent variable from independent variables in the involved model. Mathematically , this measure is calculated as follows: 
\begin{equation}
    R^2 = 1 - \frac{\sum_{i=1}^{n} (y_i - \hat{y}_i)^2}{\sum_{i=1}^{n} (y_i - \bar{y})^2} \raisebox{0.5ex}{,}
\end{equation}
\\
where, for an observation $i$:
\begin{itemize}[leftmargin=*]
    \item $y_i$ is the real value of the dependent variable,
    \item $\hat{y}_i$ is the value predicted by the model,
    \item $\bar{y}$ is the average of the real values over the n observations of the dependent variable.
\end{itemize}
\subsubsection{Confusion matrix}
This is a visualization of the predictions of a classification model with respect to ground truth values. It has the following form: 
\footnotesize
\[
\begin{tabular}{c|cc}
\textbf{Given class} & \textbf{Predicted: Negative} & \textbf{Predicted: Positive} \\
\hline
\textbf{Ground truth: Positive} & False Negative & True Positive \\
\hline
\textbf{Ground truth: Negative} & True Negative & False Positive \\
\end{tabular}
\]
\normalsize
Confusion matrix gives a detailed overview of the predictions obtained, while clearly identifying the best predicted classes and the errors made \cite{townsend1971theoretical}.
\subsubsection{Precision}
Precision is an indicator measuring the degree of identification of positive predictions for a classification model \cite{sokolova2009systematic}. In the case of binary classification, it is obtained as follows: 
\begin{equation}
\text{Precision} = \frac{TP}{TP + FP} \raisebox{0.5ex}{,}
\end{equation}
where TP represents the number of true positives, i.e. cases correctly identified as positive, and FP is the number of false positives, i.e. cases incorrectly identified as positive. In the case of multiple classes, precision is calculated by averaging the precision obtained for each class using arithmetic or weighted average.
\subsubsection{Recall}
Recall is another measure quantifying the level at which a classification model identifies positive classes from the set of positive instances in measurement sample \cite{sokolova2009systematic}, for binary classification:  
\begin{equation}
\text{Recall} = \frac{TP}{TP + FN}\raisebox{0.5ex}{,}
\end{equation}
where FN is the number of false negatives, i.e. cases incorrectly identified as negative.
\\
Like precision, recall in generic cases is deduced by averaging the recalls of each concerned class. 
\subsubsection{Accuracy}
It is a performance metric, defined as the ratio between the correct predictions of a classification model and the total number of predictions, for the case of binary classification:
\begin{equation}
\text{Accuracy} = \frac{TP+TN}{TP + TN + FP + FN}\raisebox{0.5ex}{.} 
\end{equation}
\subsubsection{F1-score}
This metric evaluates the ability of a classification model to effectively predict positive individuals, by making a compromise between precision and recall \cite{sokolova2009systematic}. The combination of precision and recall is presented in the form of a harmonic mean:
\begin{equation}
F1\text{-}score =2 \times \frac{\text{Precision} \times \text{Recall}}{\text{Recall} + \text{Precision}}\raisebox{0.5ex}{.}
\end{equation}
\noindent F1-score is a metric giving a balanced evaluation of model's performance, useful when there is an imbalance between positive and negative classes.
\subsubsection{Jaccard index}
Jaccard's index, often known as Intersection over Union (IoU), is an evaluation metric used in segmentation and detection tasks. It relates to the ratio between the intersection and the union of predictions and the ground truth \cite{jaccard1901etude}:
\begin{equation}
\text{IoU} = \frac{\text{Area of Intersection}}{\text{Area of Union}}\raisebox{0.5ex}{.}
\end{equation}
\subsubsection{mean Average Precision (mAP)}
It is an evaluation metric often used for detection tasks. It corresponds to the average value of mean accuracies of each class \cite{wang2022parallel}: 
\begin{equation}
    \text{mAP} = \frac{1}{N} \sum_{i=1}^{N} AP_i.
\end{equation}
\noindent For a determined object class i, the average accuracy  $AP_{i}$ is obtained by plotting the model's Accuracy-Recall curve, it corresponds to the area under this curve.
\paragraph{Note:}
It is very common to use the mAP50 metric, measuring the mAP value for an IoU threshold of 50\%, since it offers a compromise between high object location accuracy and error tolerances of this location.

\subsubsection{EAO (Expected Average Overlap)}
It is a metric assessing the performance of object tracking models on videos. It is calculated by averaging the IOUs obtained on the image sequence in question \cite{kristan2020eighth}.
\begin{equation}
\text{EAO} = \frac{1}{N} \sum_{i=1}^{N} \text{IoU}(i),
\end{equation}
\noindent where:
\begin{itemize}[leftmargin=*]
    \item N is the number of image sequences making up the video,
    \item \text{IoU}(i) is the IOU of the ith image.
\end{itemize}

\subsection{Geospatial datasets}
The development of accurate and reliable models depends essentially on the quality of training datasets. In addition, the credibility of the measures used for evaluation also depends on the reliability of the data used. Hence, the need to employ data that is reliable, normalized, uniform in terms of format and structure and distributed evenly to ensure that each category of data involved is representative. Furthermore, a dataset must be consistent, i.e. rich and diversified, to train the desired model properly with regard to all possible scenarios. Also, it must be representative of all possible real-life scenarios, so that the trained model is able to predict never-before-seen scenarios. In addition, a data set must be well labeled, in the case of supervised learning. For example, a dataset of remote sensing images used for aircraft detection needs to be able to cope with a multitude of configurations, i.e. hidden aircraft, different markings, different colors, etc.
\\
Given the specific nature of geospatial data, particularly in terms of data volume and diversity of formats available, standardizing and adjusting these data is essential for producing accurate GeoAI models. Various geospatial datasets are available, allowing to train and validate models on one hand, and to compare the performance of different AI models on the other.
\\
In this subsection, different categories of geospatial data are explored and the main existing geospatial datasets are examined, whether datasets proposed as part of challenges, research or datasets submitted as part of space programs and therefore fed as and when new acquisitions are made. It is noted that, in many cases, geospatial datasets are not defined by the number of instances, but rather by their geographic coverage and spatial resolution.
\subsubsection{Sattelites images datasets}
Earth observation sattelites provide various users with image data at different resolution scales, ranging from low resolution (greater than 100 meters) or medium resolution (typically between 10 and 100 meters) to very high resolution (less than 1 meter) \cite{richards2022remote}. These images, acquired using a variety of techniques, whether passive cameras such as optical or thermal, or active sensors such as radar, are used to capture terrestrial information and exploit it for a variety of applications, including agriculture \cite{shanmugapriya2019applications}, natural resources management \cite{verbyla2022satellite} and urban planning \cite{netzband2007applied}. To benefit from the added value offered by these images, a number of data centers and datasets are publicly available:
\begin{itemize}[leftmargin=*]
    \item \textbf{EOSDIS datasets} : These are public datasets from twelve Distributed Active Archive Centers (DAACs), and belonging to the Earth Observing System Data and Information System (EOSDIS) program of the National Aeronautics and Space Administration (NASA), whose mission is to archive, manage and share various geospatial data. For instance, these data centers include:
    \begin{itemize}[leftmargin=*]
        \item Land Processes Distributed Active Archive Center (LPDAAC) \cite{lpdaac},
        \item Global Hydrology Resource Center DAAC (GHRCDAAC) \cite{ghrcdaac},
        \item National Snow and Ice Data Center Distributed Active Archive Center (NSIDCDAAC) \cite{nsidc_daac},
        \item Goddard Earth Sciences Data and Information Services Center (GES DISC) \cite{nasa_ges_disc},
        \item Physical Oceanography DAAC (PODAAC) \cite{NASA_PO_DAAC}.
    \end{itemize}
    These data centers have their own Application Programming Interface (API) to ensure seamless data consultation and integration into customized platforms. Most of these data are remote sensing images, but other types of data are also available, such as ground station measurements. Table 2 shows the different types of data downloadable free of charge from these data centers, together with their nature and sources.
\end{itemize}
\begin{table*}[ht]
\centering
\captionsetup{font=scriptsize} 
\caption{EOSDIS main data.}
\label{tab:EOSDIS_main_data}
\scriptsize 
\renewcommand{\arraystretch}{1.5} 
\begin{tabular}{|m{7cm}|m{1.5cm}|m{2.5cm}|m{5.5cm}|} 
\hline
\multicolumn{1}{|c|}{\textbf{Data source}} & 
\multicolumn{1}{c|}{\textbf{Datacenter}} & 
\multicolumn{1}{c|}{\textbf{Data type}} & 
\multicolumn{1}{c|}{\textbf{Applications}} \\ 
\hline

Moderate Resolution Imaging Spectroradiometer (MODIS) & LPDAAC & Satellite images. & Environmental applications, including monitoring of vegetation cover, aerosols, land surface. \\ \hline

Ozone Monitoring Instrument (OMI) & GES DISC & Satellite images. & Environmental monitoring and air quality. \\ \hline

Landsat satellites (1 to 9) & LPDAAC & Satellite images. & A wide range of applications including agriculture and urban mapping. \\ \hline

Gravity Recovery and Climate Experiment (GRACE) and GRACE Follow-on satellites & PO DAAC & Satellite images. & Calculation of terrestrial water reserves. \\ \hline

Tropical Rainfall Measuring Mission (TRMM) satellite & GES DISC & Satellite images. & Precipitation. \\ \hline

Global Food Security-support Analysis Data (GFSAD) & LPDAAC & Satellite images. & Agriculture and food. \\ \hline

AErosol RObotic NETwork (AERONET) & GES DISC & Ground station data. & Aerosol measurement. \\ \hline

Integrated Multi-satellite Retrievals for Global Precipitation Model  (IMERG) & GES DISC & Hybrid data from several satellites. & Precipitation. \\ \hline

Climate Hazards Group InfraRed Precipitation with Station data (CHIRPS) & GES DISC & Hybrid data (satellite and ground stations). & Precipitation. \\ \hline

Surface Water and Ocean Topography satellite (SWOT) & PO DAAC & Satellite images. & Water resources management, environment and hydrology. \\ \hline

Soil Moisture Active Passive satellite (SMAP) & NSIDC DAAC & Satellite images. & Ground humidity measurement. \\ \hline
\end{tabular}
\end{table*}
\begin{itemize}[leftmargin=*]
    \item \textbf{Copernicus Data Space Ecosystem}: is a dataset of updated satellite data taken at different dates, namely data from Sentinel satellites (from 1 to 5P) and data from Copernicus Contributing Missions (CCM), implemented in the context of Copernicus program of the European Space Agency (ESA) \cite{Copernicus2024}. As well as EOSDIS datasets, Copernicus Data Space Ecosystem has its own API, enabling users to freely download and use data of various sizes and formats \cite{musial2024overview}.
    \item \textbf{Geo-ImageNet}: is an extension of ImageNet dataset \cite{deng2009imagenet} containing multispectral satellite images and Digital Elevation Models (DEM) \cite{li2023geoimagenet}. Geo-ImageNet is developed to train and validate AI models for geographic object recognition, comprising more than 876 instances spread over 6 object classes. Table 3 shows the distribution of instances according to these classes.
\begin{table}[H]
\centering
\captionsetup{font=scriptsize}
\scriptsize
\renewcommand{\arraystretch}{1.5}
\caption{Geo-ImageNet instances.}
\label{tab:geo_imagenet_instances}
\begin{tabular}{|>{\centering\arraybackslash}m{1cm}|>{\centering\arraybackslash}m{2cm}|}
\hline
\textbf{Class} & \textbf{Instances No.} \\
\hline
Basin   & 155 \\
\hline
Crête   & 171 \\
\hline
Valley  & 181 \\
\hline
Bay     & 93  \\
\hline
Island  & 106 \\
\hline
Lac     & 170 \\
\hline
\end{tabular}
\end{table}

    \item \textbf{xView datasets}: these datasets, launched as part of 3 different challenges, use annotated optical and radar satellite images for classification and object detection tasks \cite{lam2018xview, ritwik2019xbd, paolo2022xview3}. Table 4 provides detailed informations on each dataset. 
\end{itemize}
In addition to public datasets, several earth observation sattelites offering high-resolution images have limited access, given their high accuracy and the sensitivity of some of the areas captured, since these sattelites have a heliosynchronous orbit, as shown in Figure 13, which enables them to acquire any geographical area around the world. Table 5 shows main examples of these sattelites.

\begin{figure}[H]
  \centering
  \includegraphics[width=0.6\linewidth]{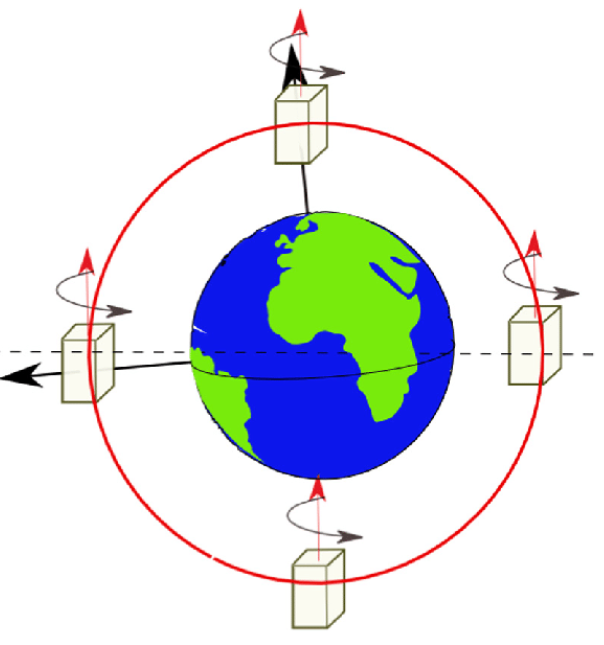}
  \captionsetup{font=scriptsize}
  \caption{Heliosynchronious orbit schema (\cite{heliosynchronious_farrahi}).}
  \label{fig:Year_chart}
\end{figure}
\begin{table*}[ht]
\centering
\captionsetup{font=scriptsize} 
\caption{xView datasets.}
\label{tab:xView_datasets}
\scriptsize 
\renewcommand{\arraystretch}{1.5} 
\begin{tabular}{|m{1.25cm}|m{1.5cm}|m{5cm}|m{1.1cm}|m{1.5cm}|m{1cm}|m{3cm}|} 
\hline
\multicolumn{1}{|c|}{\textbf{References}} & 
\multicolumn{1}{c|}{\textbf{Dataset}} & 
\multicolumn{1}{c|}{\textbf{Task}} & 
\multicolumn{1}{c|}{\textbf{\makecell{Images \\ number}}} & 
\multicolumn{1}{c|}{\textbf{\makecell{Annotations \\ number}}} & 
\multicolumn{1}{c|}{\textbf{\makecell{Classes \\ number}}} & 
\multicolumn{1}{c|}{\textbf{Images source}} \\ 
\hline

\cite{lam2018xview} & xView-1 & Detection of several object classes. & 1127 & One Million & 60 & WorldView-3 30 cm images. \\ \hline

\cite{ritwik2019xbd} & xBD & Assessing the damage caused by natural disasters. & 22068 & 850736 & 6 & Maxar open images. \\ \hline

\cite{paolo2022xview3} & xView-3 SAR & Detecting dark fishing. & 1000 & 243018 & 2 & Sentinel-1 images. \\ 
\hline

\end{tabular}
\end{table*}
\begin{table*}[ht]
\centering
\captionsetup{font=scriptsize} 
\caption{HR and VHR EO satellites samples.}
\label{tab:EO_satellites_samples}
\scriptsize 
\renewcommand{\arraystretch}{1.5} 
\begin{tabular}{|m{2cm}|m{1.5cm}|m{5cm}|m{4.5cm}|m{3.25cm}|}
\hline
\multicolumn{1}{|c|}{\textbf{Satellite}} & \multicolumn{1}{c|}{\textbf{Resolution}} & \multicolumn{1}{c|}{\textbf{Acquisition bands}} & \multicolumn{1}{c|}{\textbf{Organization Manufacturer}} & \multicolumn{1}{c|}{\textbf{Launch date}} \\ \hline

Pléiades 1A/B & 50 cm & Panchromatic (0.48 – 0.83 µm) and Multispectral & Airbus Defence and Space (ADS) & December 17, 2011 (1A) / December 2, 2012 (1B) \\ \hline

GeoEye-1 & 41 cm & Panchromatic (0.45 – 0.90 µm) and Multispectral & General Dynamics & September 6, 2008 \\ \hline

SkySat satellites (1 to 16) & 50 cm & Panchromatic (0.45 – 0.90 µm) and Multispectral & Skybox Imaging / Planet Labs & November 21, 2013 (SkySat-1), as an example  \\ \hline

WorldView-1 & 50 cm & Panchromatic (0.45 – 0.90 µm) & Ball Aerospace & September 18, 2007 \\ \hline

WorldView-2 & 46 cm & Panchromatic (0.45 – 0.80 µm) and Multispectral & Ball Aerospace & October 8, 2009 \\ \hline

WorldView-3 & 31 cm & Panchromatic (0.45 – 0.80 µm) and Multispectral & Ball Aerospace & August 13, 2014 \\ \hline

WorldView-4  & 31 cm & Panchromatic (0.45 – 0.80 µm) and Multispectral & Lockheed Martin & November 11, 2016 \\ \hline

RADARSAT-1 & 8 m & C-band SAR  & MacDonald, Dettwiler and Associates (MDA) & November 4, 1995 \\ \hline

RADARSAT-2 & 1m & C-band SAR  & MacDonald, Dettwiler and Associates (MDA) & December 14, 2007 \\ \hline

ALOS PALSAR-1 & 10 m  & L-band SAR  & Japan Aerospace eXploration Agency (JAXA) & January 24, 2006 \\ \hline

ALOS PALSAR-2 & 1 m  & L-band SAR  & Japan Aerospace eXploration Agency (JAXA) & May 24, 2014 \\ \hline

TerraSAR-X & 1 m  & X-band SAR  & Airbus Defence and Space and German Aerospace Center (DLR) & June 15, 2007 \\ \hline

TanDEM-X & 1 m  & X-band SAR  & Airbus Defence and Space and German Aerospace Center (DLR) & June 21, 2010 \\ \hline
\end{tabular}
\end{table*}

\noindent As well as sattelites data, several geospatial data sources, such as aerial imagery, LIDAR, trajectory and address data, provide an ideal entry point for training different forms of models. Table 6 summarizes the main examples of these datasets, presented in the remainder of this subsection.

\begin{table*}[ht]
\centering
\captionsetup{font=scriptsize} 
\caption{Geospatial Datasets with Characteristics and Applications.}
\label{tab:geospatial_datasets}
\scriptsize 
\renewcommand{\arraystretch}{1.5} 
\begin{tabular}{|m{2.25cm}|m{4.5cm}|m{3cm}|m{4.75cm}|m{2cm}|} 
\hline
\multicolumn{1}{|c|}{\textbf{Dataset Category}} & \multicolumn{1}{c|}{\textbf{Dataset Name}} & \multicolumn{1}{c|}{\textbf{Principal Characteristics}} & \multicolumn{1}{c|}{\textbf{Applications}} & \multicolumn{1}{c|}{\textbf{References}} \\ \hline

\multirow{5}{*}{Aerial and UAV images} 
& Aerial Image Dataset (AID) & Ten thousand images. & Object detection and classification tasks. & \cite{xia2017aid} \\ \cline{2-5}
& National Agriculture Imagery Program (NAIP) & Resolution of 1m. & Agricultural lands classification. & \cite{NAIP} \\ \cline{2-5}
& University of California Merced (UCM) land use dataset. & 2100 images. & Object detection and segmentation tasks. & \cite{yang2010bag} \\ \cline{2-5}
& UAVid & 30 videos and 300 images. & Segmentation of urban areas. & \cite{lyu2020uavid} \\ \cline{2-5}
& UAVSAR & Resolution of 1 and 3m. & Monitoring coastal deformations. & \cite{uavsar} \\ \hline

\multirow{3}{*}{Data Elevation Models} 
& Copernicus DEM & Resolution of 90m / 30m. & Terrain modelling. & \cite{cepernicusdem2022} \\ \cline{2-5}
& ASTER Global DEM & Resolution of 30m. & Orthorectification. & \cite{usgsearthexplorer} \\ \cline{2-5}
& Shuttle Radar Topography Mission (SRTM) & Resolution of 30m. & Elevation modelling. & \cite{opentopography} \\ \hline

\multirow{5}{*}{LIDAR Datasets} 
& NASA's Global Ecosystem Dynamics Investigation (GEDI) & Altimetric resolution of 1m. & Forest analysis. & \cite{NASAGEDILiDAR} \\ \cline{2-5}
& USGS 3DEP & Resolution of 10 / 30m. & Precise 3D modelling. & \cite{heidemann2012lidar} \\ \cline{2-5}
& UK Environment Agency LIDAR data & Resolution of 1m / 50 cm. & Precise 3D modelling. & \cite{UKEnvironmentAgencyLIDAR} \\ \cline{2-5}
& Toronto-3D LIDAR dataset & Centimetric resolution. & Urban modelling. & \cite{Tan2020CVPRWorkshops} \\ \cline{2-5}
& NSW Marine LiDAR Topo-Bathy 2018 & Resolution of 5m. & Hydrographic mapping. & \cite{NSWMarineLiDARTopoBathy2018} \\ \hline

\multirow{2}{*}{Vector Datasets} 
& OpenStreetMap & Global coverage. & Navigation and urban exploration. & \cite{openstreetmap} \\ \cline{2-5}
& Google Maps vector data & Global coverage. & Navigation. & \cite{googlemaps} \\ \hline

\multirow{3}{*}{Trajectory Datasets} 
& Geolife Trajectories dataset & 17621 trajectories. & Geolocation and tracking. & \cite{geolife2012} \\ \cline{2-5}
& ExactEarth Satellite AIS tracking system & Global coverage. & Maritime traffic. & \cite{marinetrafficexactearth} \\ \cline{2-5}
& EuRoC MAV dataset & Multimodal dataset from 3 sources. & Autonomous vehicle navigation. & \cite{Burri25012016} \\ \hline

\multirow{4}{*}{Address Datasets} 
& GeoNames & 11 million addresses. & Geocoding and addressing systems. & \cite{geonameswebsite} \\ \cline{2-5}
& Australia Post’s GNAF & 13 million addresses. & Nationwide geocoding and addressing systems. & \cite{datagovau} \\ \cline{2-5}
& National address database (France) & 25 million addresses. & Precise nationwide addressing. & \cite{adressedatagouv2024} \\ \cline{2-5}
& ISPARK dataset & Istanbul city coverage. & Parking lot management. & \cite{ispark2024} \\ \hline

\end{tabular}
\end{table*}

\subsubsection{Aerial and UAV images}
One of the main advantages of aerial photography and drone imagery is the very high resolution it offers \cite{zongjian2008uav}. Thanks to the low altitude of acquisition, the images produced enable more robust information extraction and more powerful analysis capabilities. Several datasets offer users the possibility of exploiting the performance of these images.
\begin{itemize}[leftmargin=*]
    \item \textbf{Aerial Image Dataset (AID)}: as shown in Figure 14, it is an annotated dataset of aerial images from different areas around the world. It is created by Xia et al. \cite{xia2017aid} to help AI researchers benchmark the different models to be exploited. AID dataset contains 10,000 images of 0.5m to 2m resolution belonging to 30 land use classes, and publicly available in \cite{aiddataset}.
\begin{figure}[H]
  \centering
  \includegraphics[width=\linewidth]{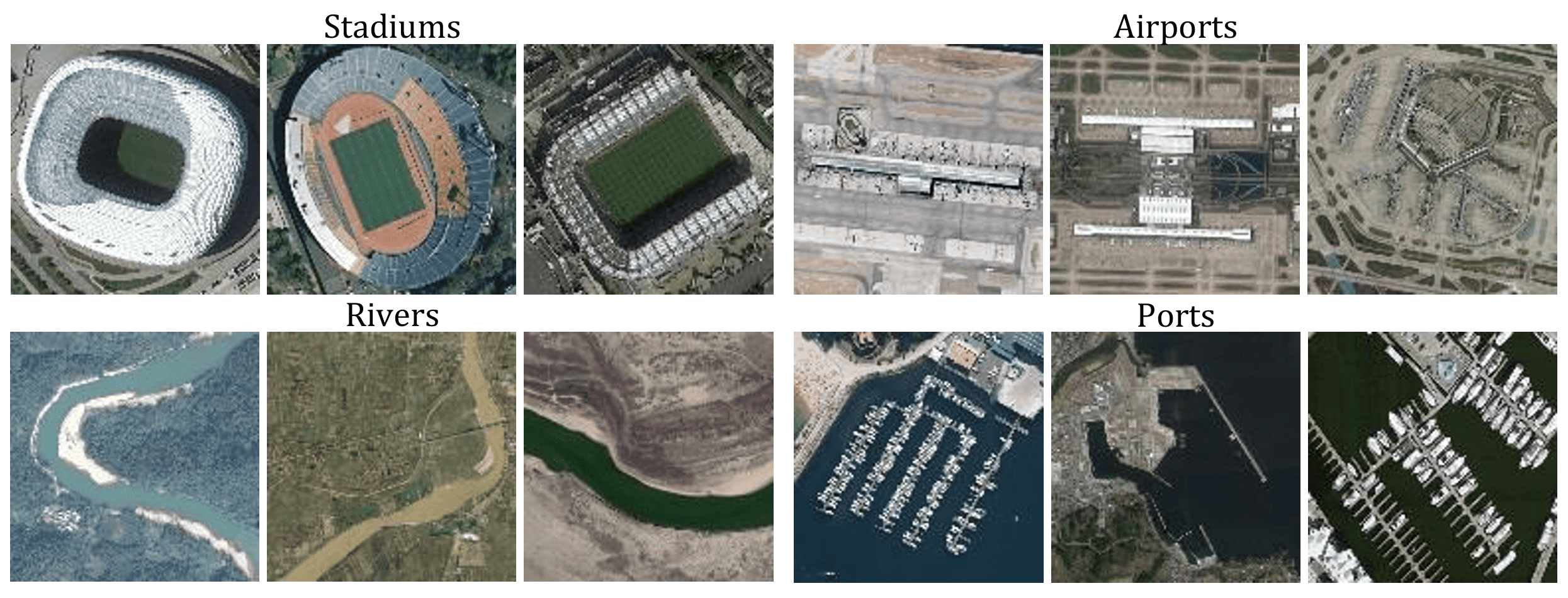}
  \captionsetup{font=scriptsize}
  \caption{Examples of AID classes (\cite{xia2017aid}).}
  \label{fig:Year_chart}
\end{figure}
    \item \textbf{National Agriculture Imagery Program (NAIP) public dataset}: is a public program launched in 2003 by the United States Department of Agriculture Farm Service Agency (FSA) \cite{usdafsa}. The purpose of this dataset is to provide free, very high resolution aerial imagery of agricultural areas in the U.S, providing a suitable entry point for a variety of applications, such as land classification and yields estimation \cite{NAIP}. Available in GeoTIFF format and supported by various platforms, such as United States Geological Survey (USGS) Earthdata \cite {usgsearthdata}, Google Earth Engine (GEE) \cite{googleearthengine} or its official portal \cite{NAIP}, NAIP dataset images have a spatial resolution of around 1m and offer RGB and Near InfraRed (NIR) band images.
    \item \textbf{University of California Merced (UCM) land use dataset}: is an annotated database of 0.3 m resolution aerial images from the United States. This dataset, comprising 21 land use classes such as buildings, airports and parking lots, and totalling 2100 images \cite{yang2010bag}, is used for training and benchmarking classification models, and it is accessed via Kaggle platform \cite{ucmercedkaggle}. 
    \item \textbf{UAVid}: is a UAV benchmarking dataset for semantic segmentation tasks. This annotated dataset, created by Lyu et al. \cite{lyu2020uavid} in 2020, totals over 8 classes with 30 videos and 300 drone images, captured from urban scenes, and it is accessible to the public via the official website \cite{uaviddataset}.
    \item \textbf{UAVSAR}: is a public dataset relating to the Uninhabited Aerial Vehicle Synthetic Aperture Radar program launched by NASA's Jet Propulsion Laboratory (JPL) \cite{jplnasa}, enabling users to use high-resolution images, from 1 to 3m, through SAR sensors mounted mainly on UAVs, but also on manned aircraft. The dataset is utilized in a variety of applications, such as landslide studies, flood monitoring and natural risk assessment \cite{uavsar}.
\end{itemize}
\subsubsection{Data elevation models}
According to Guth et al. \cite{guth2021digital}, Data Elevation Models (DEM) refers to a georeferenced representation of heights at the earth's surface. As shown in Figure 15, this raster data is representative of the bare ground, known as Data Terrain Models (DTM), or of the surface, otherwise known as Data Surface Models (DSM). DEMs are extracted from optical or radar satellite images, LIDAR point clouds, or from aerial images, providing a precise topographic representation of the terrain, enabling the construction of 3D visualizations, contour lines or slope maps, while facilitating the extraction of geographic information \cite{rayburg2009comparison}.
\begin{figure}[H]
  \centering
  \includegraphics[width=\linewidth]{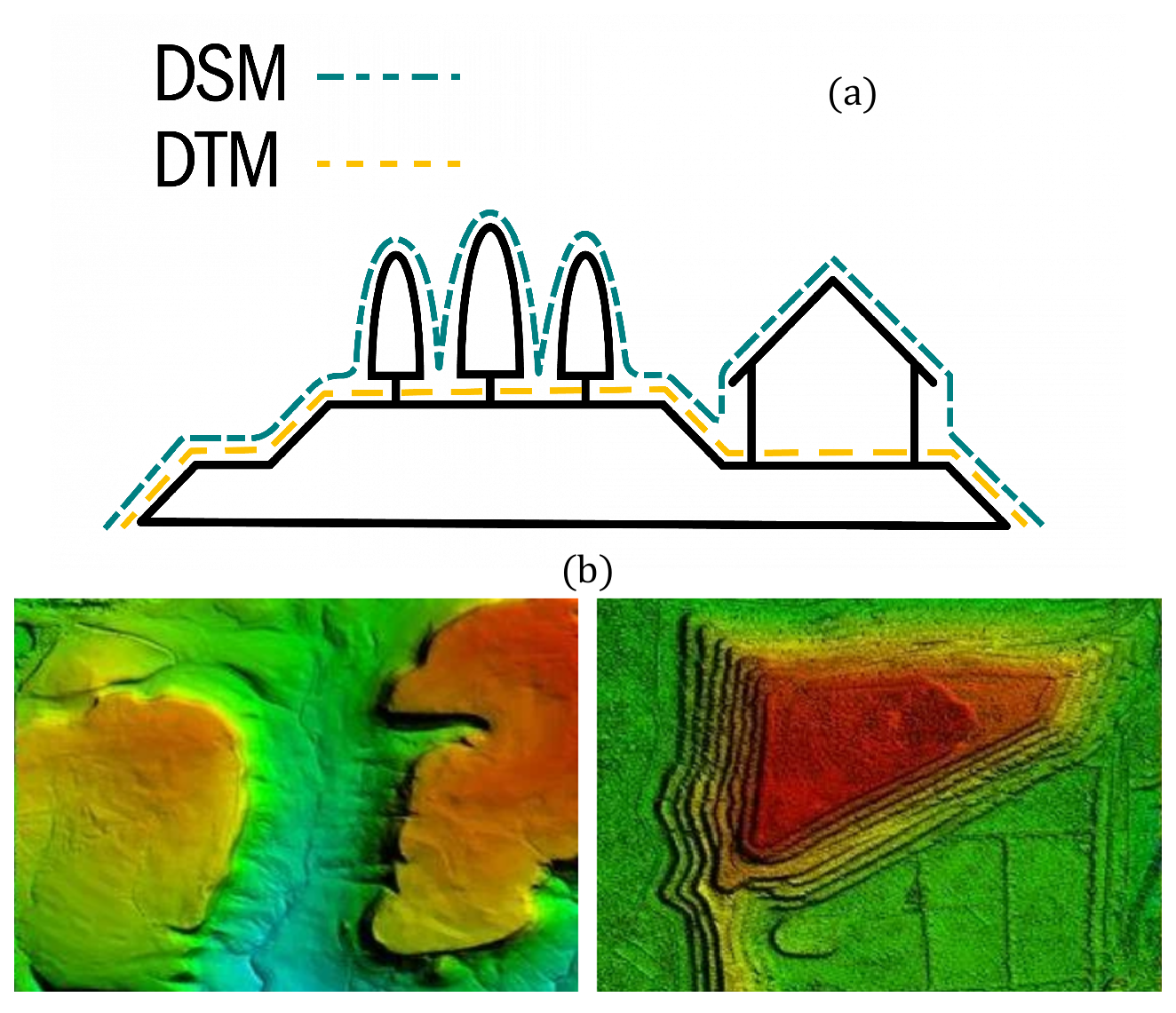}
  \captionsetup{font=scriptsize}
  \caption{Explanations of the difference between DSM and DTM by horizontal view (a) \cite{NRCanElevationData} and by raster view (b) \cite{shustrik2024elevation}.}
  \label{fig:Year_chart}
\end{figure}
\begin{itemize}[leftmargin=*]
    \item \textbf{Copernicus DEM}: contains digital surface models (DSMs) from TanDEM-X satellite between 2011 and 2015 \cite{cepernicusdem2022}. These DSMs offer a resolution of 90 meters for GLO-30 products and 30 meters for GLO-90 products, as well as a resolution of 10 meters only on the European continent for EEA-10 products, with a good altimetric accuracy Linear error 90\% (LE90) of 4m. Note that LE90 is the distance within which 90\% of linear altimeter positioning errors lie in relation to their actual position on the ground \cite{federal1998national}.
\begin{figure}[H]
  \centering
  \includegraphics[width=\linewidth]{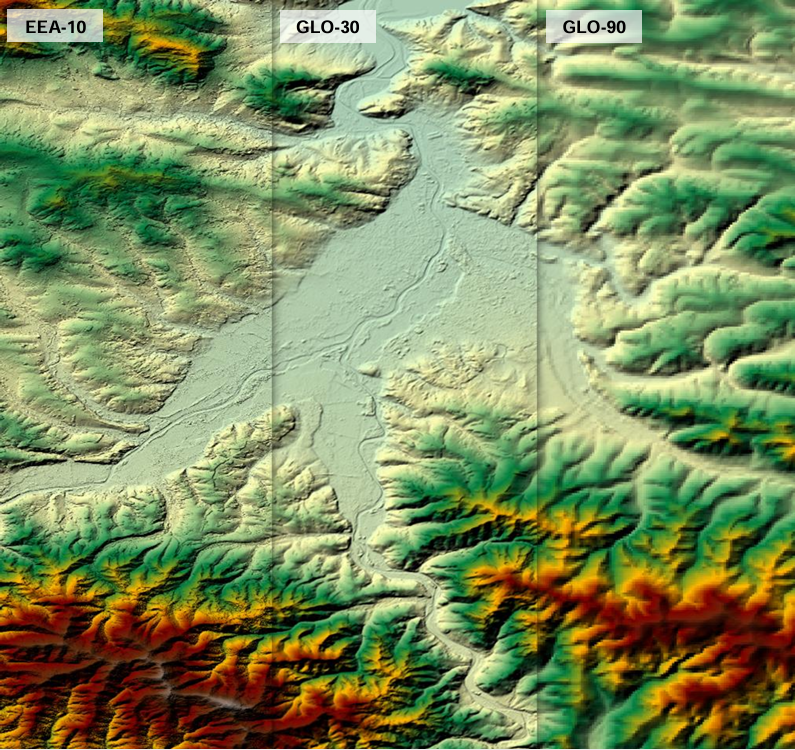}
  \captionsetup{font=scriptsize}
  \caption{Extract from a 3 categories copernicus DEM (\cite{cepernicusdem2022}).}
  \label{fig:Year_chart}
\end{figure}
    \item \textbf{ASTER Global Data Elevation Model (DEM)}:  is a public global DTM produced from optical stereo pair images from NASA's Advanced Spaceborne Thermal Emission and Reflection Radiometer (ASTER) \cite{abrams2010aster}. This model has a resolution of 30 meters and it is accessible via various platforms such as United States Geological Survey (USGS) earthexplorer \cite{usgsearthexplorer} and NASA earthdata portal \cite{nasaearthdata}.
    \item \textbf{Shuttle Radar Topography Mission (SRTM)}: this SAR Band-X mission, combining the efforts of NASA and the German space agency, aims to produce a global DTM. As of 2014, the resulting model is publicly available in 30-meter resolution, and it is used in numerous applications such as orthorectification and topographic correction. SRTM data are publicly available for 80\% of the earth's surface on various platforms such as Open Topography \cite{opentopography}.
\end{itemize}
\subsubsection{LIDAR datasets}
Thanks to their precision, Light Detection And Ranging (LIDAR) data provides a detailed 3D representation of objects and surfaces in all directions, through resolved point clouds of up to thousands of points per square meter. The main LIDAR acquisition categories are:
\begin{itemize}[leftmargin=*]
    \item Space-based LIDAR, i.e. on board of sattelites,
    \item Airborne LIDAR, via aircraft or drones,
    \item Land-based LIDAR, including mobile or stationary LIDAR,
    \item Bathymetric LIDAR, operating wavelengths in the green spectrum to penetrate water.
\end{itemize}
Multiple LIDAR datasets with different resolutions are made available to users to train and to validate models of all kinds:
\begin{itemize}[leftmargin=*]
    \item \textbf{NASA's Global Ecosystem Dynamics Investigation (GEDI) LIDAR} : it is an open dataset to assess the world's forestry structure, using a LIDAR instrument on board of the International Space Station (ISS) \cite{NASAGEDILiDAR}. Each laser pulse has a circular footprint of 25m, and the footprints have a spatial sampling frequency of 60m and an altimetric resolution of 1m. Several levels of processing are provided from the raw LIDAR data, for instance, Figure 17 illustrates an extraction of GEDI data, corresponding to Arial Ground Biomass Density (AGBD), referring to the quantity of plant biomass per unit area \cite{sialelli2024agbd}. This dataset is publicly accessible and it is available in LPDAAC datacenter \cite{lpdaac}.
\begin{figure}[H]
  \centering
  \includegraphics[width=\linewidth]{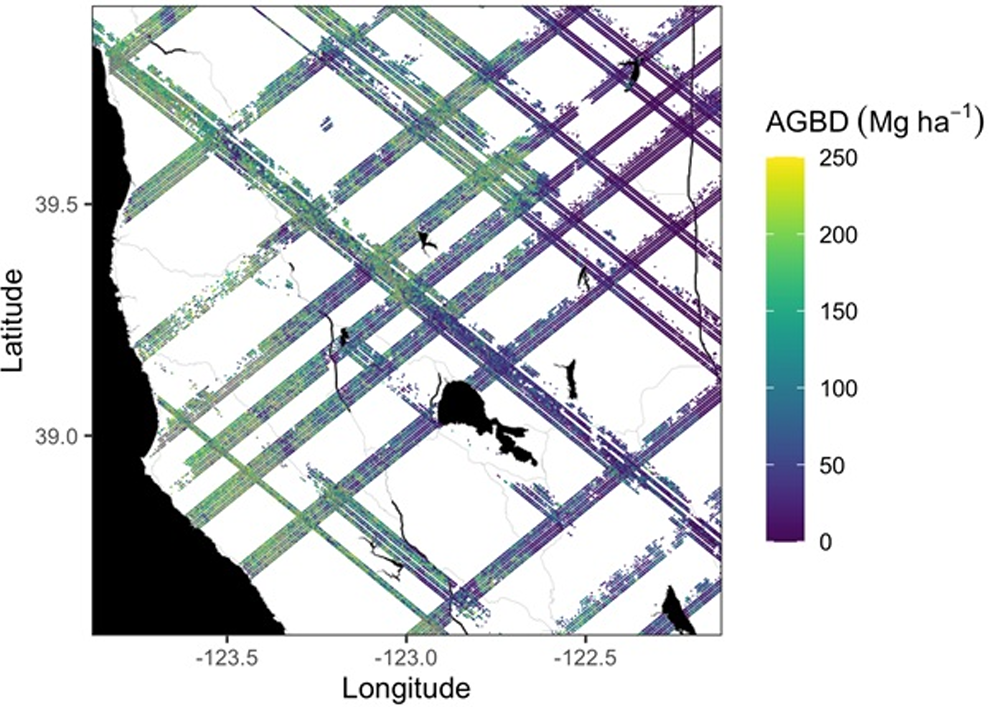}
  \captionsetup{font=scriptsize}
  \caption{Footprints of AGBD values (in Mg ha-1), acquired from April to July 2019 (\cite{dubayah2022gedi}).}
  \label{fig:Year_chart}
\end{figure}
    \item \textbf{USGS 3DEP LIDAR data} : is a USGS open dataset shared in the context of 3D Elevation Program (3DEP), the aim of this program is to provide users with 3D airborne LIDAR data for the whole of the United States \cite{heidemann2012lidar}. This dataset is available at a resolution of 30 m for rural areas and 10 m for cities \cite{callahan2022vertical}. 
    \item \textbf{The UK Environment Agency LIDAR Data} : this public dataset contains airborne-LIDAR data at the scale of England, with a spatial resolution of 1m, enhanced to 0.5m for the densest areas, as well as derived products, i.e. DSMs and DTMs \cite{UKEnvironmentAgencyLIDAR}.
    \item \textbf{Toronto-3D LIDAR dataset} : is an annotated dataset comprising mobile terrestrial LIDAR data to train semantic segmentation and classification models. This dataset features centimetric horizontal and vertical resolution, while capturing the urban roads of the city of Toronto in Canada \cite{Tan2020CVPRWorkshops}, an example of which is shown in Figure 18.
\begin{figure}[H]
  \centering
  \includegraphics[width=\linewidth]{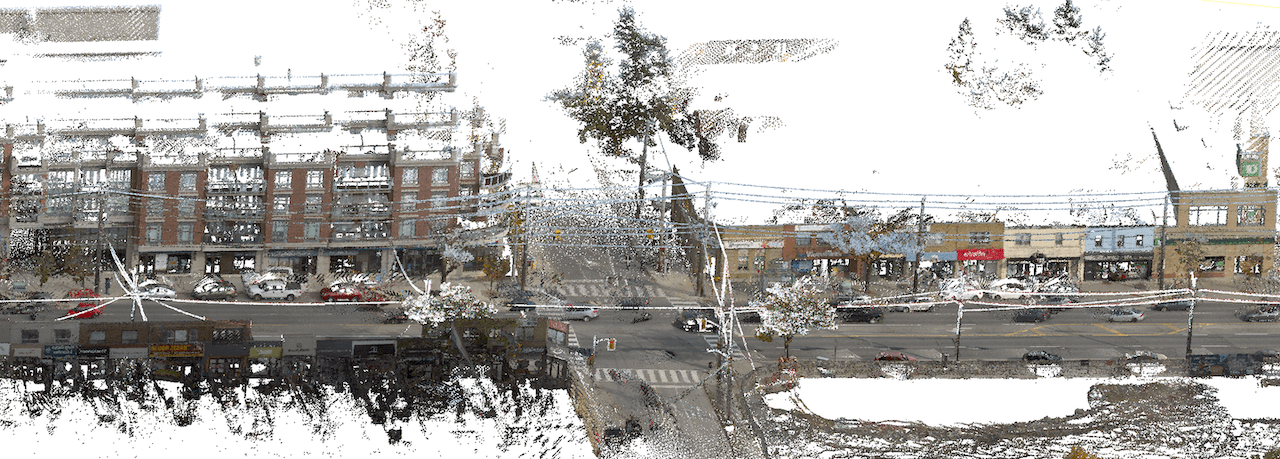}
  \captionsetup{font=scriptsize}
  \caption{Example of Toronto-3D LIDAR dataset (\cite{Tan2020CVPRWorkshops}).}
  \label{fig:Year_chart}
\end{figure}
    \item \textbf{NSW Marine LiDAR Topo-Bathy 2018} : This dataset combines airborne topographic radar and bathymetric LIDAR, covering the coastal areas of New South Wales (NSW), Australia. With a horizontal resolution of 5m and vertical resolution in the decimeter range, this dataset enables precise coastal mapping, and serves as an ideal entry point for training mapping models \cite{NSWMarineLiDARTopoBathy2018}.
\end{itemize}
\subsubsection{Vector data}
Vector data, as digital representations of geographic reality in the three standard geometric forms of point, line and polygon, provide a rich source of information while facilitating the geospatial analysis process. The reasonable size of these data in memory and the possibility of extracting them from several sources, including images and maps, encourage users to employ them in various applications, among which vector geospatial datasets:
\begin{itemize}[leftmargin=*]
    \item \textbf{Microsoft Global Building Footprint} : is a collection of vectorial building footprints extracted from satellite images using semantic segmentation, achieving an average accuracy of 90\% \cite{GlobalMLBuildingFootprints}. It is a vast collection of over 1.3B footprints across 5 continents, using images derived from Maxar \cite{maxar}, Airbus \cite{airbus_defence_space} and IGN France \cite{ign_france}. Figure 19 illustrates a concrete example of this data.  
\begin{figure}[H]
  \centering
  \includegraphics[width=0.8\linewidth]{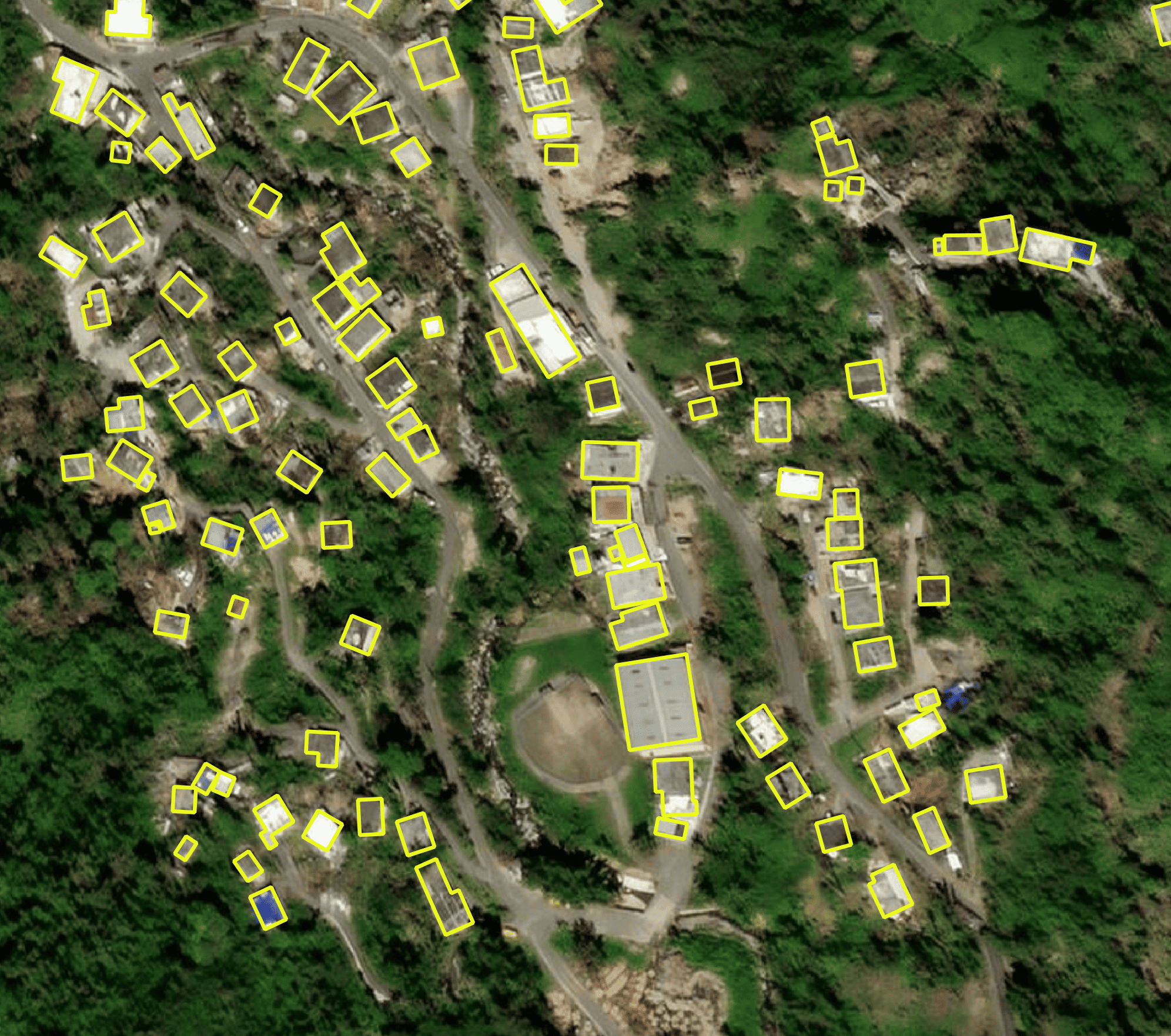}
  \captionsetup{font=scriptsize}
  \caption{Microsoft building footprints samples (\cite{GlobalMLBuildingFootprints}).}
  \label{fig:Year_chart}
\end{figure}
    \item \textbf{OpenStreetMap} : this is vector data from the OpenStreetmap collaborative mapping project, founded in 2004. The goal of this project is to provide a public representation of the world according to the land use in three standard vector forms, i.e. points such as restaurants and hotels, lines including roads and waterways, and polygons for instance parks and lakes \cite{mooney2017review}. This data is downloadable from the official Openstreetmap website \cite{openstreetmap}, via multiple platforms such as Geofabrick \cite{geofabrik}, and BBBike \cite{bbbike}.
\begin{figure}[H]
  \centering
  \includegraphics[width=0.7\linewidth]{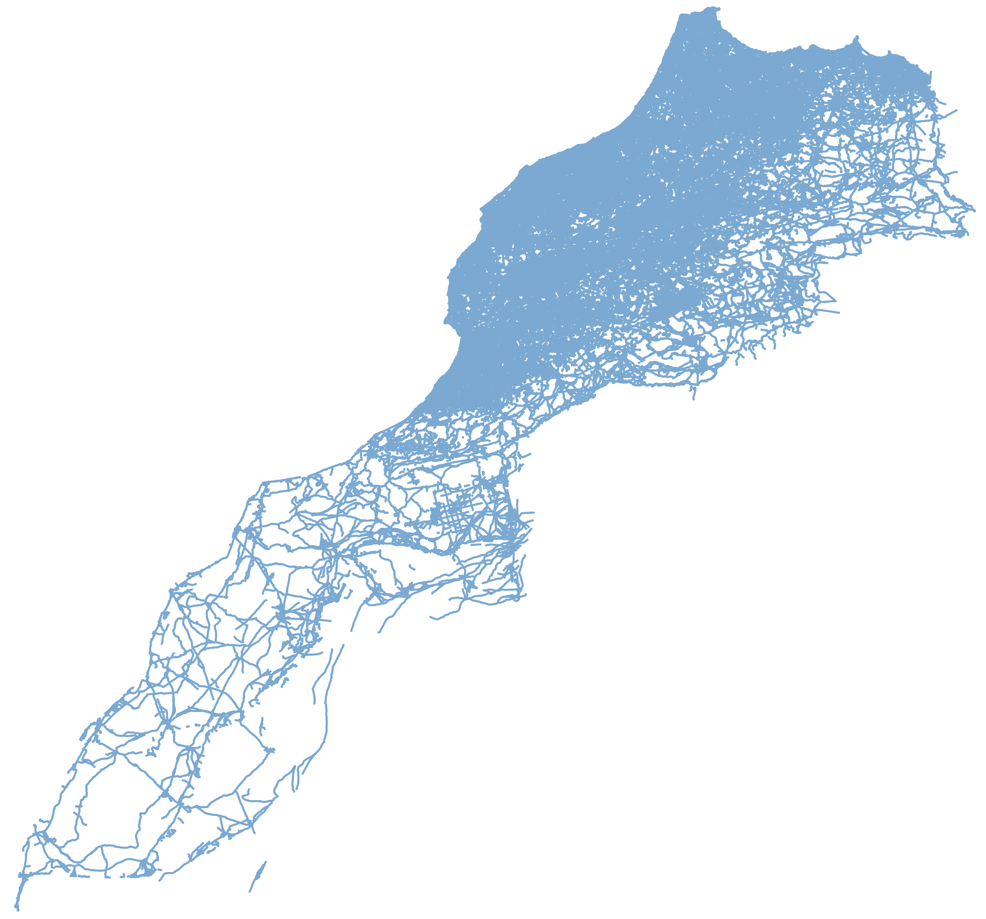}
  \captionsetup{font=scriptsize}
  \caption{OSM roads network, example from Morocco, updated 08/23/2024.}
  \label{fig:Year_chart}
\end{figure}
    \item \textbf{Google maps vector data} : as part of its mapping services, i.e. Google maps \cite{googlemaps} and Google earth \cite{googleearth}, Google offers vector tiles representing various geographical features, including roads, buildings, etc. Although this data is protected by user rights, due to the cost of its collection and production, it can be used via Google Maps APIs, such as Places API \cite{googleplacesapi} and Directions API \cite{googledirectionsapi}.
\end{itemize}
\subsubsection{Trajectory datasets}
Trajectory data, such as Global Navigation Satellite System (GNSS) and Inertial Navigation System (INS) data, enable the precise location of objects and people. This is particularly useful for analyzing the behavior of human groups and the flow of resources and energy over time and space. In this subsection, the main examples of trajectory datasets are presented.
\begin{itemize}[leftmargin=*]
    \item \textbf{Geolife Trajectories dataset}: is a set of GPS trajectories from more than 178 users in over 30 Chinese cities. Collected by Microsoft Research \cite{microsoftresearch}, this dataset contains a total of 17621 trajectories, covering 1251654 Km over a period of 48203 hours. Figure 21 shows the density of trajectories recorded in the city of Beijing, also, it is noted that this dataset is freely available on Kaggle platform \cite{geolife2012}.
\begin{figure}[H]
  \centering
  \includegraphics[width=0.8\linewidth]{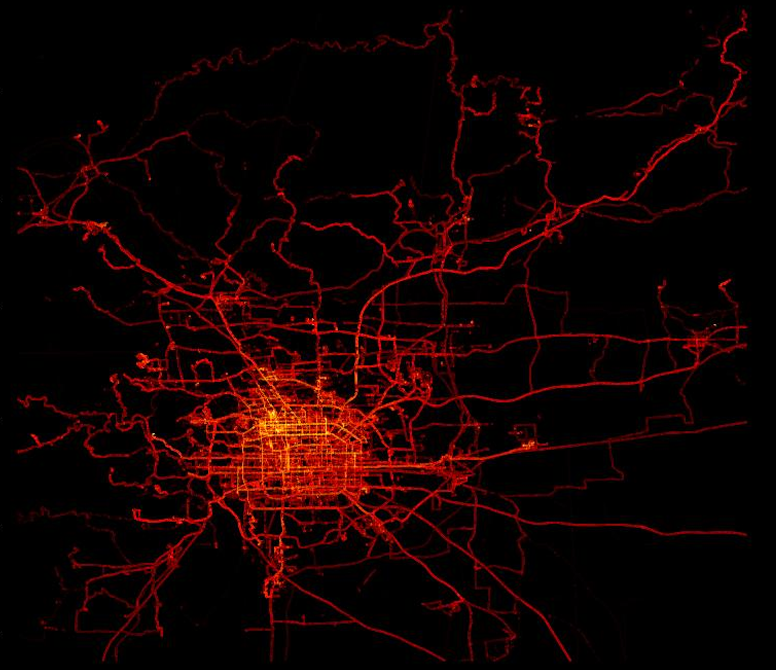}
  \captionsetup{font=scriptsize}
  \caption{Geographical extent of GeoLife dataset for the city of Beijing.}
  \label{fig:Year_chart}
\end{figure}
    \item \textbf{European Robotics Challenges - Micro Aerial Vehicles (EuRoC MAV)}: is a public dataset containing mainly inertial navigation system (INS) data, acquired from inertial measurement unit (IMU) sensors such as accelerometers and gyroscopes, as well as GPS trajectories and stereoscopic images, from micro-aerial vehicles (MAVs). The aim of this dataset is to analyze the behavior of these vehicles, in order to further research in the context of associated navigation, localization and operational models \cite{Burri25012016}.
    \item \textbf{ExactEarth Satellite AIS tracking system}: is a private dataset from the satellite-based Automatic Identification System (AIS) \cite{fournier2018past}. Using radio frequency (RF) waves, ships can transmit their GPS position, speed and heading, providing real-time global data on the trajectory of maritime traffic through the AIS system \cite{marinetrafficexactearth}.
\end{itemize}
\subsubsection{Adress datasets}
These datasets correspond to a set of information about specific geographical locations, allowing to pinpoint their precise location and describe them in detail. This is particularly important when an analysis of locations is suggested, such as collecting customer reviews, responding to emergency alerts, managing water and electricity services, etc. Examples of such datasets include the following:
\begin{itemize}[leftmargin=*]
    \item \textbf{GeoNames} : is a geographic database containing over 11 million addresses worldwide, including geographic coordinates, zip codes, population and more. These data, whose global density is illustrated in Figure 22, are available under \cite{geonameswebsite}.
\begin{figure}[H]
  \centering
  \includegraphics[width=\linewidth]{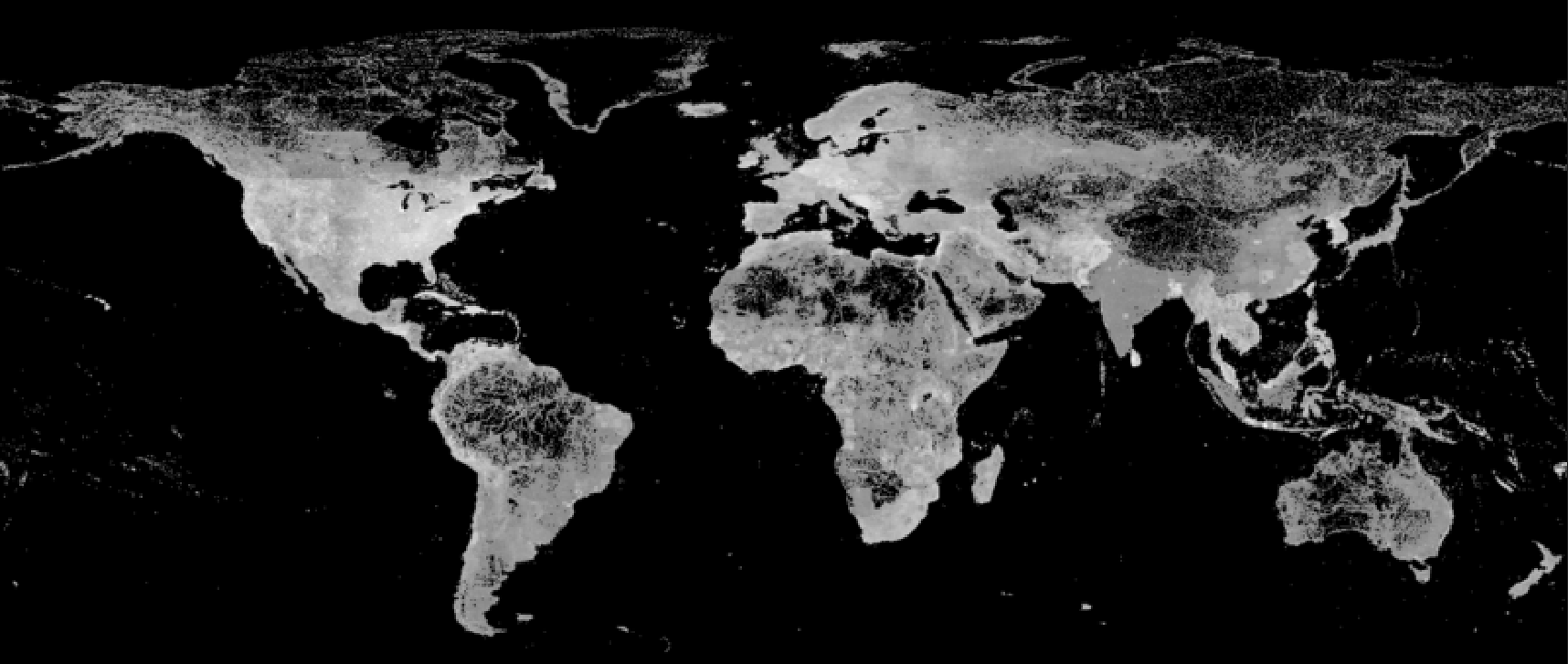}
  \captionsetup{font=scriptsize}
  \caption{Geonames density map (\cite{geonamesdensitymap2006}).}
  \label{fig:Year_chart}
\end{figure}
    \item \textbf{Australia Post’s Geocoded National Address File (GNAF)} : is a public database containing over 13 million addresses in Australia. This dataset is characterized by two elements: geographic coordinates and location labeling according to the Australian administrative system. GNAF is freely accessible via \cite{datagovau}.   
    \item \textbf{National address database} : The French government have launched a collaborative project between the Interministerial Digital Department \cite{dinum2024}, the National Agency for Territorial Cohesion \cite{anct2024} and the National Geographic Institute \cite{ign2024}, in order to provide the public with a geolocated, daily-updated national address database. This open database, as illustrated in Figure 23, is available in several formats, containing over 25 million addresses, and it is downloadable from \cite{adressedatagouv2024}.
\begin{figure}[H]
  \centering
  \includegraphics[width=\linewidth]{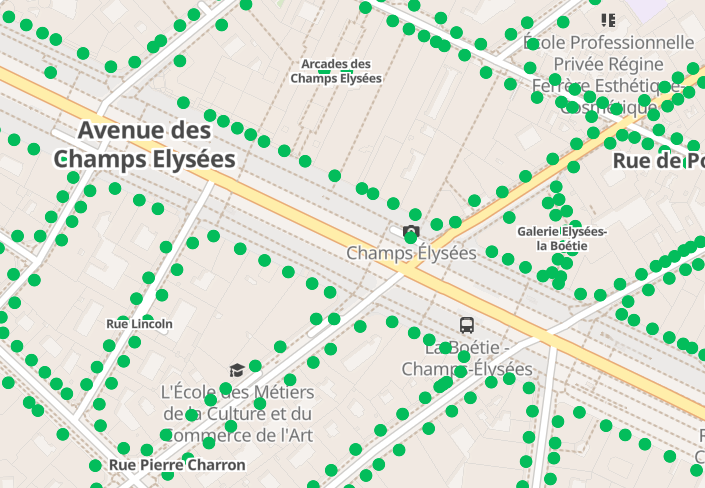}
  \captionsetup{font=scriptsize}
  \caption{French adress database sample from Paris city, OpenStreetMap basemap.}
  \label{fig:Year_chart}
\end{figure}
    \item \textbf{ISPARK dataset} : is a public dataset created by the municipality of Istanbul, Turkey, containing all the city's parking lot addresses, their capacities and the associated tariffs \cite{ispark2024}.
\end{itemize}
\noindent An essential point to raise is the relevance of public-access datasets, given that sharing data helps to improve its scope and reliability, especially in the case of collaborative datasets such as Openstreetmap \cite{openstreetmap} and Geonames \cite{geonameswebsite}. Moreover, the transparency of these datasets enables their reliability to be measured autonomously, via a variety of third parties and methods. In return, the use of these data in a wide range of research projects validates the results of these studies, guaranteeing the scientific community a reference base from which to compare the studies carried out. In the next subsection, the hardware aspect is explored, including its different categories and impact on the performance of various types of models, as well as the latest hardware infrastructures designed for optimal AI efficiency.
\subsection{Hardware}
Pioneering progress in the development of hardware architectures and components has boosted the performance and efficiency of AI, both in terms of training models and predicting results for different cases. An understanding of available hardware opportunities is therefore necessary to choose the right hardware configuration for a given AI task. In this context, a range of hardware components involved in these processes are distinguished, including hardware accelerators, defined as hardware performing a special function faster than CPUs. To illustrate, numerous hardware accelerators are cited: 

\begin{itemize}[leftmargin=*]
    \item Graphics Processing Unit (GPU) \cite{owens2008gpu},
    \item Field-Programmable Gate Array (FPGA) \cite{rose1993architecture},
    \item Digital Signal Processor (DSP) \cite{eyre2000evolution},
    \item Tensor Processing Unit (TPU) \cite{jouppi2017datacenter},
    \item Application-Specific Integrated Circuit (ASIC) \cite{einspruch2012application}. 
\end{itemize}

\noindent In addition to these advanced accelerators, there are storage devices such as Solid-State Drives (SSDs) \cite{lee2008case}, revolutionizing the optimization and performance of data entry and access, together with networking hardware, useful in cases where, for example, numerous users request predictions of a given model.
\\
Several researchers have carried out comparative studies of hardware implementation of AI algorithms. For example, Abu Talib et al. \cite{talib2021systematic} provided a systematic review comparing the performance of a multitude hardware accelerators when running DNNs, especially FPGA, GPU and ASIC, and concluded that FGPAs provide a high reconfiguration capacity, despite their limitation in terms of computing power. GPUs, on the other hand, demonstrated a high execution capacity thanks to their parallel architecture, optimising high computation operations. However, ASICs show high efficiency in specific tasks, for instance video processing on mobile devices, object recognition using embedded systems, while demonstrating significant energy optimization.
\\
In a different context, Dally et al. \cite{dally2018hardware} have taked the example of CNNs to examine the requirements of the basic operations involved, essentially convolution and matrix multiplication. For training operations, multiplication operations require 16-bit floating point precision and results are summed using 32 bits, making computational precision a priority in the design of electronic circuits. For prediction operations, a much lower precision (8 bits) is more than sufficient to perform calculations, making energy optimization and time reduction a priority in this case. This difference in design priorities for each type of operation makes it necessary to optimize them in order to adapt to all requirements, i.e. energy savings, speed and calculations efficiency. Additionally, Zhao et al. \cite{zhao2023tectonic} present Techtonic-shift, a novel infrastructure to improve I/O storage request procedures for ML training tasks, while reducing energy demand. The proposed infrastructure combines Tectonic large-scale distributed file systems \cite{pan2021facebook} with integrated Flash storage \cite{klimovic2016flash}, while introducing memory cache management protocols in order to optimize invloved demands. 
\\
Given the size and diversity of geospatial data, GeoAI requires high-performance hardware to ensure that the data in question can be trained effectively. The use of accelerators such as GPUs , FPGAs and TPUs with considerable RAMs is recommended to perform the concerned tasks. For example, GPUs are perfectly suited for computer vision tasks such as remotely sensed images segmentation, while FPGAs are more suited for GeoAI tasks that require real-time processing, such as vehicle tracking and AIS data census. In addition, storage devices, whether local or in the cloud, must take into account the huge volume of data, while enabling the data used to be stored, accessed and manipulated quickly enough. There is also a strong need to optimize hardware architectures, as this naturally implies improving GeoAI model training and inference processes, both in terms of time and energy consumption.
\section{GeoAI applications}
Having identified research questions and corresponding research directions, this section paves the way for a deeper understanding of GeoAI applications for precision agriculture, environment and natural disaster monitoring, water resource management, urban planning and healthcare. Figure 24 shows the details of these research axes and sub-axes. Indeed, the theme of artificial intelligence applied to geospatiality is addressed not only as an asset for processing and manipulating complex data, but also as a powerful and advanced decision-making tool that is methodological, effective and efficient. 
\begin{figure}[H]
    \centering
    \includegraphics[width=\linewidth]{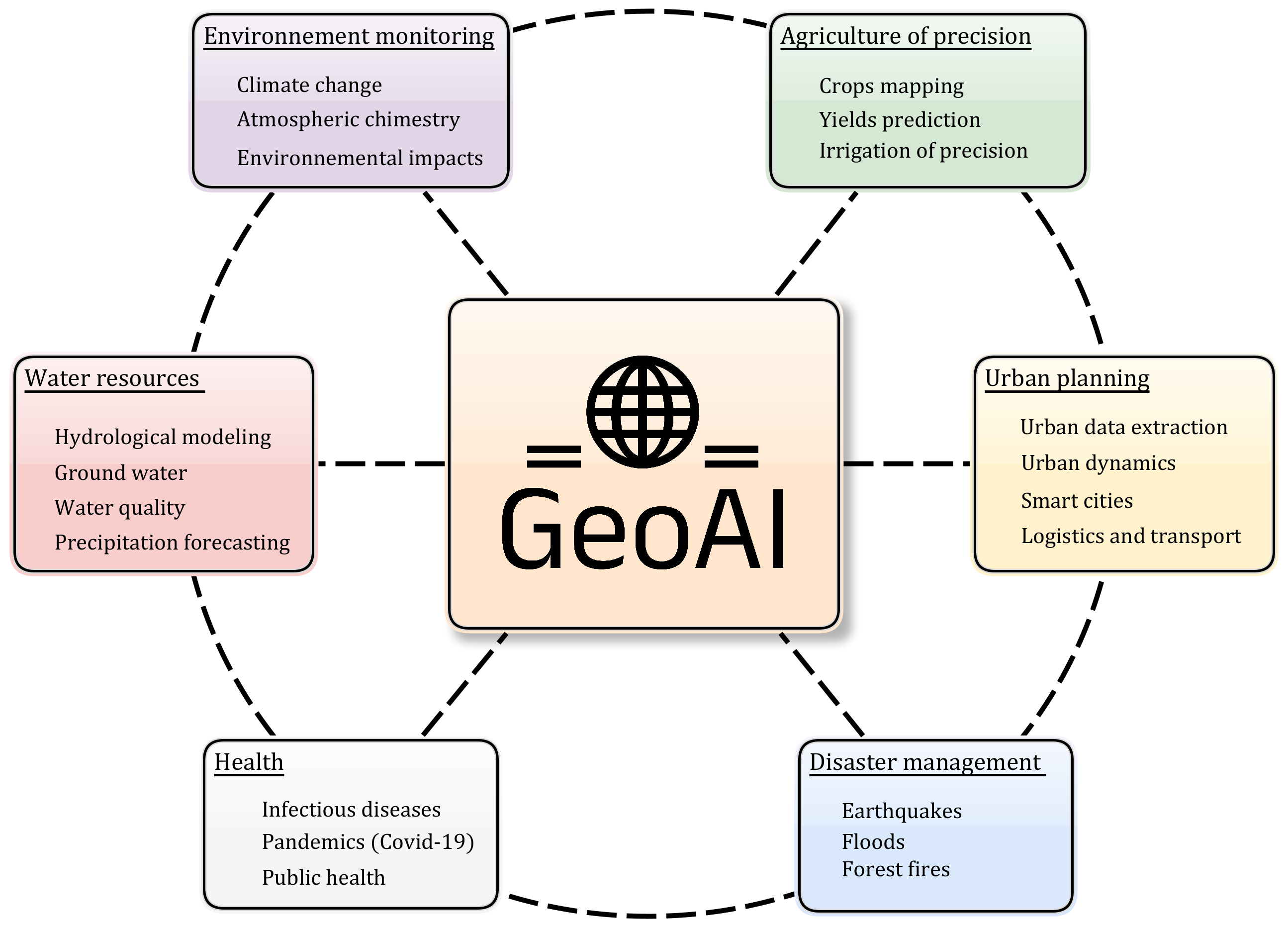}
    \captionsetup{font=scriptsize}
    \caption{Research axes for GeoAI.}
    \label{fig:label-de-votre-figure3}
\end{figure}
\subsection{Precision agriculture}
According to Pierce and Nowak \cite{pierce_aspects_1999}, precision agriculture refers to methods that use new technology to manage different aspects of agricultural production in order to improve productivity, boost crop yields and control the associated environmental quality. Consequently, the close collaboration between precision agriculture and GeoAI has led to a better understanding of the needs and challenges of the agricultural sector, enabling to offer cutting-edge support, whether through precision mapping at the finest scales, yield prediction or precision irrigation. Table 7 shows the main methods and models explored in this field.
\begin{table*}[ht]
\centering
\captionsetup{font=scriptsize} 
\caption{Proposed AI methods for precision agriculture.}
\label{tab:AI_methods_agriculture}
\scriptsize 
\renewcommand{\arraystretch}{1.5} 
\begin{tabular}{|m{3.25cm}|m{10.35cm}|m{1.25cm}|} 
\hline
\multicolumn{1}{|c|}{\textbf{Applications}} & \multicolumn{1}{c|}{\textbf{Methods and algorithms}} & \multicolumn{1}{c|}{\textbf{References}} \\ \hline

\multirow{2}{*}{Crops mapping and classification} 
& Mask Region-based Convolutional Neural Network combined with Random Forest, Support Vector Machine and Multiple Linear Regression. & \cite{peng_combination_2023} \\ \cline{2-3}
& Linear mixed-effect. & \cite{wang_estimation_2019} \\ \hline

\multirow{2}{*}{Yields prediction} 
& Random forest. & \cite{carneiro_soil_2023} \\ \cline{2-3}
& Random Forest and boosting methods. & \cite{ramzan_multimodal_2023} \\ \hline

\multirow{2}{*}{Precision irrigation} 
& Mask Region-based Convolutional Neural Network and ResNext-101. & \cite{de_albuquerque_dealing_2021} \\ \cline{2-3}
& U-Net architecture. & \cite{raei_deep_2022} \\ \hline                       
\end{tabular}
\end{table*}

\subsubsection{Crops mapping}
It is clear that the integration of artificial intelligence methods into the process of mapping agricultural yields, thereby producing added value geographical data, enables significant progress to be made in detecting the key factors influencing yields, including land preparation and fertilisation, crop protection against perticides, the automation of tedious tasks, the identification of large-scale crop types and the effective planning of targeted activities. Relevant studies successed to achieve a high level of control while using different categories of AI models and geospatial data. Han et al. \cite{han_spatio-temporal_2023} presented an innovative spatio-temporal multi-level attention (STMA) model for crop mapping using Sentinel 1 SAR images, including three datasets from Germany \cite{wolff2020identifying}, France \cite{garnot2021pastis}, and South Africa \cite{westerncapeagriculture2021}, covering various crop types. The proposed model includes ResNet network \cite{he2016deep} for feature extraction, a cross-attention mechanism to regularize data dimensions, a spatio-temporal self-attention module to extract relationships between spatial and temporal dimensions and a decoder based on U-Net \cite{ronneberger2015u} architecture to reconstruct the feature map. Overall Accuracy (OA) and F1-Score metrics are used during the evaluation phase, obtaining average values across all classes of 96\% and 74\% respectively.
\begin{figure}[H]
    \centering
    \includegraphics[width=0.9\linewidth]{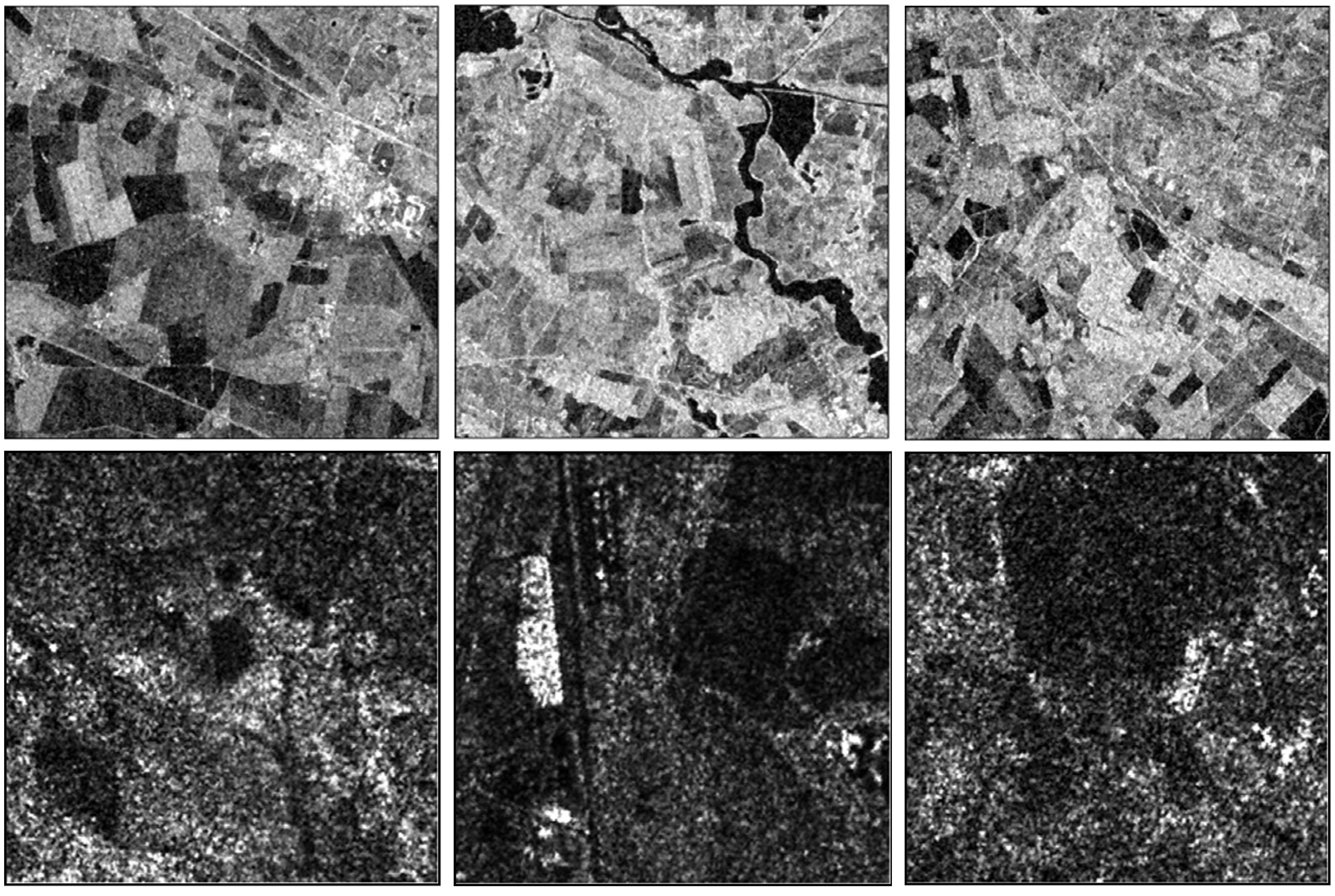}
    \captionsetup{font=scriptsize}
    \caption{Examples of SAR Sentinel1 images used in \cite{han_spatio-temporal_2023}.}
    \label{fig:label-de-votre-figure2}
\end{figure}
\noindent A classificaton model based on satellite imagery is proposed by Gallo et al. \cite{gallo_-season_2023} for the automatic mapping of seasonal crops through Sentinel-2 images along with old crop maps. The concerned model is composed of two parts, a 3D Feature Pyramid Network (FPN3D), which is a 3D extension of Feature Pyramid Network (FPN) \cite{lin2017feature}, and an aggregator CNN allowing annual segmentation. As shown in Figure 26, training dataset is constructed using phenological analysis to enhance historical crop maps, as well as recent Sentinel-2 satellite imagery. FPN3D model is then generalized through a CNN aggregator to produce an accurate seasonal map of agricultural crops. A test phase is carried out while obtaining an overall accuracy (OA) of 73\% over four successive seasons. Tiwari et al \cite{tiwari_automated_2024} combined Sentinel-1 SAR and Sentinel-2 optical imagery to identify the transplation and peak seasons of rice crops, using two approaches, random forest classification and multi-Otsu approach \cite{liao2001fast}. Results showed a classification accuracy of up to 94\%.
\begin{figure*}[ht]
    \centering
    \includegraphics[width=0.75\linewidth]{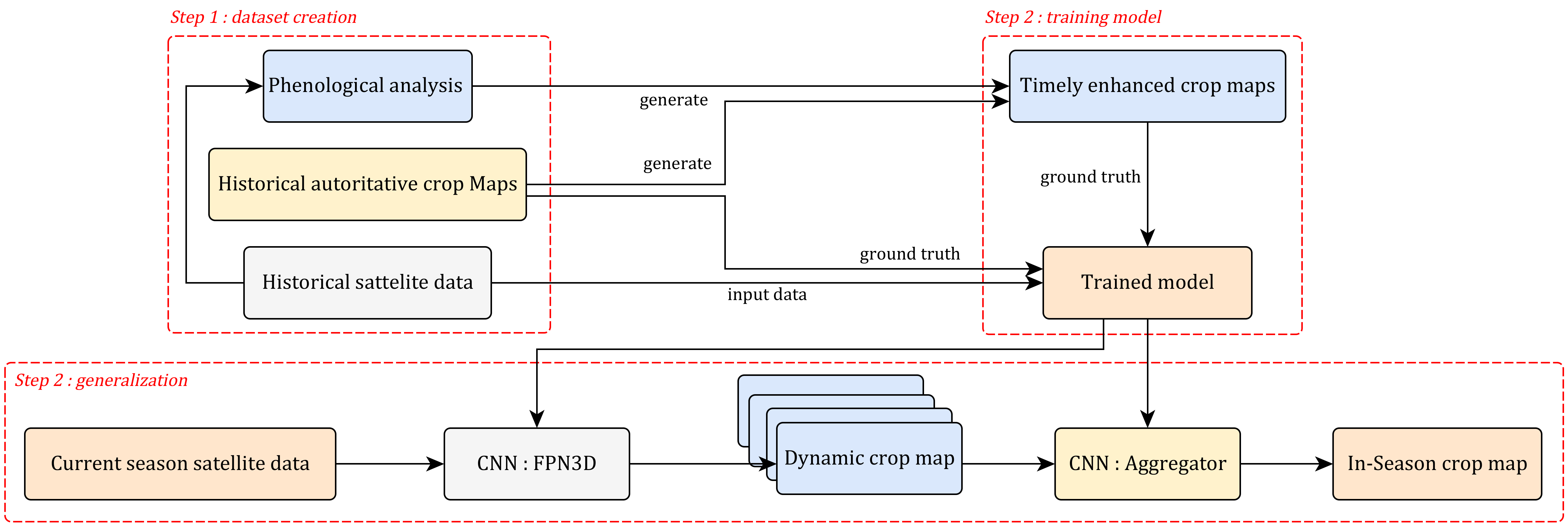}
    \captionsetup{font=scriptsize}
    \caption{Workflow for the automatic mapping  of seasonal crops (\cite{gallo_-season_2023}).}
    \label{fig:label-de-votre-figure2}
\end{figure*}
\noindent Following the same logic, Mohammadi et al. \cite{mohammadi_improvement_2023} used time series of Landsat 7 and 8 data, as well as old geographic yield layers to improve the efficiency and accuracy of cartographies, especially when the model is confronted to unseen data. The approach is to supervise the proposed 3D Fully Convolutional Network (FCN) layers to understand the spatio-temporal relationships between data series. An evaluation carried out on unseen samples demonstrates a high classification performance, with an average IOU of 81\% and an F1-Score of 90\%.
\\
Apart from satellite imagery, several studies explored the importance of high-resolution UAV imagery, for more accurate crop mapping, particularly in small-scale study areas. Ribero et al. \cite{ribeiro_automated_2023} evaluated three different traditional convolution-based networks, specifically U-Net \cite{ronneberger2015u}, LinkNet \cite{chaurasia2017linknet} and Pyramid Scene Parsing Network (PSPNet) \cite{zhao2017pyramid} for the automatic segmentation of crop rows in sugarcane fields. To train and evaluate the models, the authors used senseFly Sensor Optimized For Drone Application (SODA) images, an example of which is shown in Figure 27. Results of validation on four samples showed marked stability, with U-Net network performing relatively better than the other two models in terms of Dice coefficient, being a statistical similarity metric \cite{dice1945measures}, used in this context to assess segmentation capacity. Table 8 shows the Dice coefficient values on the said samples.

\begin{table}[H]
\centering
\captionsetup{font=scriptsize}
\scriptsize
\renewcommand{\arraystretch}{1.5}
\caption{Comparison between U-Net, LinkNet and PSPNet on SODA dataset (\cite{ribeiro_automated_2023}).}
\label{tab:comparison_unet_linknet_pspnet}
\begin{tabular}{|>{\centering\arraybackslash}m{1cm}|>{\centering\arraybackslash}m{1cm}|>{\centering\arraybackslash}m{1cm}|>{\centering\arraybackslash}m{1cm}|}
\hline
\textbf{Sample} & \textbf{U-Net} & \textbf{LinkNet} & \textbf{PSPNet} \\ \hline
A & 90.8\%  & 89.1\%  & 87.1\% \\  \hline
B & 92.3\%  & 90.3\%  & 88.6\% \\  \hline
C & 90.4\%  & 88.4\%  & 86.6\% \\ \hline
D & 86.7\%  & 83.6\%  & 84.1\% \\ \hline
\end{tabular}
\end{table}

\begin{figure}[H]
    \centering
    \includegraphics[width=\linewidth]{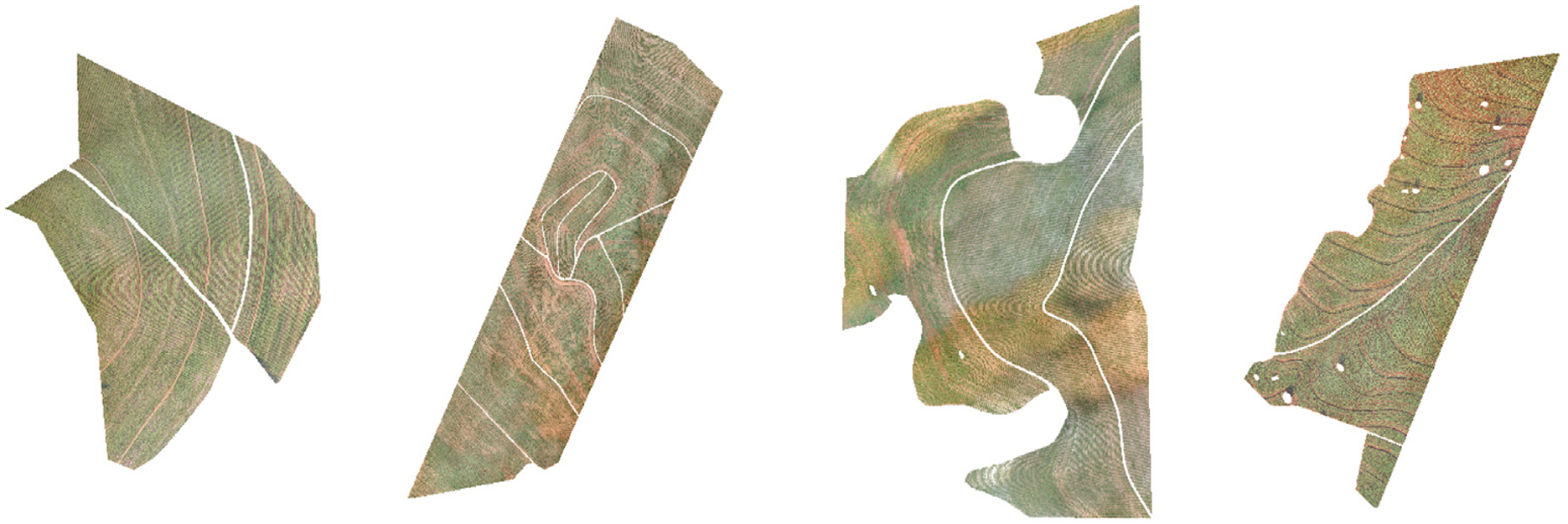}
    \captionsetup{font=scriptsize}
    \caption{Samples of data used for crops segmentation (\cite{ribeiro_automated_2023}).}
    \label{fig:label-de-votre-figure2}
\end{figure}
\noindent Research works cited in this subsection has demonstrated the capabilities of GeoAI for predictive mapping of agricultural crops, including the use of satellite imagery at different resolutions, historical crop maps for fairly accurate classification, and ground truth data for validating the performance of the AI models in question. This mapping technique facilitates the evaluation of the impact of drought and climate change on agricultural production, along with the identification of resulting yields, this is of direct relevance to the next subsection.
\subsubsection{Yields prediciton}
Several studies exploited advanced AI techniques to develop accurate models for yields prediction. Carneiro et al. \cite{carneiro_soil_2023} used a random forest classifier to identify the most significant factors in predicting cotton yield. Additionally, a random forest regressor is used to generalize this prediction on the basis of the classifier results. The authors have exploited data of various types and have shown that indicators derived from Sentinel-2 satellite images, real-time kinematic (RTK) GPS and LIDAR data are the most decisive. Subsequently, results showed an R² of 62\%, an MAE of 0.16 and a MAPE of 5\% for the five most relevant factors. Ramzan et al. \cite{ramzan_multimodal_2023} have combined the exploitation of Landsat-8 satellite data converted to Normalized Difference Vegetation Index (NDVI) \cite{carlson1997relation}, a vegetation index using red and near-infrared (NIR) bands to measure vegetation density on a given surface, along with agro-meteorological data, to train a multi-modal AI model. In this way, the authors use MLP, support vector regression \cite{drucker1996support} and Gaussian random projection, referring to a dimensuality reduction technique based on matrices whose elements follow a Gaussian distribution \cite{bingham2001random}. Ensemble learning methods are also employed, specifically random forest as a bagging method, extreme gradient boosting and XGBoost as boosting methods. Finally, a DNN with 3 hidden layers is used to estimate tea yield at farm scale. The resulting model proved its performance by achieving an R² of 99\%.
\\
Given the high data accuracy provided by UAV imagery, several works address yield estimation using UAV data. Peng et al. \cite{peng_combination_2023} used an original approach to conveniently and economically estimate wheat crop yields by using UAV images, example of which are shown in Figure 28. The first goal of this study is to extract the phenotypic characteristics of wheat spikes, using a Region-based Convolutional Neural Network (R-CNN) \cite{he2017mask}. Then, the yield is estimated using classical ML algorithms, specifically Random Forest Regression (RFR), Multiple Linear Regression (MLR) and Support Vector Regression (SVR). A test phase is executed, resulting in an F1-Score of 83\% for the Mask-RCNN. In addition, R² coefficient is used to examine the performance of RFR, MLR and SVR models, obtaining values of 83\%, 85\% and 86\% respectively. In the same context, Wang et al. \cite{wang_estimation_2019} used a Linear Mixed-Effect (LME) model, referring to a regression model considering the relationships between the independent variables and a dependent variable, modelling the intrinsic variability of the data by means of an intrinsic term in the regression equation \cite{bennington1994use}. The ultimate aim of this research is to accurately estimate the leaf area index (LAI) \cite{zheng2009retrieving} of rice crop. Diverse ML models are used to compare the performances obtained, in particular simple regression (SR), artificial neural network (ANN) and RF classifier and regressor, while using RGB UAV images and rice variety informations. The results showed that LME model is the most accurate, achieving an R² of 76\% to 81\% and an RMSE of 1.04 to 1.16.
\begin{figure}[H]
    \centering
    \includegraphics[width=\linewidth]{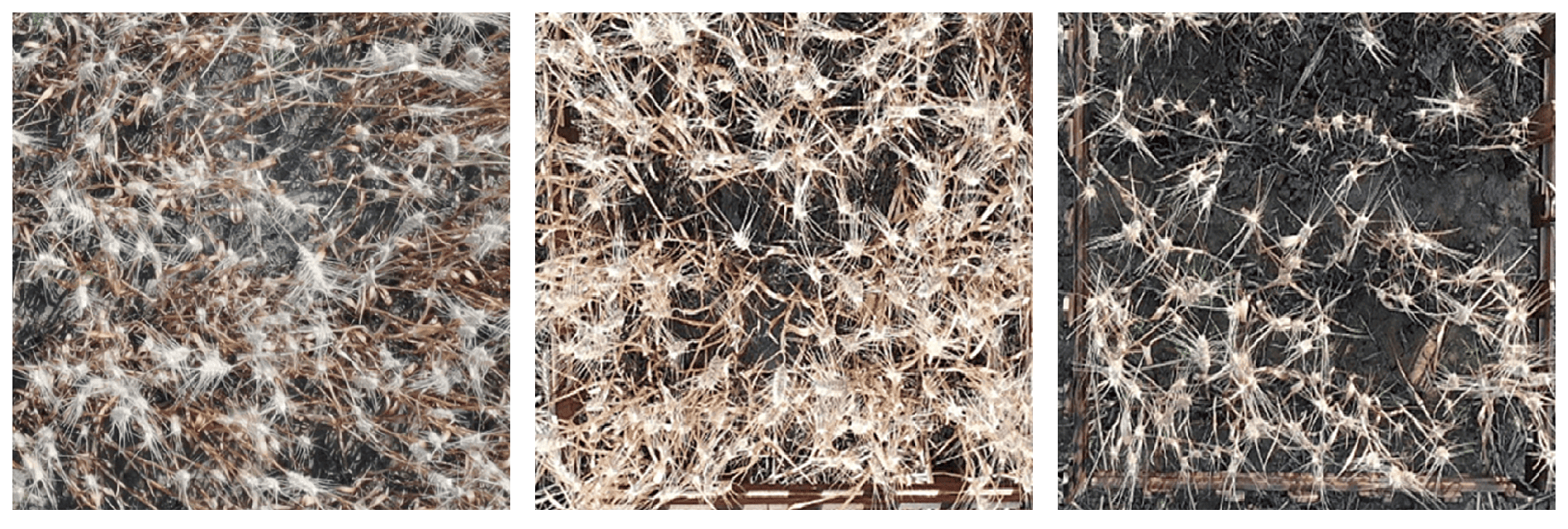}
    \captionsetup{font=scriptsize}
    \caption{Extractions of an UAV image used for estimating wheat crop yields (\cite{peng_combination_2023}).}
    \label{fig:label-de-votre-figure2}
\end{figure}
\noindent This subsection gives, through the cited papers, a global overview of the latest GeoAI techniques for the quantification and estimation of agricultural yields. Different Machine Learning and Deep learning models contribute in the said task using various geospatial data, mainly satellite and UAV images, proving to be very interesting for accurate and reliable prediction of large-scale agricultural productivity.
\subsubsection{Precision irrigation}
Abioye et al. \cite{abioye_review_2020} defined precision irrigation as the optimal use of water resources through the introduction of modern technologies in the agricultural procedures considered. Indeed, the use of prediction performance offered by GeoAI methods in new irrigation technologies is becoming more applicable and efficient. In this context, de Albuquerque et al. \cite{de_albuquerque_dealing_2021} used a Mask-RCNN \cite{he2017mask} with ResNeXt-101 backbone \cite{xie2017aggregated} for automatic detection of Center Pivot Irrigation Systems (CPIS), donating automatic watering systems for large agricultural areas by rotating large pipelines around a central point. The global model takes as input time series of Sentinel-2 satellite images with data augmentation as input, besides, it is conceptualized to take into account different environmental situations, including clouds and seasonal variations. While averaging the simulations results over six tests, the model reached a mAP50 value of 88\%. Raei et al. \cite{raei_deep_2022} presented an U-Net segmentation model \cite{ronneberger2015u} with ResNet30 backbone \cite{he2016deep}, combined with a transfer learning approach for irrigation systems segmentation on a regional scale. 8 600 aerial images from NAIP dataset \cite{NAIP} are used to train, evaluate and test this model. Therefore, final results show a high accuracy of up to 94\% while proving the use of transfer learning and the unbalanced dataset of several irrigation systems categories.
\\
In addition to the automatic detection of irrigation systems, several studies tackled the subject of optimizing water resources dedicated to irrigation. Jalajamony et al. \cite{jalajamony_drone_2023} proposed an intelligent system enabling selective irrigation of the driest areas while optimizing consumption. This system uses several data sources, including RGB images integrated into the Raspberry Pi 4, Thermal InfraRed (TIR) images from the Lepton 3.5 camera, under different levels of hydration, so as to visualize the behavior of the thermal camera in relation to this application. In addition, Global Positioning System (GPS) data from a customized quadricopter is employed to geolocate the areas to be used. Moreover, an AI component is implemented to exploit the embedded results by adjusting the angle of rotation of the flow control valve, this component uses Random Forest (RF) as a base model resulting in an MSE of 0.06, compared to values of 0.14 and 86.5 for KNN and SVM respectively.
\begin{figure}[H]
    \centering
    \includegraphics[width=\linewidth]{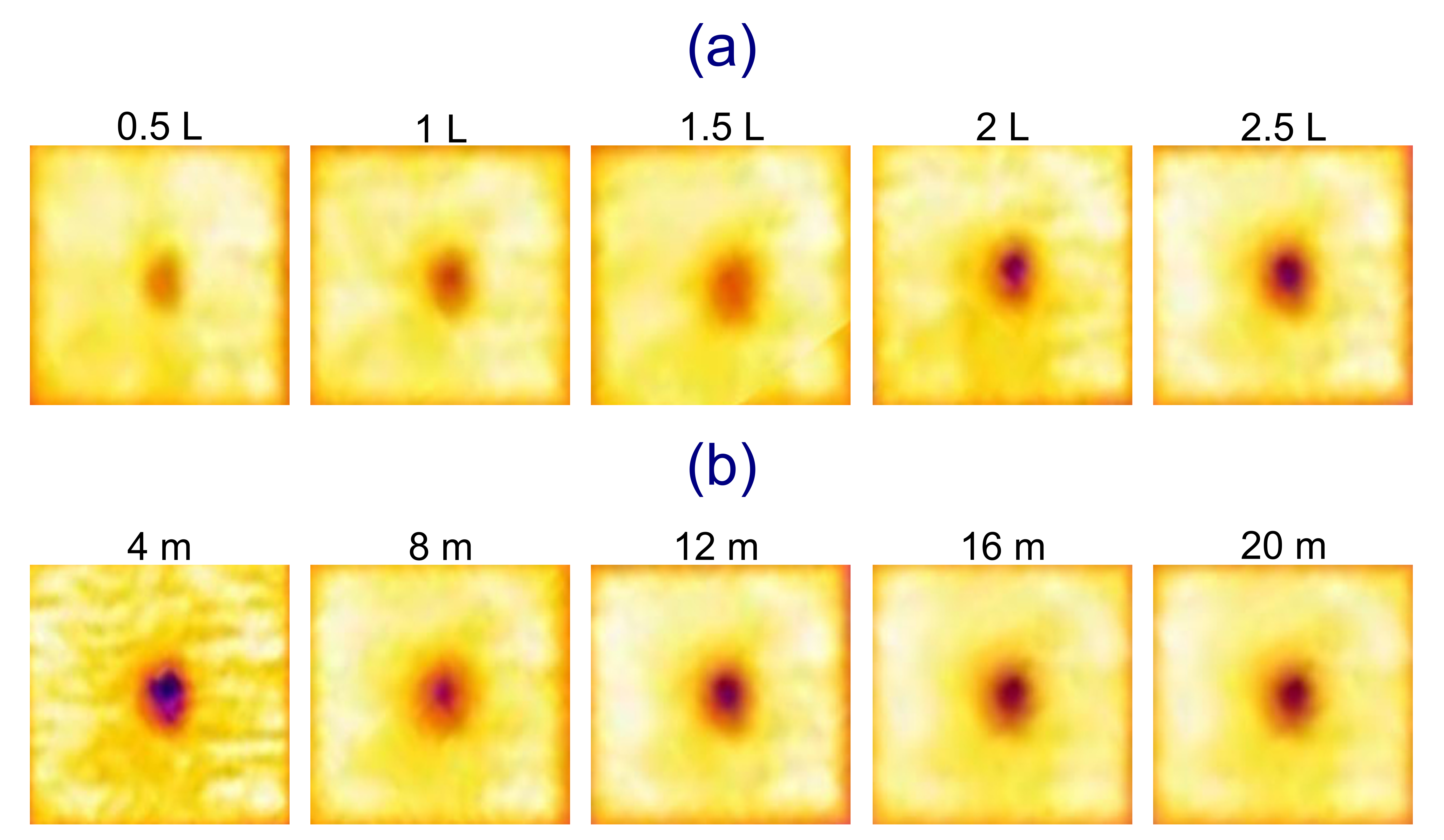}
    \captionsetup{font=scriptsize}
    \caption
    {Examples of thermal images under different levels of hydration (a) and at different altitudes (b) (\cite{jalajamony_drone_2023}).}
    \label{fig:label-de-votre-figure2}
\end{figure}
\noindent Despite the complexity of irrigation procedures, GeoAI has proven to be a high-performance tool for precision irrigation. This includes the detection of irrigation systems from remote sensing images, in addition to the use of several models as a key component for the implementation of concerned systems and procedures. These same models allow for equally effective contributions to the field of urban planning, logistics and transport, enabling sustainable and optimal management of the densest urban areas, and constituting the problem to be explored meticulously in the next subsection.
\begin{table*}[ht]
\centering
\captionsetup{font=scriptsize} 
\caption{Proposed methods for urban planning.}
\label{tab:urban_planning_methods}
\scriptsize 
\renewcommand{\arraystretch}{1.5} 
\begin{tabular}{|m{3.75cm}|m{12cm}|m{1.25cm}|} 
\hline
\multicolumn{1}{|c|}{\textbf{Applications}} & \multicolumn{1}{c|}{\textbf{Methods and algorithms}} & \multicolumn{1}{c|}{\textbf{References}} \\ \hline

\multirow{3}{*}{Urban data extraction} 
& Convolutional Neural Network - Feature Extraction. & \cite{alem_deep_2022} \\ \cline{2-3}
& Multi-Scale Dilated Convolutional Neural Network and Multi-Scale Dilated Convolutional Neural Network. & \cite{chen_multi-modal_2022} \\ \cline{2-3}
& U-Net model. & \cite{zhang_3d_2022} \\ \hline

\multirow{3}{*}{Urban dynamics} 
& Artificial Neural Network. & \cite{tehrani_predicting_2024} \\ \cline{2-3}
& Long Short-Term Memory model. & \cite{hajjar_long_2023} \\ \cline{2-3}
& Pixel to pixel Generative Adversarial Network. & \cite{sun_gan-based_2021} \\ \hline

\multirow{3}{*}{Smart cities} 
& MPSiamese, Few-Shot Siamese Networks and YOLOv5-Pyramid model. & \cite{hu_spatiotemporal_2023} \\ \cline{2-3}
& Long Short-Term Memory model. & \cite{canli_deep_2021} \\ \cline{2-3}
& Random Forest classifier, Xgboost and Faster Region-based CNN. & \cite{ye_urban_2019} \\ \hline

\multirow{6}{*}{Logistics and transport} 
& Support Vector Machine, Feedforward Neural Network, Convolutional Neural Network, Recurrent Neural Network, and Long Short-Term Memory. & \cite{panovski_long_2020} \\ \cline{2-3}
& Hybrid Gated Recurrent Unit - Long Short-Term Memory. & \cite{zafar_traffic_2022} \\ \cline{2-3}
& Incremental Output Decomposition Recurrent Neural Network. & \cite{lu_iodrnn_2024} \\ \cline{2-3}
& Denoising stacked Autoencoder and a Collaborative Learning approach. & \cite{chen_personalized_2020} \\ \cline{2-3}
& SegNet model. & \cite{kearney_maintaining_2020} \\ \cline{2-3}
& Multi-modal deep learning model. & \cite{grandio_multimodal_2023} \\ \hline

\multirow{3}{*}{Green infrastructures and Bio-security} 
& Semi supervised CNN. & \cite{guo_nationwide_2023} \\ \cline{2-3}
& Faster Region based CNN. & \cite{das_geoai_2022} \\ \cline{2-3}
& Convolutional Neural Networks. & \cite{carnegie_airborne_2023} \\ \hline

\end{tabular}
\end{table*}
\subsection{Urban planning, logistics and transport}
By combining AI models with geographic information acquisition techniques, urban planners can better understand and predict possible trends in urban dynamics. This logic is perfectly synchronized with the integration of new technologies into urban management tools, while allowing researchers to implement innovative models that enables the inhabitants of major cities to live "better". In the remainder of this subsection, relevant works combining GeoAI and urban dynamics, smart-cities modelling, logistics and transport along with the management of green urban areas are listed. Table 9 summarizes these research studies.

\subsubsection{Urban data extraction}
Urban classification, i.e. the categorisation of existing input data, such as remotely sensed images or scanned maps, into several classes, enables urban data to be analysed and grouped with a high degree of precision, taking into account their complexity and diversification. Alem and Koumar \cite{alem_deep_2022} used UCM dataset \cite{yang2010bag}, to compare three classification models, comprising a Convolutional Network for Feature Extraction (CNN-FE) along with a Transfer Learning (TL) and a fine tuning models, adjusted from a pre-trained EfficientNet model \cite{tan2019efficientnet}. Performance metrics showed that the fine tuning model performed better in terms of classification, achieving an accuracy of 88\%, and an F1-Score of 89\% for 21 classes. In the same context, Chen et al. \cite{chen_multi-modal_2022} proposed a DL model to detect urban villages in the main cities of Jing-Jin-Ji region from China. The input data consists of Google Earth images merged with Street View data \cite{streetview_reference}. This model has two branches, a neural network based on multi-scale dilated convolutions \cite{yu2015multi} for the Google images and a Recurrent Attention Network based on multi-view learning \cite{blum1998combining} for the Street View data, while achieving an accuracy of 92\%. 
\begin{figure}[H]
    \centering
    \includegraphics[width=0.9\linewidth]{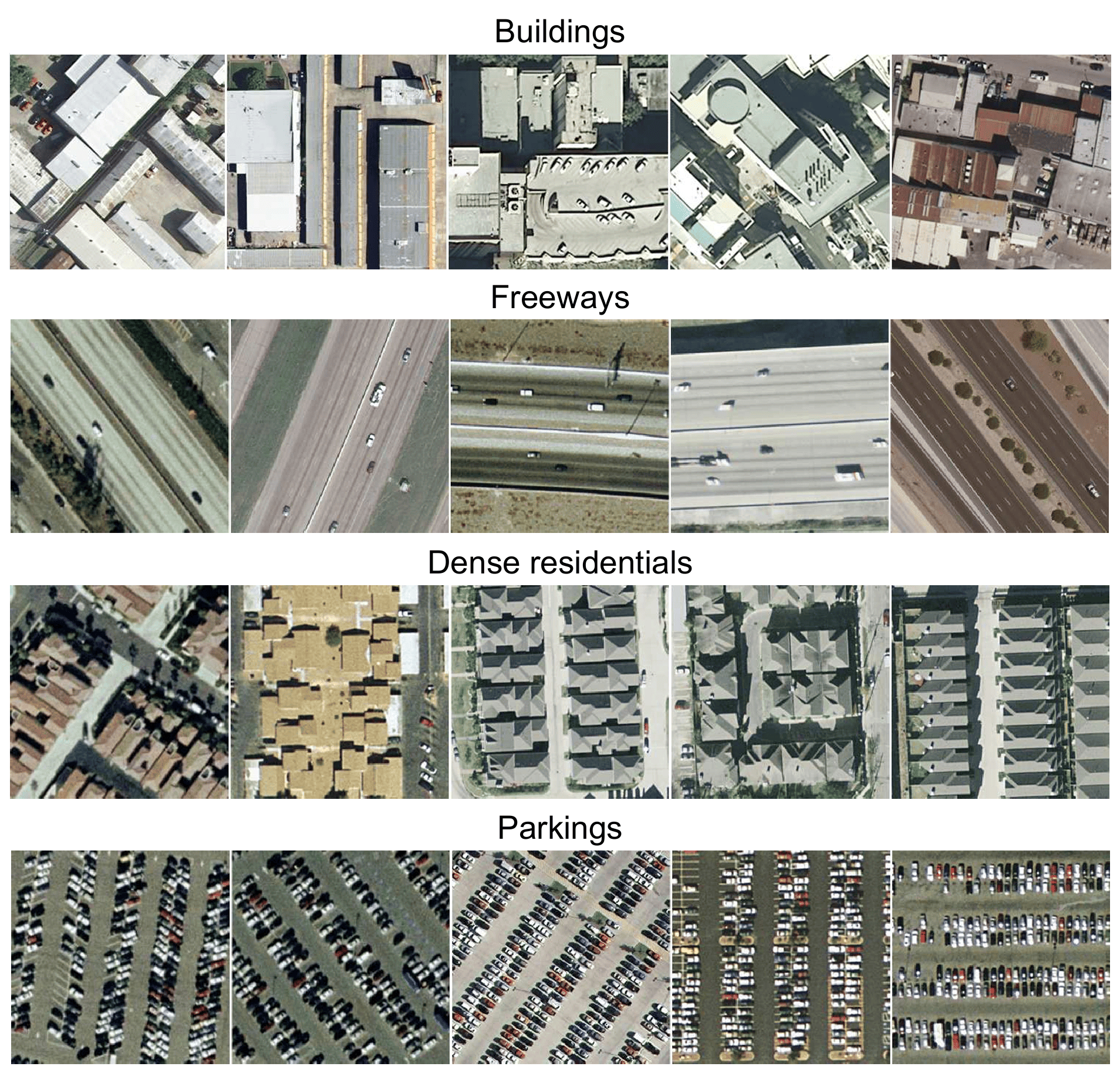}
    \captionsetup{font=scriptsize}
    \caption
    {Examples of UCM dataset (\cite{yang2010bag}).}
    \label{fig:label-de-votre-figure2}
\end{figure}
\noindent As well as classification, a number of studies have focused on creative methods for building extraction from geographic data. For instance, a U-Net architecture \cite{ronneberger2015u} is used by Zhang et al. \cite{zhang_3d_2022} for 3D buildings extraction from LIDAR data and photogrammetric point clouds. This approach extracts the shapes of real colored buildings with high efficiency, obtaining an overall accuracy of 87\%.
\\
Through the census implemented in this subsection, it is clear that the extraction of highly filtered data from multiple datasets represents a topical issue. In addition, it is remarked that multiple categories of geospatial data are used, including aerial, satellite imagery and LIDAR point clouds, while noting the particular contribution of resolved data in classification and segmentation tasks, given the inherent need for much finer data extraction.

\subsubsection{Urban dynamics}
Thanks to their ability to process complex data "intelligently", new ML and DL techniques are able to completely change the classical understanding of urban dynamics, enabling the anticipation of the main needs on this field. Tehrani et al. \cite{tehrani_predicting_2024} proposed a novel ANN to predict the energy of solat electromagnetic radiations in urban areas, using a fruitful combination of different types of data, precisely three-dimensional coordinates of constructions, average building heights, inhabited and unoccupied areas coupled with azimuth angle. This study constitutes a creative work to integrate solar energy into future and sustainable urban planning. Simulations results showed that this model achieves an MSE of 0.01 and an R² of 85\%. Hajjar et al. \cite{hajjar_long_2023} proposed an LSTM to process Land-Use Change (LUC) data in time and space. Using 3 raster maps with dates of 2000, 2010 and 2020, the tests have showed significant effectiveness for the classification task in question.
\\
In a relatively different context, prediction of urban development is a key element in the preparation of urban master plans, as they indicate recommendations to be followed in the future. Sun et al. \cite{sun_gan-based_2021} developed a pix2pix GAN \cite{isola2017image} for Land-Use and Land-Cover Change (LUCC) prediction by integrating vectorial planning indicators, as well as land-use probability maps from Landsat 5 and 8 satellite images. Model in focus performed well, achieving a pixel-by-pixel prediction accuracy of 85\%.
\\
To conclude, various models are used to explore the different aspects of urban dynamics, including the distribution of energy over urban environments, the analysis of spatio-temporal changes, whether in relation to land cover, linking to the occupation of the urban surface in focus, or in relation to land use. In short, GeoAI techniques are a key tool for monitoring and analyzing urban changes, and for launching a new era of intelligent cities, a theme analyzing in depth in the following subsection.

\subsubsection{Smart cities}
The integration of cutting-edge AI models in smart cities is increasingly indispensable to automate urban infrastructure. Indeed, a multitude of research projects are underway to combine geospatial data with AI techniques In this regard. Hu et al. \cite{hu_spatiotemporal_2023} presented an intelligent spatio-temporal framework based on GeoAI. The proposed framework integrates multi-video geo-referenced approach along with the virtual city based on Augmented Reality (AR) in urban digital twins. Siamese \cite{bromley1993signature} video tracking models are utilized, in particular MPSiam \cite{li2021mpsiam} and Four-Stream Siamese Network (FSSiamese) \cite{wang2021multiple}, together with YOLOv5-Pyramid architecture \cite{yang2020face} for small object detection. The presented models demonstrated significant performance while obtaining a mean offset error for video georeferencing of 0.14m. In a different context, Canli and Toclu \cite{canli_deep_2021} have developed a mobile application based on deep and cloud learning to refine parking searches in the city of Istanbul. The authors used an LSTM fed by the ISPARK dataset collected by Istanbul Metropolitan Municipality is used in this context. Test results showed the high efficiency of LSTM model in predicting car park occupancy rates, compared with RF, SVM and Autoregressive Integrated Moving Average (ARIMA) models. The parking searches model achieves an accuracy of 99\% and an RMSE of 1.59.
\\
Ye et al. \cite{ye_urban_2019} introduced the factor of urban commerce in smart cities by analyzing Street View annotated images, illustrated in Figure 30, and OpenStreetMap road network. Random Forest and Xgboost classifiers are applied to the images and a Faster Region-based CNN (Faster-RCNN) model \cite{ren2015faster} is employed to detect commercial signs. Results showed satisfactory accuracy in different scenarios, while obtaining an average precision (AP) of 88\% for the detection task and a mAP of 85\% for Random Forest classification.
\begin{figure}[H]
    \centering
    \includegraphics[width=1\linewidth]{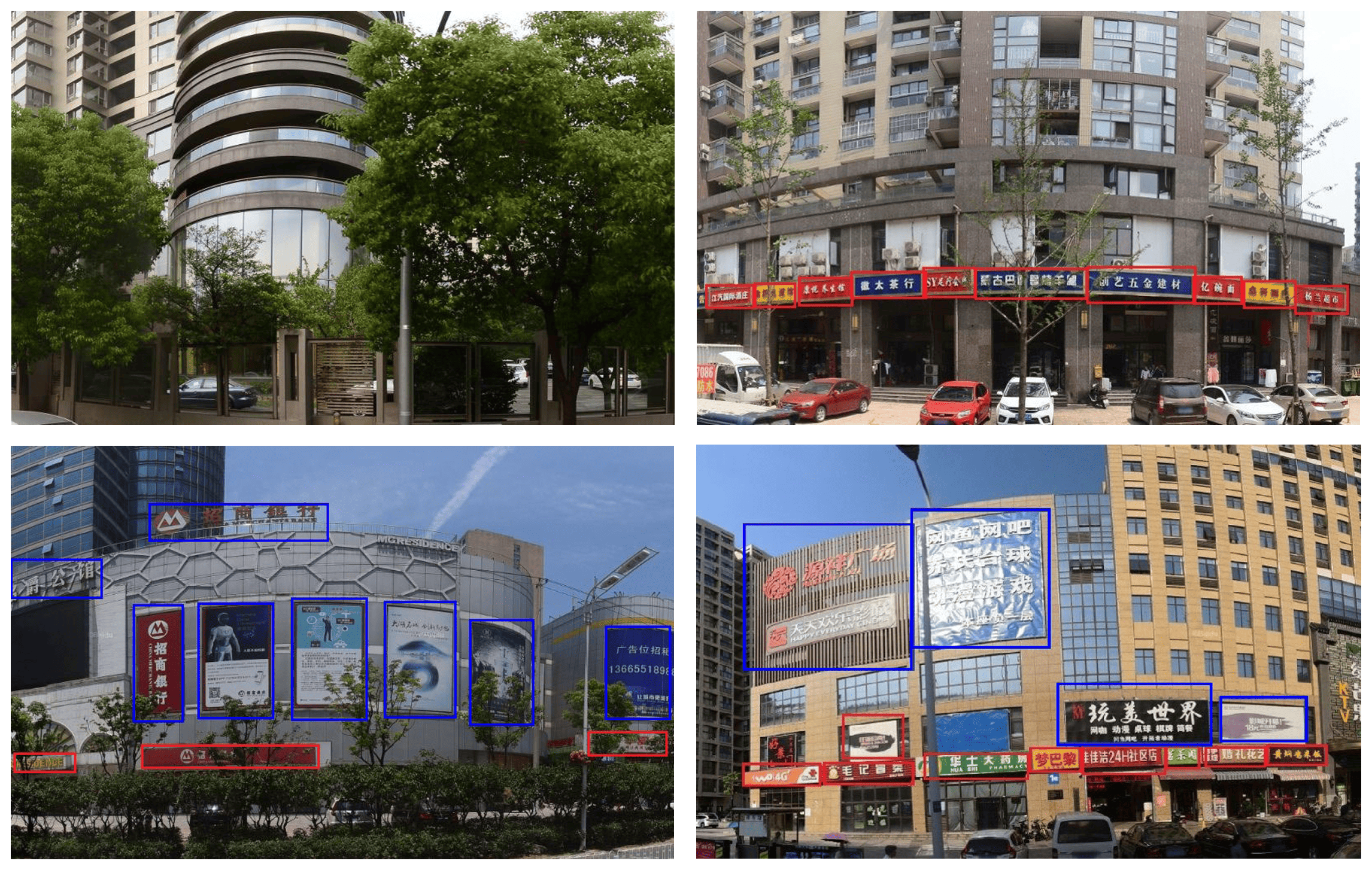}
    \captionsetup{font=scriptsize}
    \caption
    {Examples of Street View images used in \cite{ye_urban_2019}. The blue boxes represent road signs and the red ones shop signs.}
    \label{fig:label-de-votre-figure2}
\end{figure}
\noindent In conclusion, this subsection provides a complete overview of the contribution of GeoAI to the development of smart cities. The synergy between AI models and geospatial data makes it possible to exploit these data in a multitude of task, such as detecting key objects, exploring the availability and status of the emphasized services, and improving the attractiveness of commercial offers.
\subsubsection{Logistics and transport}
In addition to urban classification and the identification of the concerned dynamics, GeoAI provides productive solutions for optimizing and forecasting traffic, calculating routes and managing various transportation systems. Several AI models using geographic records are employed to improve the safety of road networks, to ensure the sustainability of transportation modes and to manage logistics operations. In this settings, Panovski and Zaharia \cite{panovski_long_2020} have proposed a method for predicting bus arrival times in short term, i.e. from a few minutes to two hours, and in long term, i.e. throughout the day, using a traffic density matrix (TDM). As part of a comparative study, numerous models are used including Support Vector Regression (SVR), CNN, Feedforward Fully Connected Neural Network (FNN), RNN and LSTM to improve the accuracy of predictions, giving a Mean Absolute Error (MAE) values of 71 s, 48.7 s, 46.9 s and 45.3 s, 44.6 s for the SVR, CNN, RNN, LSTM and FNN respectively. Likewise,  Zafar et al. \cite{zafar_traffic_2022} have developed a hybrid Gated Recurrent Unit LSTM (GRU-LSTM) for predicting traffic in smart cities. The input data is various OSM data and GIS maps for road segmentation. The model is 95\% accurate. Lu et al. \cite{lu_iodrnn_2024} proposed an Incremental Output Decomposition Recurrent Neural Network (IODRNN), referring to a specific RNN based on incremental learning, for traffic prediction in Intelligent Transport Systems (ITS), attempting to overcome the shortcoming of time delay in predictions. GNSS data is employed as input to the model, with an MAE and an RMSE of 16\% and 17\% respectively.
\\
In addition to predicting traffic density, route calculation aims to reduce the number of trips and the consistency of congestion, constituting a field of attack for GeoAI tasks. Chen et al. \cite{chen_personalized_2020} proposed a DL model for recommending optimized itineraries for tourists, while integrating points Of interest (POI) location with user history. The concerned model is composed of a Denoising Stacked AutoEncoder (SDAE) \cite{vincent2010stacked} and a collaborative learning model \cite{laal2012benefits} to predict user interests and visit durations. As shown in Figure 32, the SDAE architecture consists of an encoder that learns the robust representations of noisy textual descriptions of POIs, while reducing the associated dimensuality. The decoder then reconstructs the original dimension, while correcting the noise introduced. The model results in an average F1-score of 48\%, obtained over 8 different cities, namely Budapest, Delhi, Edinburgh, Glasgow, Osaka, Perth, Toronto and Vienna. The performance obtained is considered very favourable given the complexity of the task in question.
\begin{figure}[H]
    \centering
    \includegraphics[width=1\linewidth]{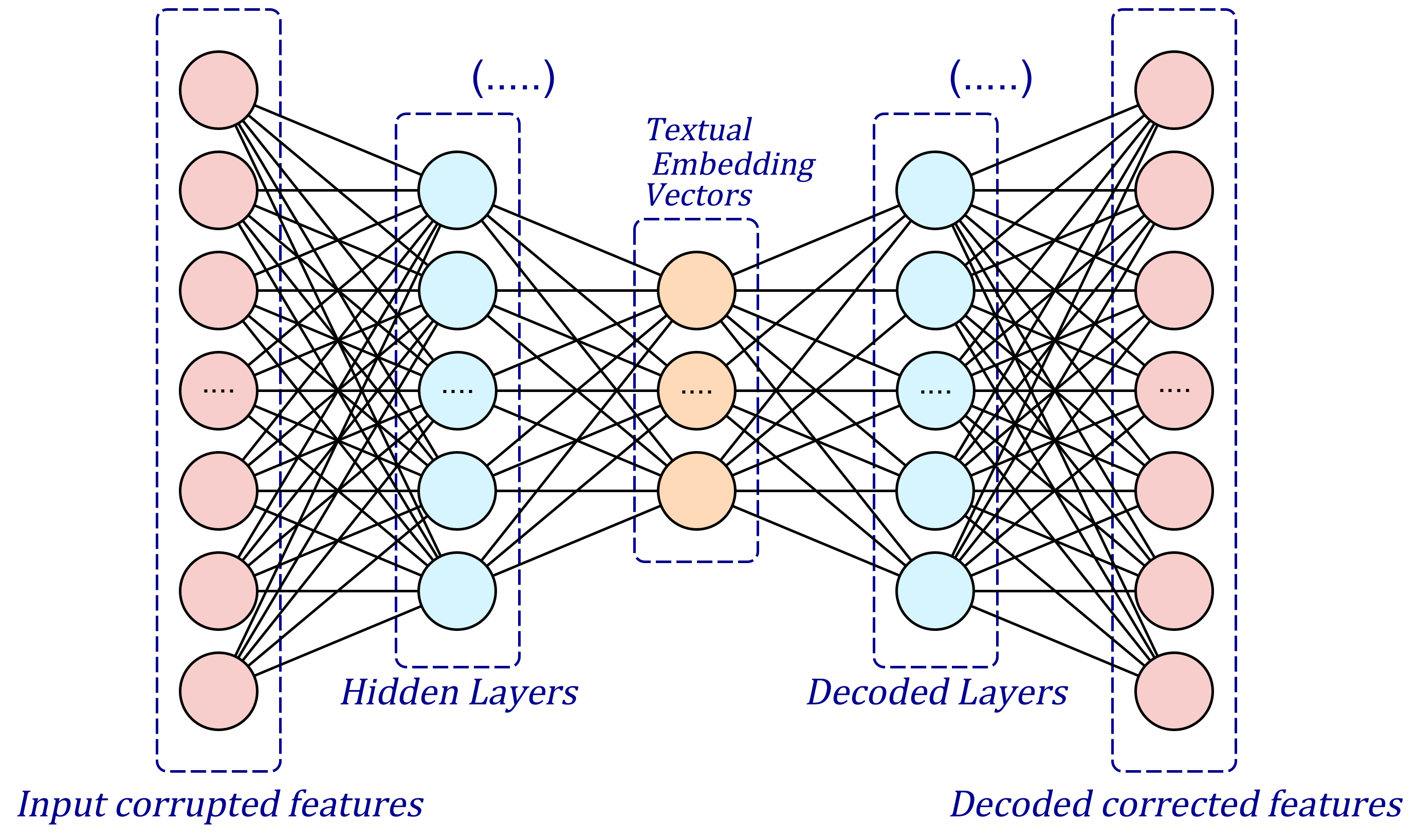}
    \captionsetup{font=scriptsize}
    \caption
    {SDAE architecture for route calculation, proposed in \cite{chen_personalized_2020}.}
    \label{fig:label-de-votre-figure2}
\end{figure}
\noindent The inventory of transportation infrastructure seems necessary for quickly identifying areas in need of maintenance, and to quickly update road conditions. Rúa et al. \cite{rua_automatic_2023} estimated the road areas likely to be affected by rockfalls, while analyzing the slopes of roads. The authors first detect the slope through a cloud of LIDAR points, estimate the areas susceptible to falls, and then simulate the event using RockGIS software. Indeed, the F1-Score of the Proposed model is 96\%. In addition, Kearney et al. \cite{kearney_maintaining_2020} have presented a segmentation model based on SegNet \cite{badrinarayanan2017segnet}, an encoder decoder architecture based on CNN, using rapidEye satellite images with the aim of updating rural road networks for forest landscapes. The test phase showed high capabilities, with a Recall of 89\% and accuracy of 87\%. In another context, Grandio et al. \cite{grandio_multimodal_2023} presented a multimodal DL model for panoptic segmentation of railway features, including signposts, rails, information boards, etc. A point cloud from two LIDAR sensors is utilized to generate railway infrastructures. As illustrated in Figure 33, a U-Net architecture \cite{ronneberger2015u} is applied to rasterize the said point cloud. Subsequently, a Pointnet++ \cite{qi2017pointnet++} model is exploited for a detailed semantic segregation. This workflow have showed remarkable performance, achieving a precision of 68\%, a recall of 86\% and an F1 score of 74\%.
\begin{figure}[H]
    \centering
    \includegraphics[width=0.7\linewidth]{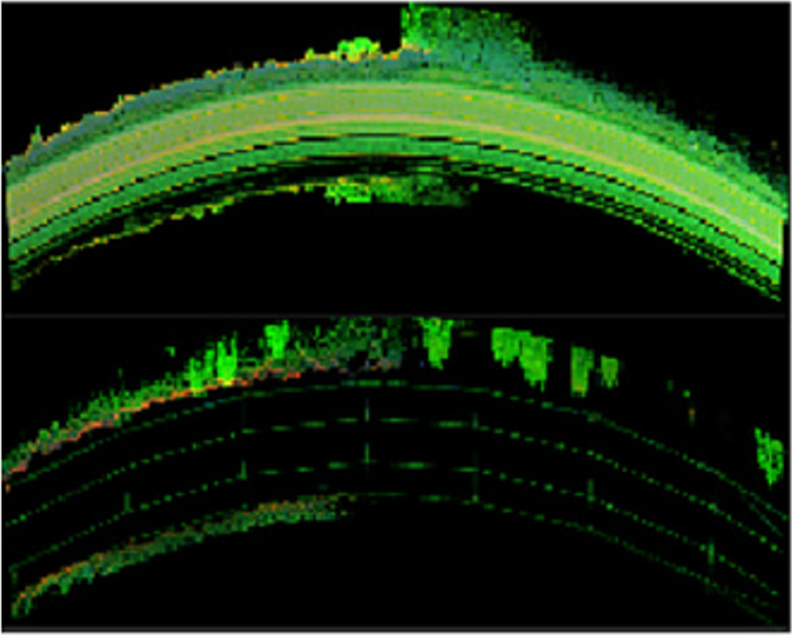}
    \captionsetup{font=scriptsize}
    \caption
    {Sample of rasterized LIDAR data used for railway features segmentation (\cite{grandio_multimodal_2023}).}
    \label{fig:label-de-votre-figure2}
\end{figure}
\noindent Through the inventory presented above, various researches have exploited GeoAI models for traffic management, route optimisation and inventory of existing infrastructure, enabling decision-makers to improve the accuracy of their estimates and to implement high-calibre strategic planning. In the next subsection, a rather interesting area is explored, in particular that of green infrastructure and biodiversity and their impact on a healthy urban environment.
\subsubsection{Green Infrastructure and Bio security}
Optimal management of green spaces in urban areas is crucial to enhance the urban landscape and to improve significantly the quality of life of city residents. Many research studies have contributed to this perspective. For instance, Guo et al. \cite{guo_nationwide_2023} proposed a semi-supervised CNN based on Deeplabv3+ architecture \cite{chen2018encoder} for the Urban Tree Cover (UTC) mapping. Images from Digital Globe's WorldView-2 and 3, GeoEye-1, Planet Labs' SkySat and Airbus' Pleiades satellites are adopted to train the model. Evaluation phase showed encouraging results, with an overall accuracy of 95\%. Furthermore, an individual tree inventory (ITI) model is developed by Das et al. \cite{das_geoai_2022} to quantify the contribution of tree cooling on urban heat. Aerial high resolution and Landsat-8 Land Surface Temperature (LST) images together with LIDAR data are utilized to describe the Canopy Height Model (CHM) with a resolution of 1m. The proposed model architecture is based on a Faster Region based CNN (R-CNN) architecture \cite{ren2015faster} coupled with Inception ResNet V2 \cite{szegedy2017inception} in order to identify seven individual tree canopy species. Additionally, Pearson correlation \cite{cohen2009pearson}, Linear Regression (LR) and Geographically Weighted Regression (GWR) \cite{brunsdon1998geographically} are deployed to identify the relationship between tree characteristics, species and local temperature. The resulting identification model performed very well, achieving a mean Average Precision (mAP) of 85\% and an average F1-Score of 80\%.
\\
In addition to inventorying green spaces in urban environments, Carnegie et al. \cite{carnegie_airborne_2023} have developed a tree classification model for forest pest biosecurity monitoring. The model uses two CNNs for tree classification (M1) and instance segmentation (M2), with multispectral aerial imagery as input. Results of this comparison favoured Model M2 over M1 in terms of F1-Score metric, attaining an average value of 79\%, in comparison with a value of 58\% for M1.
\\
In summary, various ML and DL models are used to optimize the location of green infrastructure and to guarantee bio-diversity in urban environments, thereby ensuring the ecological balance of the environment in question, improving the quality of citizens life and accurately monitoring urban trends. Cited Research work includes urban vegetation mapping, correlation measurements between green infrastructure and environmental parameters, as well as monitoring biodiversity within the most densest cities. The control of green infrastructures and biodiversity leads to focus on another topic, that of environmental monitoring, comprising climate change, atmospheric chemistry and environmental impacts.
\subsection{Environmental monitoring}
Complex data from a variety of sources such as satellite imagery, environmental sensors and climate variables, can be used to monitor critical aspects of the environment, such as climate change measurements, air quality variables and environmental impacts of projects, activities and events. In fact, Figure 34 shows an example of environmental data exploited in this context, more precisely the Aerosol Optical Depth (AOD), referring to the portion of light attenuated by aerosol particles in the atmosphere \cite{wei2020satellite}, measured by OMI instrument and downloadable via EOSDIS datasets \cite{musial2024overview}. Conversely, Table 10 summarizes AI methods explored in this subsection.
\begin{figure}[H]
    \centering
    \includegraphics[width=1\linewidth]{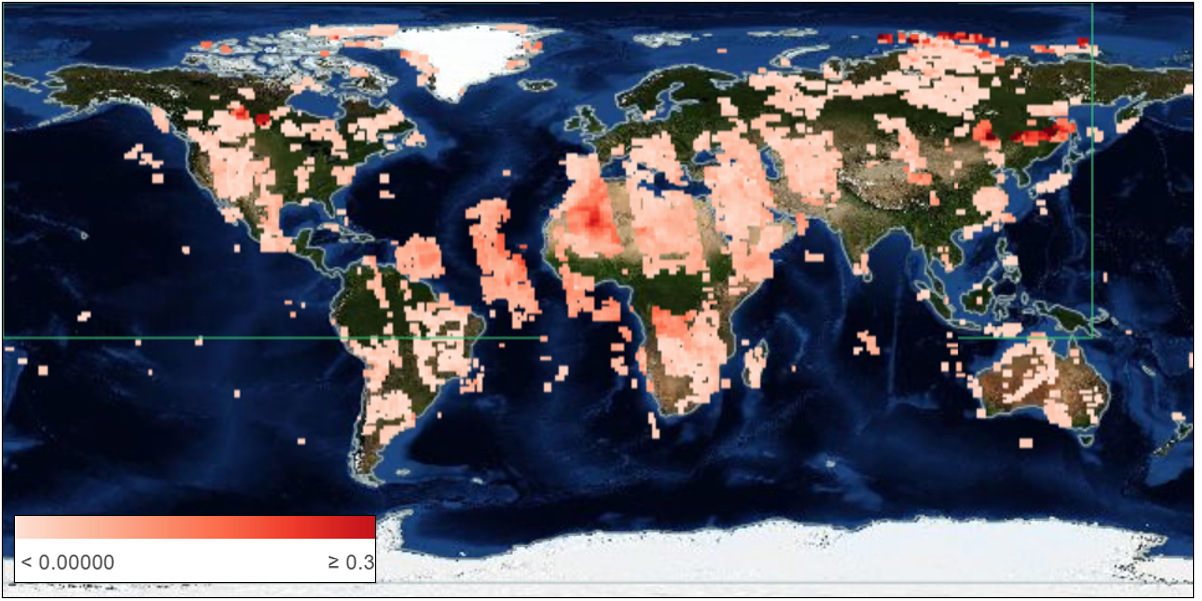}
    \captionsetup{font=scriptsize}
    \caption{OMI Near UV Aerosol Optical Depth dated 07/24/2024.}
    \label{fig:label-de-votre-figure3}
\end{figure}
\begin{table*}[ht]
\centering
\captionsetup{font=scriptsize} 
\caption{Models and proposed methods used for environment monitoring.}
\label{tab:environment_monitoring_methods}
\scriptsize 
\renewcommand{\arraystretch}{1.5} 
\begin{tabular}{|m{2.25cm}|m{10.5cm}|m{1.5cm}|} 
\hline
\multicolumn{1}{|c|}{\textbf{Applications}} & \multicolumn{1}{c|}{\textbf{Methods and algorithms}} & \multicolumn{1}{c|}{\textbf{References}} \\ \hline

\multirow{3}{*}{Climate change} 
& Deep Neural Network and Recursive Feature Elimination method. & \cite{lou_deep_2023} \\ \cline{2-3}
& Residual Channel Attention Network. & \cite{wang_applying_2023} \\ \cline{2-3}
& Rule-based ML, Convolutional Neural Network and Random Forest. & \cite{xin_evaluations_2020} \\ \hline

\multirow{5}{*}{Atmospheric chemistry} 
& Deep Convolutional Autoencoder. & \cite{opio_modeling_2023} \\ \cline{2-3}
& XGBoost model. & \cite{kim_importance_2021} \\ \cline{2-3}
& Random Forest and K-Means clustering algorithm. & \cite{dou_estimates_2021} \\ \cline{2-3}
& Deep Neural Network. & \cite{rowley_predicting_2023} \\ \cline{2-3}
& Long Short-Term Memory. & \cite{seng_spatiotemporal_2021} \\ \hline

\multirow{4}{*}{Environmental impacts} 
& Artificial Neural Network. & \cite{alqadhi_applying_2024} \\ \cline{2-3}
& Multiscale Geographically Weighted Regression, Random Forest along with SHapley Additive exPlanation model. & \cite{gianquintieri_implementation_2024} \\ \cline{2-3}
& Logistic Regression, XGBoost, Light Gradient Boosting Machine, Multi-Layer Perceptron, Long Short-Term Memory, Convolutional Long Short-Term Memory and transformers. & \cite{shevchenko_climate_2024} \\ \cline{2-3}
& DeeplabV3+ architecture. & \cite{cao_ecological_2023} \\ \hline

\end{tabular}
\end{table*}
\subsubsection{Climate change}

Faced with the major challenges of climate change and its effects on the planet, several research projects have addressed this issue by proposing novel methods and models for monitoring environmental events. Lou et al. \cite{lou_deep_2023} presented a Deep Neural network (DNN) using a recursive feature elimination (RFE) algorithm \cite{guyon2002gene}. The aim of the proposed model is to explore vegetation changes using Landsat 5, 8, Sentinel-1 satellites and MODIS images, from 1988 to 2018. Over the explored period, the model's performance shows an overall accuracy of 83\%. Wang et al. \cite{wang_applying_2023} used a Residual Channel Attention Network (RCAN) \cite{zhang2018image}, it is a model improving image resolution while using attention mechanisms and image fusion, combining details from several images. The aim of this work is to provide a more accurate representation of the vegetation cover from Danjang river basin. Landsat 4, 8 and MODIS images are used to train the presented model. It is worth mentioning that the NDVI index \cite{carlson1997relation} is used for the physical representation of this vegetation cover. Then, the resulting refined representation enabled better reconstruction of the NDVI index for more accurate exploration of change. In the same context, Xin et al. \cite{xin_evaluations_2020} evaluated several methods, in particular neural network and random forest, as well as six rule based methods to retrieve vegetation phenology using MODIS images and USA national phenology network data \cite{betancourt2005implementing}. The results show a big difference between the different methods in terms of efficiency, while highlighting the performance of RF model, with an MAE of 15.16 days. In addition to NDVI, Leaf Area Index (LAI) \cite{zheng2009retrieving} is a powerful ecological indicator for measuring the interactions of flora with the environment, it represents the ratio of the total leaf area of vegetation to the ground surface. Ma and Liang \cite{ma_development_2022} presented an improved leaf area index at 250 m resolution. For this, the authors used a bidirectional long-term memory (Bi-LSTM) \cite{schuster1997bidirectional} as a base model. MODIS surface reflectance data is exploited to form a spatio-temporal series as a training dataset, while producing 79 LAI maps to test the resulting model.
\\
Despite the difficulties associated with managing the problem of climate change, several AI models using geospatial data are implemented to monitor, analyze and predict the causes and consequences of this phenomenon. Accurate and comprehensive analysis is carried out to manage the resulting effects and to plan appropriate coping strategies.
\subsubsection{Atmospheric chemistry}
Atmospheric chemistry is a sub-discipline of atmospheric science quantifying the composition of chemical elements in the atmosphere. Besides, artificial intelligence, in particular GeoAI, guarantees accurate prediction of atmospheric chemical components, through training based on historical data. Opio et al. \cite{opio_modeling_2023} have developed a Weather Research and Forecasting model for the atmospherical Chimestry (WRF-Chem) based principally on OMI images. In addition, the proposed architecture has been enhanced by a deep convolutional autoencoder for WRF (WRF-DCA) to model the atmospheric dispersion of sulphur dioxide (SO2). This model showed a strong performance with an RMSE of $1.5 \times 10^{16} \, \text{molecules/cm}^2$. Kim et al. \cite{kim_importance_2021} used an XGBoost model to estimate the hourly Nitrogen Dioxide gaz (NO2) concentrations from Sentinel 5P images, along with meteorological data and ground sensor measurements. Results showing the importance of accurate near-surface NO2 mapping, particularly during the COVID-19 containment period. For the same reason, Dou et al. \cite{dou_estimates_2021} have combined an RF model with K-Means algorithm to predict NO2 concentrations in China. Using data from Ozone Monitoring Instrument (OMI), combined with socio-economic factors and anthropogenic emission inventories, the model provided detailed estimates of NO2 near the ground, subsequently used for environmental quality and epidemic management in China.
\\
Moreover, air quality qualification is a key element in governing and sustainable environmental management. Indeed, Rowley and Karakuş \cite{rowley_predicting_2023} propose air quality network (AQNet), a deep neural network trained on multispectral images from Sentinel-2 satellite, merged with NO2 concentrations from Sentinel-5P satellite data. The goal of this research is to form an air quality index from NO2 concentrations, ozone gases (O3) and particulate matter (PM10), i.e. airborne particles with an aerodynamic diameter of less than 10 µm. In addition, Seng et al. \cite{seng_spatiotemporal_2021} exploited an LSTM model to predict concentrations of PM2.5, referring also to airborne particles with an aerodynamic diameter of less than 2.5µm, Sulfur Dioxide (SO2), NO2, O3, and Carbon Monoxide (CO). The input data are ground measurement stations constituting a large spatio-temporal series. The test phase showed outstanding efficiency with an RMSE of 12.48 mg/m³ and an R² of 91\%.
\\
Besides the area of air quality, quantification of aerosols, i.e. the constituents of fine particles suspended in the atmosphere, is essential for studying aerosols impact on the radiation balance, and for understanding the dispersion of pollutants. With this in mind, Sun et al. \cite{sun_aerosol_2021} dealed with the aerosol absorption properties using UltraViolet Aerosol Index (UVAI) \cite{torres1998derivation}, an index reflecting the absorption capacity of aerosols in the ultra-violet wavelengths. The authors have implemented a DNN using ozone monitoring instrument (OMI) data for training and ground-based aerosol robotic network (AERONET) \cite{holben1998aeronet} observations for test, a representative map of which is shown in Figure 35, resulting in an RMSE of 0.005.
\begin{figure}[H]
    \centering
    \includegraphics[width=0.9\linewidth]{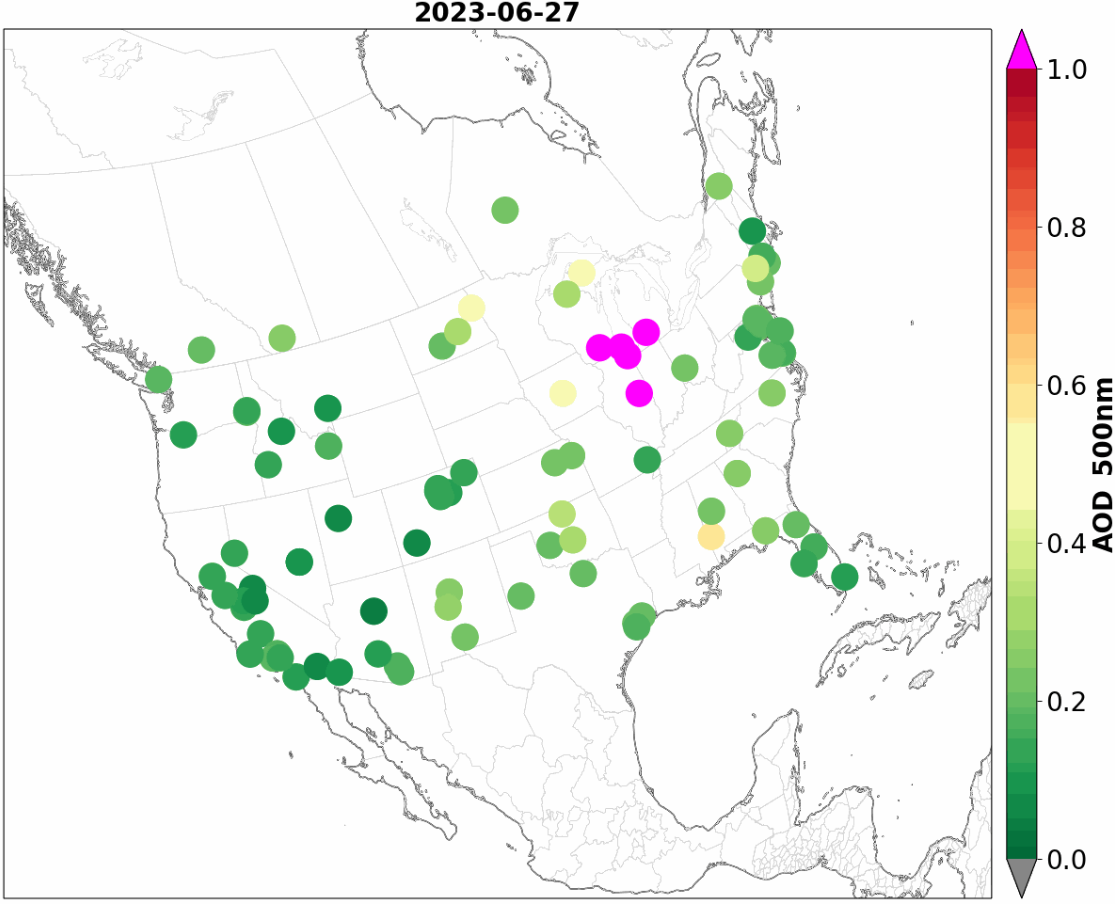}
    \captionsetup{font=scriptsize}
    \caption{Example of United States AERONET data dated 06/27/2023 (\cite{aeronet}).}
    \label{fig:label-de-votre-figure3}
\end{figure}
\noindent Finally, the methods and models presented allow for accurate models and predict the chemical composition of the atmosphere, providing a better understanding of atmospheric dynamics and processes, assessing air quality and forecasting climatic phenomena. 
\subsubsection{Environmental impacts}
Given the accuracy and reliability of its techniques, GeoAI is proving extremely valuable in assessing the interactions between the environment and human activity. Indeed, two concepts are distinguished, notably Environmental Impact Assessments (EIA), as well as the study of environmental effects on project sites, infrastructures and agricultural land. In this regard, the use of GeoAI is very useful for monitoring environmental phenomena, measuring the impact of human activities on the integrity of ecosystems, on physical factors such as air and water quality, on social services and economic values. 
\\
According to Glasson et al. \cite{glasson2013introduction}, EIA is a systematic approach aiming to analyze and predict the effects of development actions on the environment. Several AI models contributed to this examination, in fact, an Artificial Neural Network (ANN) is developed by Al Qadhi et al. \cite{alqadhi_applying_2024} to analyse the impact of urbanisation on ecosystems in Abha region of Saudi Arabia. The ANN in question is trained on Land use and Land Cover (LULC) maps using Landsat 4, 5 and 8 images over the years 1990, 2000 and 2010. The proposed ANN is able to obtain an accuracy of 94\%, using alongside biophysical, biochemical and biological indicators to develop the Remotely Sensed Urban Surface Ecological Index (RSUSEI), a qualitative index of urban ecological condition \cite{firozjaei2020remotely}. In the same context as Environmental Impact Assessment, Gianquintieri et al. \cite{gianquintieri_implementation_2024} proposed a GeoAI model with a multi-block architecture in order to assess the impact of agricultural activity on the distribution of PM2.5. Data used are PM2.5 concentrations from Copernicus database and land use maps of Lombardy region in Italy. The blocks in question are Spearman rank-based correlation \cite{spearman1961proof} for feature extraction, multiscale geographically weighted regression (MGWR) \cite{fotheringham2017multiscale, oshan2019mgwr} for spatial data enhancement, Random Forest (RF) for identification of areas with high PM2.5 concentration, and a SHapley Additive exPlanation (SHAP) method \cite{lundberg2017unified} to compare the impact of several attributes such as agricultural land use, rate of urbanisation and presence of water resources. Results of SHAP model showed that natural areas followed by agricultural land with maize and cereals have a strong impact on the concentration of PM 2.5. 
\\
To perform a so-called inverse task, i.e. measuring the environmental effects on agricultural activities, Shevchenko et al. \cite{shevchenko_climate_2024} presented a meta-classifier based on several ML models, specifically Logistic Regression (LR), CatBoost \cite{prokhorenkova2018catboost}, XGBoost, LightGBM \cite{ke2017lightgbm}, along with various DL models, particularly MLP, LSTM, ConvLSTM \cite{shi2015convolutional} and a Transformer. The goal of this study is to measure the impact of climate change on the suitability of agricultural lands in Europe and Asia. The authors use agricultural classification products from Global Food Security-support Analysis Data at 1 km resolution (GFSAD1km) \cite{thenkabail2016nasa}, together with climatic and morphological data to train and evaluate the model, resulting in an overall accuracy of 86\%.
\\
Cao et al. \cite{cao_ecological_2023} assessed the ecological safety of ice and snow tourist destinations, particularly in the face of environmental threats and climate change. To do this, a DeeplabV3+ classification model \cite{chen2018encoder} is introduced using Landsat satellite images. As shown in Figure 36, the presented model consists of an encoder to extract image characteristics through successive convolutions and pooling, along with a decoder refining the predictions made by applying convolutions. Then, a factor 4 upsampling and concatenation with the features produced by the decoder are applied, followed by resampling to reduce information loss. The model's predictions achieved an overall accuracy of 91\%, demonstrating remarkable efficiency compared with traditional methods such as maximum likelihood \cite{fisher1922mathematical} and minimum distance method \cite{wolfowitz1957minimum}.
\\ 
The models presented in this subsection reflect the state of exploration of environmental impact assessment, including both the identification of the impacts of urbanization on ecosystems \cite{alqadhi_applying_2024}, the exploration of the effects of agriculture on environmental parameters \cite{gianquintieri_implementation_2024}, and the measurement of environmental effects on agricultural activity \cite{shevchenko_climate_2024}. 
\\
The explored examples provide a comprehensive overview of GeoAI's contribution to the aforementioned theme, not only in terms of applications, but also with regard to the integration of AI models at the heart of the processes in question. The presented models have proven their high applicability in environmental monitoring, including the use of geospatial data of all types to monitor various environmental phenomena, to model environmental impacts and to preserve ecosystems, enabling the engaged parties to take the necessary measures to protect the environment. Following the same logic, the next subsection explores the synergy between GeoAI and another cutting-edge field, that of water resource management, enabling to respond differently to current challenges in this area.
\subsection{Water resources management and precipitation forecasting}
Given the vital importance of water resources, a considerable set of research is carried out using innovative GeoAI techniques and methods. This research is aimed at sustainable management of water resources and more targeted interpretation of the water-related phenomena, regarding hydrological modelling, water quality, groundwater management and precipitation. Implementing these techniques enables informed decision-making in the face of today's challenges, especially quantifying demand, optimizing use and proactively protecting against pollution. Table 11 summarizes the methods and algorithms used.
\begin{figure*}[ht]
    \centering
    \includegraphics[width=0.75\textwidth]{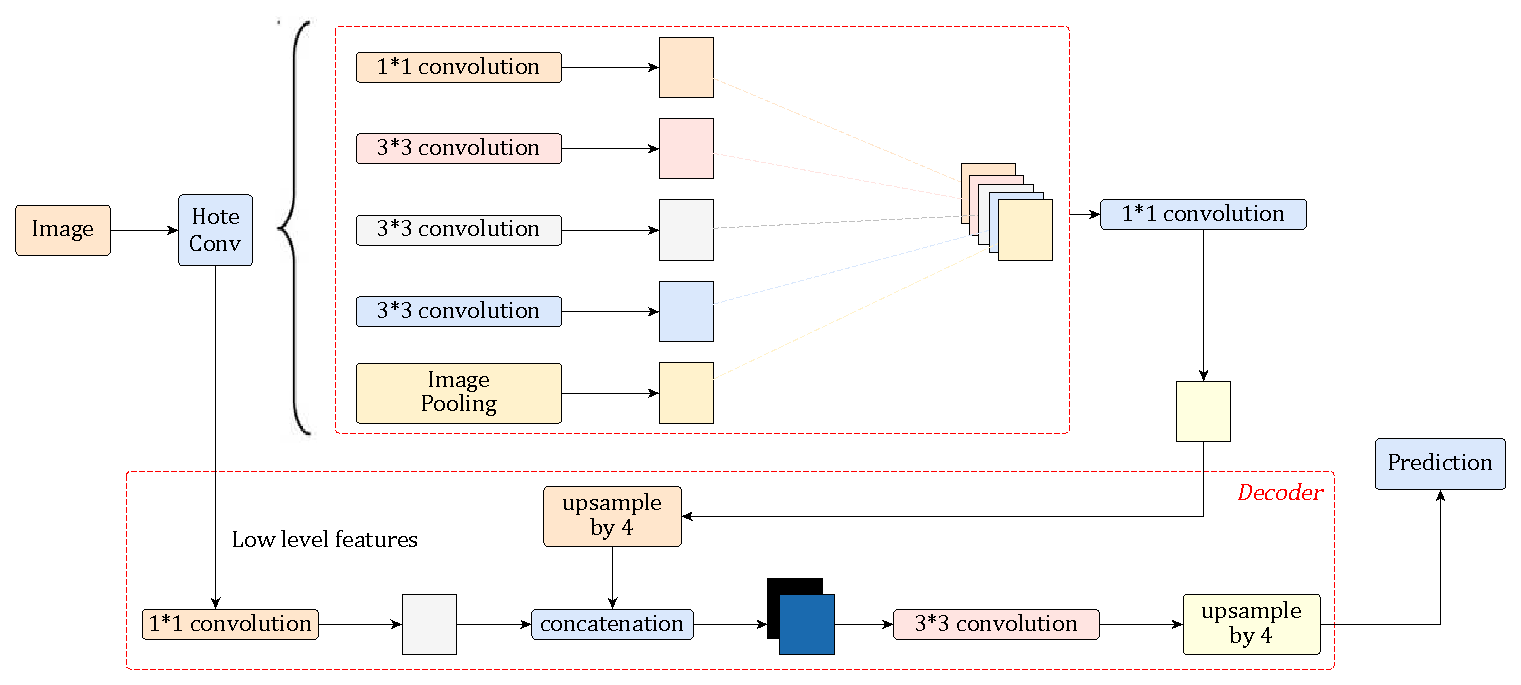}
    \captionsetup{font=scriptsize}
    \caption{DeepLab V3+ flow chart (\cite{cao_ecological_2023}).}
    \label{fig:label-de-votre-figure2}
\end{figure*}
\begin{table*}[ht]
\centering
\scriptsize 
\captionsetup{font=scriptsize} 
\caption{Methods and algorithms for water resources management.}
\label{tab:water_resources_methods}
\renewcommand{\arraystretch}{1.5} 
\begin{tabular}{|m{2.5cm}|m{13.5cm}|m{1.25cm}|} 
\hline
\multicolumn{1}{|c|}{\textbf{Applications}} & \multicolumn{1}{c|}{\textbf{Methods and algorithms}} & \multicolumn{1}{c|}{\textbf{References}} \\ \hline

\multirow{4}{*}{Hydrological modelling} 
& U-Net model. & \cite{du_drainage_2024} \\ \cline{2-3}
& Quadtree Decomposition. & \cite{sun_quadtree_2024} \\ \cline{2-3}
& Reflectance Transformation combined with Convolutional Neural Network. & \cite{on_chan_novel_2024} \\ \cline{2-3}
& Weighted Average Super Ensemble, Extra Tree Regression Super Ensemble and Bayesian Model Averaging Super Ensemble. & \cite{wegayehu_super_2023} \\ \hline

\multirow{4}{*}{Ground water} 
& Convolutional Neural Network - eXtreme Gradient Boosting. & \cite{al-ruzouq_hybrid_2024} \\ \cline{2-3}
& Logistic Regression, Support Vector Machine and eXtreme Gradient Boosting. & \cite{sahour_identification_2022} \\ \cline{2-3}
& Long Short-Term Memory and Convolutional Neural Network-Long Short-Term Memory. & \cite{seo_predicting_2021} \\ \cline{2-3}
& Random Forest, Support Vector Regression and Multi-Layer Perceptron. & \cite{sabzehee_enhancing_2023} \\ \hline

\multirow{3}{*}{Water quality} 
& Ensemble learning approach together with Higher Order Singular Value Decomposition, Deep Neural Networks and Support Vector Machine. & \cite{sun_aerosol_2021} \\ \cline{2-3}
& Transformers, Mixture Density Network, Random Forest, Gradient Boosting and eXtreme Gradient Boosting. & \cite{yang_monitoring_2023} \\ \cline{2-3}
& Linear Regression along with Deep Neural Networks. & \cite{vakili_determination_2020} \\ \hline

\multirow{4}{*}{Precipitation forecasting} 
& Heterogeneous Spatiotemporal Attention Fusion Prediction model. & \cite{niu_heterogeneous_2023} \\ \cline{2-3}
& Spatio Temporal Inference Network. & \cite{jin_spatiotemporal_2024} \\ \cline{2-3}
& Fully Convolutional Neural Network, Convolutional Neural Network, Recurrent Neural Network, Long Short-Term Memory, Gated Recurrent Unit, U-Net and Generative Adversarial Network. & \cite{sengoz_machine_2023} \\ \cline{2-3}
& PseudoFlow SpatioTemporal Long Short-Term Memory. & \cite{luo_pfst-lstm_2021} \\ \hline

\end{tabular}
\end{table*}
\subsubsection{Hydrological modelling}
Thanks to their efficiency and performance, GeoAI methods permit a comprehensive modelling of hydrological events. In this context, Islam et al. \cite{islam_mutual_2023} used DL-based space-time image velocimetry (DL-STIV) \cite{watanabe2021improving} in order to estimate hydrological parameters including water flow velocity and flow rate in Asahi Japanese river. UAV images are collected and used to train the model. Furthermore, Digital Elevation Models (DEM) derived from UAV-borne LIDAR data are exploited to validate the results obtained, i.e. flow velocities with MSEs between 0.01 and 0.24 m/s and flow rates estimated with accuracies of up to 11\%. For the purpose of extracting drainage network in low relief areas, Du et al. \cite{du_drainage_2024} have proposed a U-Net architecture \cite{ronneberger2015u} using LIDAR data as input. Results of comparison with Random Forest model show a good performance of the proposed model, obtaining a precision of 88\% and a recall of 89\%.
\\
The definition of water-land boundaries provides a better understanding of interactions between terrestrial and aquatic entities, giving greater precision in the calculation of flows and the qualification of environmental impacts. In this context, Sun et al. \cite{sun_quadtree_2024} presented a DL model for coastline extraction. This model is based on quadtree decomposition, an algorithm used to divide input images recursively into four equal regions \cite{shusterman1994image}. This method is used to reduce computational costs, to eliminate non-interesting areas and to manage scaling. Moreover, the authors use high-resolution (HR) satellite images from Google Map server using the Tile Map Service (TMS). A multi-scale classification is used based on two DL networks, MobileNet V3 \cite{howard2019searching} and Inception V4 \cite{szegedy2017inception}. This model proved a high classification performance, presenting an accuracy improvement of over 6\% compared to traditional architectures, for instance Fully Convolutional Network (FCN) and U-Net \cite{ronneberger2015u}. In a different context, Onchan et al. \cite{on_chan_novel_2024} proposed a Reflectance Transformation - Convolutional Neural Network (RT-CNN) for generating river bathymetric data from China. The authors use Landsat-7, 8 and Sentinel-2 satellites images to achieve an improved accuracy of over 18\% compared to other methods such as SVM, RF and Gradient Boosting Decision Tree. The goal of this work is to predict the response of water systems to various environmental conditions. Wegayehu and Muluneh \cite{wegayehu_super_2023} have used data from MODIS instrument to calculate a set of environmental indices, including the Normalized Difference Water Index (NDWI) \cite{mcfeeters1996use}, i.e. a remotely sensed index used to identify water surfaces, calculated in the same way as NDVI but using green and near-infrared bands \cite{mcfeeters1996use}.  A set of precipitation data are employed, namely IMERG data, CHIRPS precipitation measurements, a case in point is shown in Figure 37, Multi-Source Weighted-Ensemble Precipitation Version 2 (MSWEP-V2) to simulate streamflow in three river basins in Ethiopia. Super ensemble approach, being an advanced approach using multiple inferences from a multitude of models, called learners, to form a more accurate model \cite{van2007super}, is applied. Three models are exploited in this regard, in particular Weighted Average Super Ensemble (WASE), using a weighted average of predictions to form the result model, Bayesian Model Averaging Super Ensemble (BMASE), featuring Bayesian inference for the same reason and Extra Tree Regression Super Ensemble (ETRSE), exploiting Extra Tree Regression (ETR) \cite{geurts2006extremely}. This combination resulted in an improved accuracy, with an R² coefficient of up to 77\%, and a better adaptation to different scenarios.
\begin{figure}[H]
    \centering
    \includegraphics[width=0.9\linewidth]{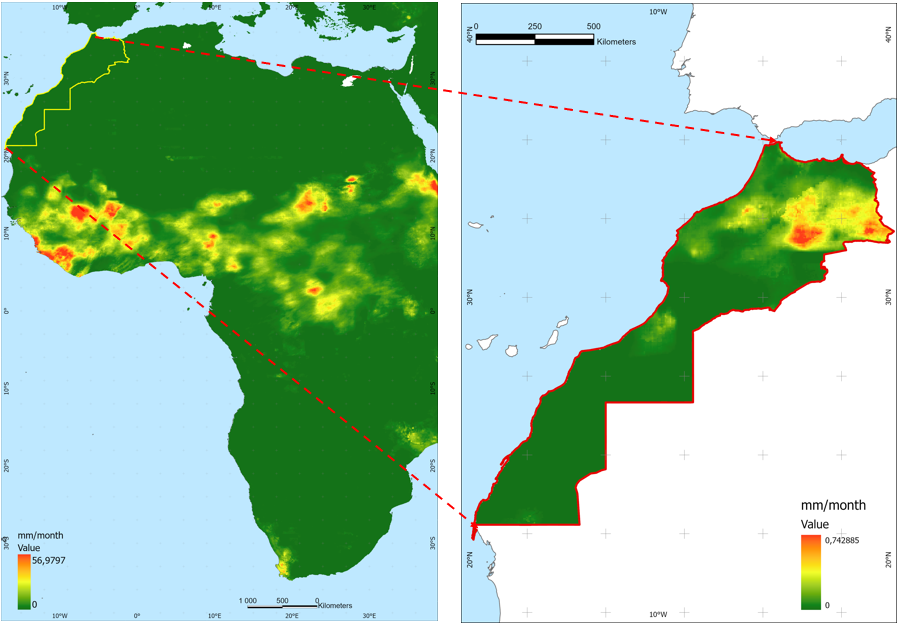}
    \captionsetup{font=scriptsize}
    \caption{CHIRPS monthly data example from Morocco and Africa, dated August 2024.}
    \label{fig:label-de-votre-figure2}
\end{figure}

\subsubsection{Groundwater}
Groundwater management is an important area of involvement for researchers in the field of AI. More specifically, and given its captivating ability to process geospatial data, GeoAI gives water specialists the opportunity to carry out numerous tasks, such as delineating groundwater recharges, predicting areas containing potential aquifers, ensuring sustainable management of these resources as well as preventive detection of possible contamination.
\\
For the purpose of groundwater recharge delineation in the United Arab Emirates (UAE), Al-Ruzouq et al. \cite{al-ruzouq_hybrid_2024} have used a hybrid of convolutional neural network and extreme gradient boosting (CNN-XGB). For model prediction and training, nine hydrogeological factors are utilized in this regard, specifically precipitation, elevation, drainage density, geomorphology, geology, water levels, Total Dissolved Solids (TDS), lineament density together with distance to residential areas. The said model is able to achieve an accuracy of 81\%, a recall of 78\% and an F1-Score of 80\%. Along the same lines, Sahour et al. \cite{sahour_identification_2022} have compared a set of models such as Logistic Regression (LR), SVM and extreme gradient boosting (XGB), with the aim of identifying areas of shallow groundwater. Data used in this comparison includes remote sensing images from Sentinel-1, and Landsat-8 satellites, MODIS data along with other geomorphological data, in particular elevation, slope, curvature, distance to undercut features, soil moisture, radar backscatter coefficient, NDVI and LST. Test results show an accuracy of 93\% for XGB method, compared to SVM (88\%) and Logistic regression (87\%).
\\
In order to predict potential groundwater resources, Nhu et al. \cite{nhu_new_2024} have based on a TensorFlow Deep Neural Network (TF-DNN), using 12 influencing factors and 733 locations of groundwater sources. Related factors includes slope, elevation, curvature, NDVI and Normalized Difference Moisture Index (NDMI), i.e. an index reflecting the water content of vegetation using the Near InfraRed (NIR) and Short-Wave InfraRed (SWIR) bands \cite{jin2005comparison}. TF-DNN demonstrated high efficiency, achieving an accuracy of 80\% and an F1-Score of 79\%. In a similar context, Seo and Lee \cite{seo_predicting_2021} compared two models, more precisely the standard LSTM and the combinatorial Convolutional Neural Network-LSTM (CNN-LSTM) model for predicting spatio-temporal variations in groundwater storage. Terrestrial water storage anomaly, recorded by GRACE and GRACE-FO satellites, precipitation from Tropical Rainfall Measuring Mission (TRMM) satellite, as well as NDVI and modified normalized difference water index (MNDWI) \cite{xu2006modification} acquired via Landsat 5 and 8 satellites are employed as input data for the models. Final results have favored the use of CNN-LSTM in predicting changes in waterground storage, acquiring an RMSE of 44.92 mm/month.
\begin{figure}[H]
  \centering
  \includegraphics[width=0.9\linewidth]{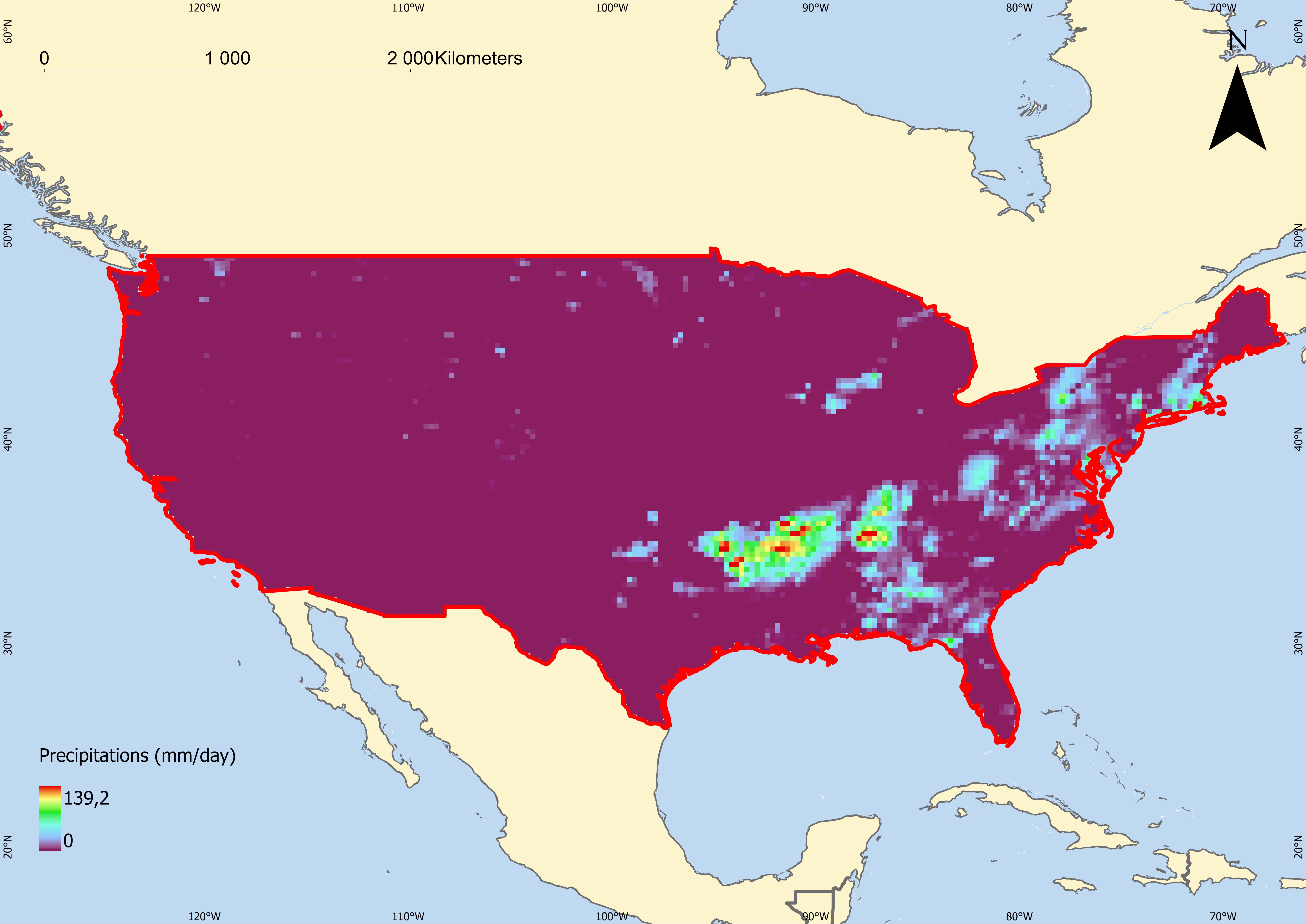}
  \captionsetup{font=scriptsize}
  \caption{Daily map of TRMM data following the Midwest floods in the USA (06/15/2008).}
  \label{fig:bar_chart2}
\end{figure}
\noindent Another application theme is to improve the sampling frequency of groundwater-related data, in which context Sabzehee et al. \cite{sabzehee_enhancing_2023} presented a multitude of models, including RF regression, Support Vector Regression (SVR) \cite{drucker1996support} and MLP. The aim of this study is to improve the spatial resolution of groundwater storage anomaly, obtained from GRACE sattelite. The model takes as input hydro-climatic variables such as precipitation, LST and NDVI. RF model have obtained the highest performance with regard to the aim of this study, with an RMSE of 18.36 mm.
\\
From the above, it has emerged that GeoAI plays an important role in forecasting the geographical delineation of groundwater resources. The results of these forecasts, through the five research works cited, reflect the performance guaranteed by AI models in combination with geospatial data, particularly with regard to the subject in question.
\subsubsection{Water quality}
Artificial intelligence techniques, combined with geo-information, can also be a powerful tool for precise quantification and qualification of water quality, using remote sensing products and in-situ measurements, referring to direct field measurements, as useful and relevant acquisition techniques. Sun et al. \cite{sun_ensemble_2021} proposed an ensemble learning model trained on remotely sensed images, based on Integrated Data and Classifier Fusion via Higher Order Singular Value Decomposition (IDCF-HOSVD). As indicated, the authors use HOSVD \cite{de2000multilinear}, a mathematical method for feature extraction and dimension reduction. In addition to generalizing Singular Value Decomposition (SVD) to multi-dimensional vectors, the classification results of several models, such as neural networks and SVM is applied. The main objective of this work is to measure the parameter "chlorophyll-a" in the water of Lake Nicaragua in Central America, reflecting the degree of nutrimants in the water \cite{bjorn2009viewpoint}. The IDCF-HOSVD model showed favourable results when comparing actual results with ground truth measurements. Yang et al. \cite{yang_monitoring_2023} have studied water quality in krastic wetlands, using multi-spectral and hyperspectral UAV images, combined with ground measurements. Concerned process includes the use of various models, namely transformers, mixture density network \cite{bishop1994mixture}, RF, XGBoost and gradient boosting. These models are analyzed using SHapley Additive exPlanations (SHAP) method \cite{lundberg2017unified}, to predict the parameters chlorophyll-a (Chla), phycocyanin (PC), as photosynthetic pigments, turbidity (Turb), relating to the density of particles suspended in water, and dissolved oxygen (DO), a significant value of which represents good water quality. In conclusion, the transformers achieved the best predictions in terms of PC and DO, with R² coefficients of 0.649 and 0.844 respectively, while the XGB and gradient boosting models achieved favorable results in terms of chlorophyll-a estimation, with an R² of 75\%. Additionally, Wu et al. \cite{wu_spatiotemporal_2022} exploited remote sensing images to measure water quality in Zhejiang coastal area from China. A Spatiotemporal Deep Belief Network is applied on a dataset of MODIS images along with field data in order to estimate concentrations of nutrients, especially Dissolved Inorganic Nitrogen (DIN) and Dissolved Inorganic Phosphorus (DIP). The inference results have provided mean values for the coefficients of determination R² of 48\%, it is noted that Figure 39 shows the prediction results over 9 years. Besides, Vakili and Amanollahi \cite{vakili_determination_2020} used Landsat 8 satellite images to measure water quality through optically inactive water variables including total Nitrogen (TN) and total Phosphorus (TP). Linear Regression and artificial neural network (ANN) are modeled to identify the correlation between Landsat 8 images as well as TP and TN concentrations. This study showed a high performance of ANN, obtaining an MAE of 5\%, an RMSE of 6\% and an R² of 86\%.
\begin{figure}[H]
  \centering
  \includegraphics[width=0.9\linewidth]{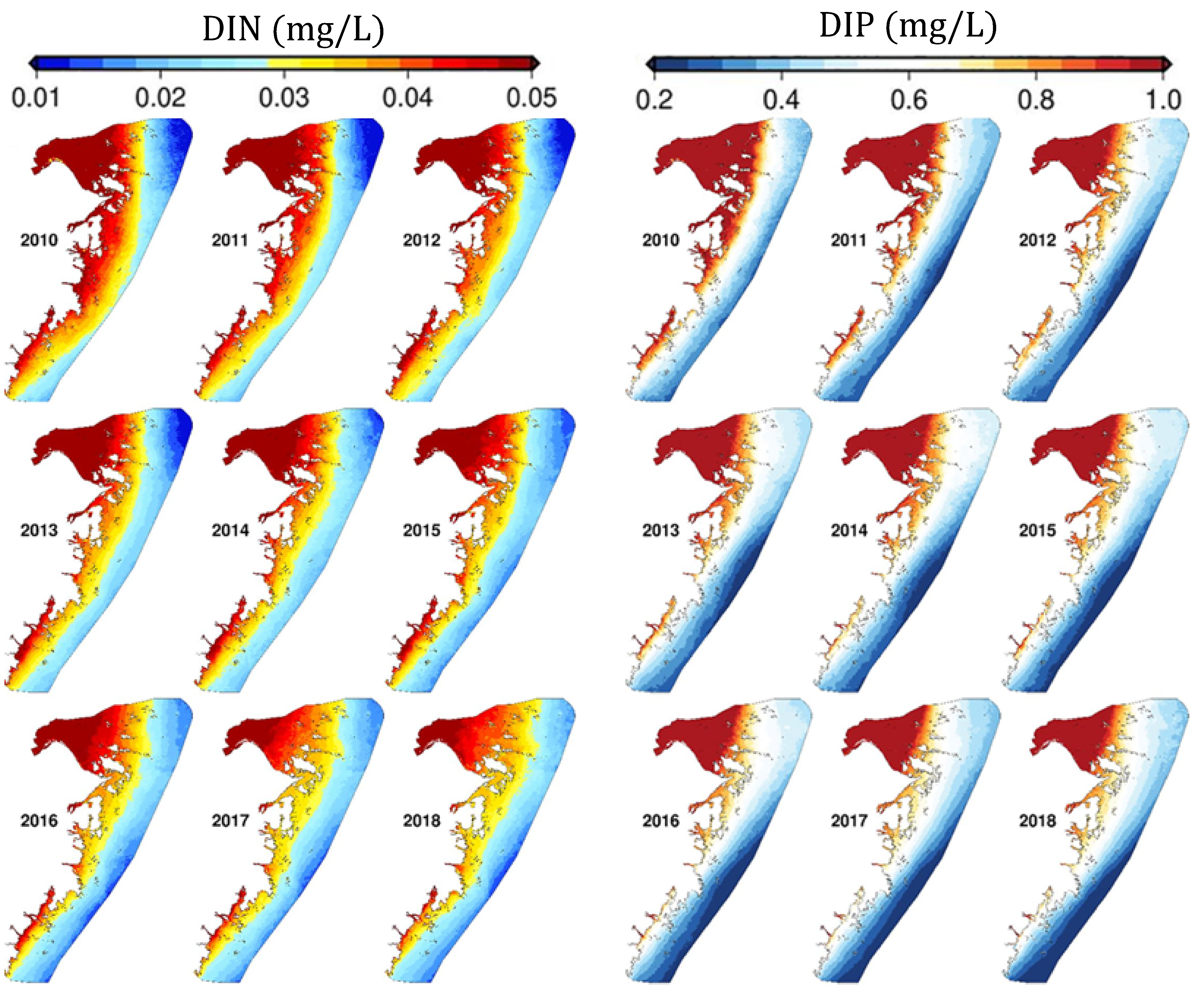}
  \captionsetup{font=scriptsize}
  \caption{Water nutrient concentrations from \cite{wu_spatiotemporal_2022}.}
  \label{fig:bar_chart2}
\end{figure}
\noindent Optical characteristics, such as diffuse attenuation coefficient and transparency, indicates precisely the composition of water and the presence or absence of various substances. Filisbino Freire da Silva et al. \cite{filisbino_freire_da_silva_machine_2021} proposed an SVM architecture combined with a sigmoïde function at the end of model to classify known optical water types (OWTs) and to detect new OWTs. Train, validation and test data are from Sentinel-2 satellite MultiSpectral Instrument (MSI) sensor. Indeed, cross-validation procedure showed an overall accuracy (OA) of the model of up to 94\%.
\\
As a summary, numerous studies have focused on the impact of GeoAI models in assessing and monitoring water quality. These models accurately predicted a number of parameters related to the subject, touching on various aspects such as eutrophication, chemical composition and physical characteristics, enabling sustainable monitoring and effective prevention of water pollution.
\subsubsection{Precipitation forecasting}
AI techniques for water resources management have applications in a wide range of fields, including quantification and prediction of precipitation. This includes nowcasting, geospatial analysis of precipitation and prediction of particle natures in rainfall. These tasks naturally lead to improved water storage procedures, planned agriculture and optimised water use. In order to carry out immediate precipitation forecasts, Niu et al. \cite{niu_heterogeneous_2023} have developed a Heterogeneous Spatiotemporal Attention Fusion Prediction (HST-AFP) network for nowcasting precipitation using input data such as weather forecasts and radar echo observations. The proposed model performed well in comparison to U-Net \cite{ronneberger2015u} and Recurrent Neural Network (RNN) methods, with a Critical Success Index (CSI) of 37\% in test phase, referring to a weather forecasting metric, corresponding to the ratio of correct forecasts to all possible events \cite{donaldson1975objective}. With the same objective, Jin et al. \cite{jin_spatiotemporal_2024} used a set of meteorological data including wind speed, humidity and temperature. As a basic model, a Spatio Temporal Inference Network (STIN) is adopted by proving high prediction capabilities, several evaluation metrics are used to validate this approach by obtaining an IOU of 43.15 and 30.37 at the end of the next hour on two different datasets. 
\\
For real-time precipitation forecasting in North America, Sengoz et al. \cite{sengoz_machine_2023} employed a multi-model approach using several neural network architectures, especially Fully Connected Neural Network (FCNN), CNN, RNN, LSTM, Gated Recurrent Unit (GRU) \cite{chung2014empirical}, U-Net \cite{ronneberger2015u} and GAN. The goal of this research is to improve the results of 8 basic Numerical Weather Prediction (NWP) \cite{lorenc1986analysis} models, referring to weather forecasting systems using mathematical models and antecedent meteorological data \cite{pu2019numerical}. The choice of this approach is made to capture spatial and temporal characteristics of data, to enhance NWP results and to reduce its errors. Results in question showed a 17\% improvement in MAE values compared to NWP outputs. Furthemore, Luo et al. \cite{luo_pfst-lstm_2021} proposed PseudoFlow SpatioTemporal Long Short-Term Memory (PFST-LSTM), in order to refine precipitation forecasts over Convolutional LSTM (Conv-LSTM) \cite{shi2015convolutional}, trajectory gated recurrent unit (TrajGRU) \cite{shi2017deep}, Spatio-Temporal LSTM (ST-LSTM) and predictive recurrent neural network (PredRNN) \cite{wang2017predrnn}. The model is a variant of LSTM introducing pseudo-flow prediction on spatio-temporal data. A 101×101 km spatial echo-radar maps from CIKM AnalytiCup 2017 competition, of which a detailed example dataset is shown in Figure 40, is employed while demonstrating a better efficiency during evaluation phase with an average MSE of 82.11. 
\begin{figure}[H]
  \centering
  \includegraphics[width=\linewidth]{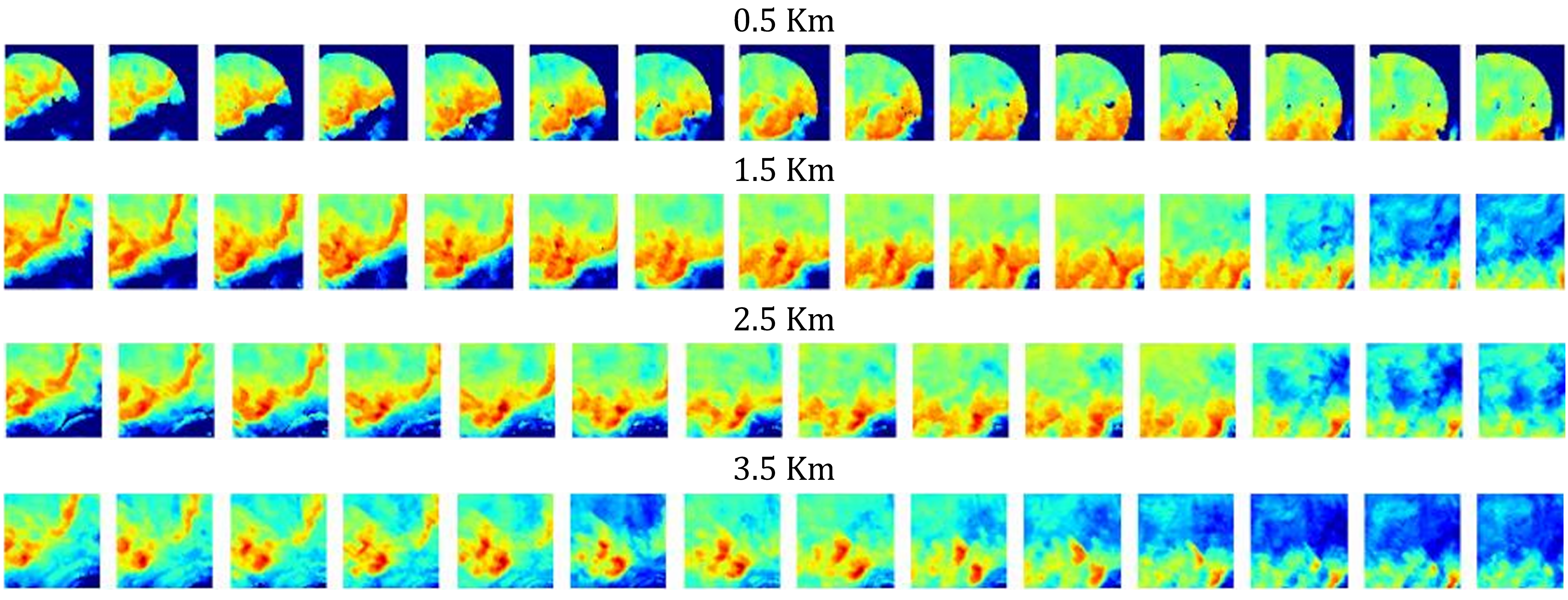}
  \captionsetup{font=scriptsize}
  \caption{Examples of Echo-radar maps of 15 time periods at 4 different altitudes, from CIKM AnalytiCup 2017 competition (\cite{yao2017cikm})}.
  \label{fig:bar_chart2}
\end{figure}
\noindent To predict the spatial distribution of stable oxygen isotopes in precipitation, Erdélyi et al. \cite{erdelyi_predicting_2023} have verified several methods, in particular Regression Kriging (RK) \cite{hengl2007regression}, corresponding to a geostatistical method combining Linear Regression (LR) and geospatial interpolation of a random variable, being the residuals of the said LR \cite{hengl2007regression}, as well as ML techniques, including standard RF, Multiple Regression Random Forest (MRRF) \cite{sekulic2020random} and Regression Enhanced Random Forest (RERF) \cite{zhang2019regression}. These methods are then validated using MSE metric. Results show varied MSE values for the RK, MRRF, RERF and RF methods, which are 2.77, 2.61, 2.99 and 3.08 respectively, demonstrating the effectiveness of ML models compared to the KR model, particularly for MRRF.
\\
Hence, a wide range of geospatial data are used to forecast precipitation, including satellite images and ground station measurements. These data are combined with traditional geostatistical methods, numerical weather models and employed by AI models to obtain accurate predictions validated by ground truth. Moreover, these AI models are a powerful tool in the service of another rather critical sector, that of disaster management. This field, requiring rapid interventions and accurate forecasting, is dissected munitiously in the following subsection, while presenting the most interesting research papers in the area.
\subsection{Disaster management}
The combination of geospatial datasets combined with AI models in monitoring of natural disasters represents a major advance forward for this type of risk. Despite this, a major limitation exists in the search and selection of geospatial data relating to natural disasters, as these data must be acquired exactly at the time of the concerned event. Figure 41 shows a concrete example of such data, presenting a pair of pre/post earthquake images taken from the xBD dataset \cite{xview2dataset}, constituting an input for related AI models. Furthermore, a number of interesting works are published on the subject, particularly in relation to earthquakes, floods and forest fires. Table 12 shows the main researches listed in this subsection.
\begin{figure}[H]
    \centering
    \includegraphics[width=1\linewidth]{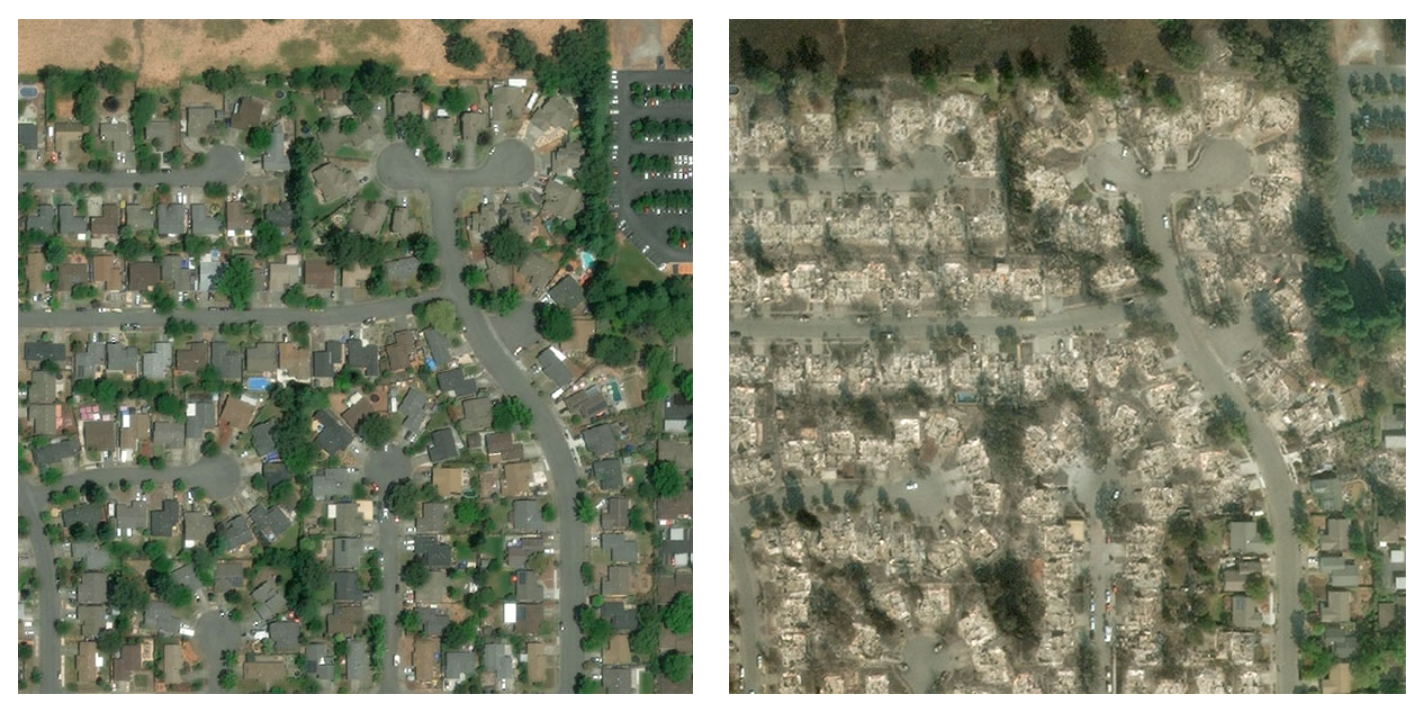}
    \captionsetup{font=scriptsize}
    \caption{Pair of xBD pre/post earthquake dataset images (\cite{ritwik2019xbd}).}
    \label{fig:label-de-votre-figure2}
\end{figure}
\begin{table*}[ht]
\centering
\scriptsize 
\captionsetup{font=scriptsize} 
\caption{Set of methods and algorithms presented for natural disasters management.}
\label{tab:methods_algorithms}
\renewcommand{\arraystretch}{1.5} 
\begin{tabular}{|m{1.5cm}|m{12cm}|m{1.25cm}|} 
\hline
\multicolumn{1}{|c|}{\textbf{Applications}} & \multicolumn{1}{c|}{\textbf{Methods and algorithms}} & \multicolumn{1}{c|}{\textbf{References}} \\ \hline

\multirow{5}{*}{Earthquakes} 
& Convolutional Neural Network. & \cite{jena_earthquake_2021} \\ \cline{2-3}
& Extended Long Short-Term Memory. & \cite{xiong_gnss_2022} \\ \cline{2-3}
& Artificial Neural Network. & \cite{hafeez_possible_2021} \\ \cline{2-3}
& Decision Trees, Support Vector Machine, Bayesian Network and Logistic Regression. & \cite{torres_integration_2019} \\ \cline{2-3}
& K-means and Random Forest. & \cite{zahs_classification_2023} \\ \hline

\multirow{4}{*}{Floods} 
& Artificial Neural Networks, Support Vector Machine, Decision Trees, Bayesian Naive, Random Forest and Fuzzy Logic. & \cite{puttinaovarat_flood_2020} \\ \cline{2-3}
& Convolutional Long Short-Term Memory. & \cite{moishin_designing_2021} \\ \cline{2-3}
& U-Net combined with attention mechanisms. & \cite{li_unet_2022} \\ \cline{2-3}
& Fully Convolutional Network-8s. & \cite{hashemi-beni_flood_2021} \\ \hline

\multirow{5}{*}{Forest fires} 
& Artificial Neural Network coupled with a Generative Adversarial Network. & \cite{rathod_multipurpose_2023} \\ \cline{2-3}
& Neural Adaptive Fuzzy Inference System with Artificial Bee Colony optimization algorithm. & \cite{pham_classifying_2024} \\ \cline{2-3}
& Logistic regression, MobileNet, Visual Geometry Group, LeNet and Residual Networks. & \cite{chen_wildland_2022} \\ \cline{2-3}
& U-Net architecture. & \cite{cho_burned_2023} \\ \cline{2-3}
& Convolutional Neural Network. & \cite{siddique_sustainable_2024} \\ \hline

\end{tabular}
\end{table*} 
\subsubsection{Earthquakes}
New AI approaches using geospatial data enable more effective earthquake prediction and more accurate quantification of material and human damage. Jena et al. \cite{jena_earthquake_2021} used DEM, vector and attribute data to build a CNN aiming to calculate earthquakes risk in northern India. This model showed a high degree of effectiveness in mapping, obtaining an accuracy of 94\% and an F1-score of 91\%. 
\\
A number of studies use positioning data to identify the ionospheric disturbances preceding sedimentation. Xiong et al. \cite{xiong_gnss_2022} proposed an extended LSTM model with encoder-decoder architecture to detect pre-earthquake ionospheric disturbances, Total Electron Content-Global Navigation Satellite System (TEC-GNSS) measurements are utilized to train and evaluate the proposed model. A comparison is made between the LSTM and other models such as the AutoRegressive Integrated Moving Average (ARIMA) \cite{box2015time}, DNN, RF and SVM, showing a clear improvement in the detection of seismic irregularities. Hafeez et al. \cite{hafeez_possible_2021} used Artificial neural networks (ANNs) to examine the atmospheric and ionospheric disturbances preceding Pakistan earthquake in 2019. Input data includes Land Surface Temperature (LST) records from MODIS instrument, Total Electron Content-Global Positioning System (TEC-GPS) and Total Electron Content-Global Ionospheric Map (TEC-GIM) measurements before and after the earthquake. Prediction results showed a notable efficiency, obtaining temperature variations from 2.8°C to 13.4°C, as well as TEC-GIM and TEC-GPS value variations of 5 TECU for the post-earthquake periods.
\\
Thanks to their precision, LIDAR data, photogrammetric and aerial images are a valuable datasource of information for risk assessment studies. Torres et al. \cite{torres_integration_2019} combined the use of LIDAR point clouds, aerial orthophotos and satellite images to estimate the fragility of buildings to earthquakes in Spain. The process comprises three phases: image segmentation, extractions of buildings and assignment of seismic vulnerability to each footprint. A multitude of ML methods are exploited, specifically decision trees, SVM, logistic regression and Bayesian networks \cite{pearl1985bayesian}, which are graphical models formalising the relationships between various variables to make probabilistic deductions \cite{pearl1988probabilistic}. Test results provides an F1-Score of 79\%, 77\%, 80\% and 77\% for the four models respectively. Zahs et al. \cite{zahs_classification_2023} developed a change-based model to quantify post-earthquake damage. The authors have used Virtual Laser Scanning (VLS) data, referring to a technique for simulating LIDAR acquisition in a virtual environment \cite{hildebrand2022simulating}. This model trains to extract changes from this data, using k-means clustering and then uses photogrammetric point clouds for Random Forest classification of damage. As shown in Figure 42, VLS data is categorized into 4 different classes according to the severity of damage, using a post- and pre-earthquake data pair. The overall process is validated through a large batch of real point clouds while achieving an accuracy of 95\%. 
\begin{figure}[H]
    \centering
    \includegraphics[width=1\linewidth]{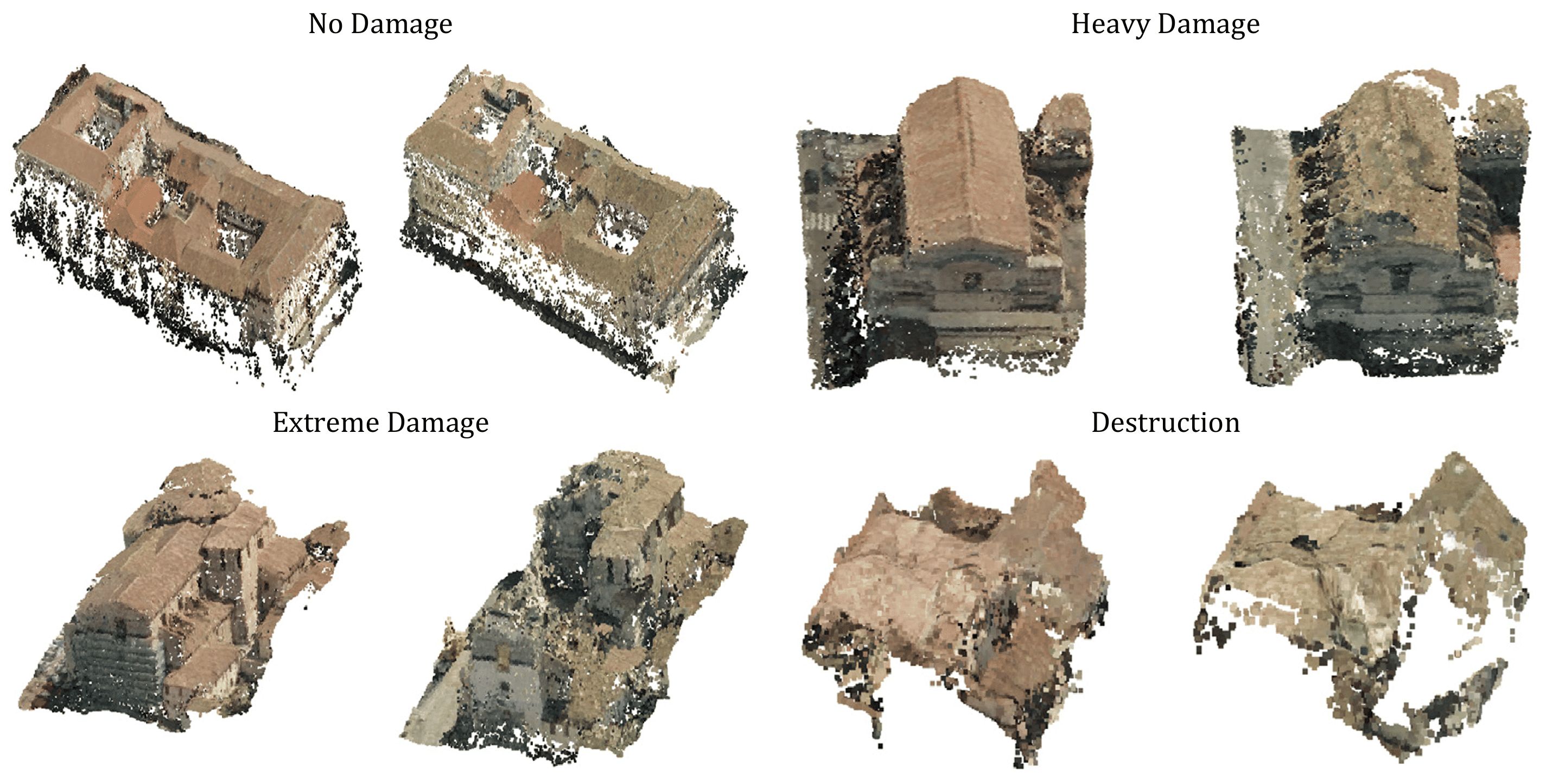}
    \captionsetup{font=scriptsize}
    \caption{Example of VLS data post (left) and pre (right) earthquake, each class of data is represented by an illustrative example (\cite{zahs_classification_2023}).}
    \label{fig:label-de-votre-figure2}
\end{figure}
\noindent
According to the above-mentioned research, GeoAI contributes effectively to the analysis and prediction of earthquakes, and to the estimation of earthquake damage. The benefits derived are essential for minimizing the effects of seismic events and predicting their occurrence.
\subsubsection{Floods}
Recent advances in GeoAI have enabled to draw up expressive vulnerability maps and precise models for flood forecasting purposes. Given this context, Puttinaovarat and Horkaew \cite{puttinaovarat_flood_2020} developed a flood and precipitation forecasting method based on big geodata, including Web Map Service (WMS), public raster records, meteorological, hydrological and rowdsourced data. Numerous models, specifically SVM, RF, decision trees, fuzzy logic \cite{zadeh1965fuzzy} and ANNs are evaluated to choose the most performant. Results of a k-fold cross-validation \cite{geisser1975predictive} favore the use of ANNs, SVM and RF. Another flood forecasting model is developed by Moishin et al. \cite{moishin_designing_2021} using a hybrid ConvLSTM \cite{shi2015convolutional} model, combining the architecture of CNN with LSTM to process spatiotemporal data, including rainfall data and Flood Index ($I_{F}$), referring to an index for real-time flood monitoring using daily rainfall measurements \cite{deo2015real}. This model has proven its ability to handle large-scale spatio-temporal datasets with notable accuracy.
\\
In addition to flood forecasting, a number of studies focused on mapping the affected areas. Li et al. \cite{li_unet_2022} presented a U-Net network \cite{ronneberger2015u} coupled with attention mechanisms to better extract inundated areas in China. Dual-polarized Sentinel 1 SAR images are employed for model training and evaluation. The results showed a remarkable contribution of this model compared to traditional methods, obtaining average precision values of 84\%, recall of 91\% and F1-Score of 87\%. Montello et al. \cite{montello_mmflood_2022} constructed a Multimodal Dataset for Flood delineation (MMFlood) from Sentinel 1 SAR images, along with elevation data, hydrographic maps and binary annotations. This dataset focuses on the reliability of SAR data combined with digital elevation models (DEMs) for flood mapping, particularly in cloudy environments. A multi-encoder architecture is employed to delineate floods from this dataset, proving the use of DEMs for greater precision, while reaching an F1 score of up to 79\% for this multimodal approach. In a similar context, Hashemi-Beni and Gebrehiwot \cite{hashemi-beni_flood_2021} proposed an advanced method to provide more accurate mapping of flood extents. To this end, UAV and aerial optical images, together with LIDAR data are collected by the National Oceanic and Atmospheric Administration (NOAA) \cite{noaa_official}. In a first phase, a particular Fully Convolutional Network (FCN-8s) is applied for image classification, the application of a tride of 8 for FCN-8s convolution layers shows better performance, proving the choice of this architecture. Then, delimitation of flooded areas is improved by a Regional Growing (RG) segmentation method using DEMs of water levels. The use of data augmentation techniques, such as cropping, rotation and translation, resulted in an overall accuracy of 97\%.
\\
Like earthquakes, multiple creative GeoAI models are implemented to map and forecast floods, essentially employing remote sensing data, aerial images and LIDAR point clouds in order to monitor and predict this hyper-descructive phenomenon.
\subsubsection{Forest fires}
Forest fires are one of the most damaging natural disasters, affecting both property and lives of animals and humans. Integrating innovative methods into real-time monitoring systems is a very useful way of identifying vulnerable areas and recognizing the most critical perimeters in the field. Numerous studies are carried out on the subject of preventing the most susceptible areas to forest fires. An ANN is used by Rathod et al. \cite{rathod_multipurpose_2023}, trained by data collected via a drone equipped with the Raspberry Pi4 module and coupled with temperature data from a Digital Humidity and Temperature DHT11 sensor. In addition, a GAN is implemented to augment the training dataset, while achieving an overall accuracy of 90\% for the overall model. Pham et al. \cite{pham_classifying_2024} aimed to predict forest fire susceptibility, through an Adaptive Neural Fuzzy Inference System (ANFIS) \cite{jang1993anfis}. Several types of data, especially Digital Elevation Models (DEM), NDWI and NDVI indices added to hydrometeorological stations are utilized in this settings. Additionally, an Artificial Bee Colony (ABC) algorithm \cite{karaboga2007powerful} is used to alleviate the problem of data multicollinearity, obtaining an RMSE of 7.28, an MAE of 5.02 and an R² of 86\% for the ANFIS-ABC model.
\\
Delineating forest fires is necessary for reliable and effective anticipation, as mapping the areas affected enables public services to better understand the severity of damages and to move swiftly on to the post-fire stage. It also enables stakeholders to better quantify the impact of these fires on biodiversity and ecological balance. Indeed, Chen et al. \cite{chen_wildland_2022} used Flame2 \cite{hopkins2022flame}, an improved version of Flame \cite{shamsoshoara2021aerial}, referring to a customized CNN for forest fire detection, while comparing it to a combination of DL models, in particular LetNet \cite{lecun1989backpropagation}, Visual Geometry Group (VGG) \cite{simonyan2014very}, MobileNet \cite{howard2017mobilenets} and ResNet \cite{he2016deep}. As illustrated in Figure 43, RGB and infrared (IR) images from drones are used to train comparative models. The evaluation stage proved the high detection efficiency of FLAME2, reaching an average F1-Score of 83\%. Also, Cho et al. \cite{cho_burned_2023} used a U-Net network \cite{ronneberger2015u} to map burnt areas from PlanetScope's satellites images (RGB+NIR). Complementary NDVI \cite{jones2010remote} and greyscale texture data are utilized to improve the accuracy of obtained results, leading to an F1 score of up to 93\%. An interesting framework for forest fire classification is developed by Siddique et al. \cite{siddique_sustainable_2024}, the chosen model in question is a three-layer CNN using fire and non-fire images from Wildfire Detection Image Dataset \cite{kaggle2024wildfire}. The proposed framework integrates Federated Stochastic Gradient Descent (FedSGD) \cite{mcmahan2017communication} to aggregate this model in the cloud, while obtaining an accuracy of 99\%.
\begin{figure}[H]
    \centering
    \includegraphics[width=1\linewidth]{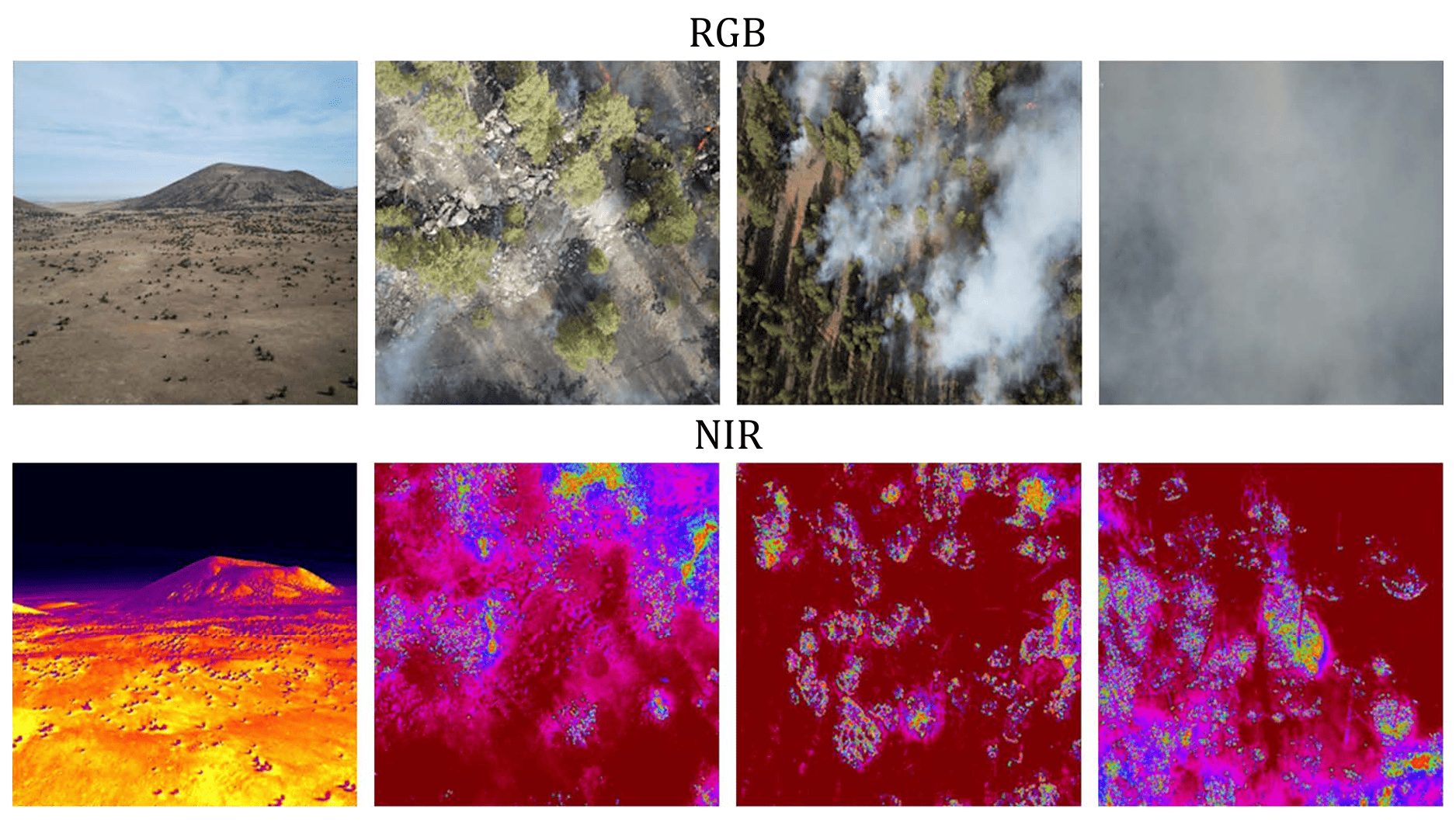}
    \captionsetup{font=scriptsize}
    \caption{Sample of RGB and IR data used for forest fires detection (\cite{chen_wildland_2022}).}
    \label{fig:label-de-votre-figure2}
\end{figure}
\noindent As mentioned above, several GeoAI models are proposed for forest fire prediction and detection, including the use of various data collected at different dates, helping to manage the adverse effects of this disaster on flora and fauna, as well as preserving the lives of thousands of people. In the same spirit of preservation, healthcare is a field enabling to predict the spread of epidemics and infectious diseases, to help analyze and exploit public health data, and to provide citizens with the care they need. The following subsection surveys the contribution of GeoAI in this field, mentioning the most interesting research in this regard.
\subsection{Healthcare}
Thanks to the reforming effect of artificial intelligence (AI), increased use of geospatial data enables the modelling of spread and impact of diseases, the planning of medical campaigns and the optimizing of resources. These advantages guarantee a “responsible governance” in front of epidemics and critical health crises. In this section, research projects adressing this issue are listed, while applying GeoAI methods to the surveillance of infectious diseases such as fevers, influenza and poisoning, together with public health monitoring, and epidemic modelling, taking the case of COVID-19 as an illustrative example. Table 13 shows all the proposed methods and algorithms. 
\begin{table*}[ht]
\centering
\scriptsize 
\captionsetup{font=scriptsize} 
\caption{Summary of methods and models in Healthcare field.}
\label{tab:methods_models_health}
\renewcommand{\arraystretch}{1.5} 
\begin{tabular}{|m{3cm}|m{13.25cm}|m{1.25cm}|} 
\hline
\multicolumn{1}{|c|}{\textbf{Applications}} & \multicolumn{1}{c|}{\textbf{Methods and algorithms}} & \multicolumn{1}{c|}{\textbf{References}} \\ \hline

\multirow{4}{*}{Infectious diseases and toxins} 
& Multi-stage Long Short-Term Memory. & \cite{venna_novel_2019} \\ \cline{2-3}
& Improved Artificial Tree Optimizing Backpropagation Neural Network. & \cite{hu_prediction_2018} \\ \cline{2-3}
& Support vector regression, Gradient Boosted Machine, Least Absolute Shrinkage and Selection Operator, Step-down Linear Regression. & \cite{guo_developing_2017} \\ \cline{2-3}
& Web search and location models. & \cite{sadilek_machine-learned_2018} \\ \hline

\multirow{5}{*}{Pandemics} 
& High-Resolution Geographical Convolutional Network. & \cite{zhao_predicting_2023} \\ \cline{2-3}
& Multi-Layer Perceptron, Support Vector Machine, Linear Regression and Random Forest. & \cite{da_silva_covid-19_2021} \\ \cline{2-3}
& Linear Regression. & \cite{atek_geospatial_2022} \\ \cline{2-3}
& Convolutional Neural Network combined with an Improved Spatial Transformer and a Multi Grid Module. & \cite{zhang_covid-19_2023} \\ \cline{2-3}
& Deep neural network and Deep regression model. & \cite{tan_environmental_2022} \\ \hline

\multirow{2}{*}{Public health} 
& Geographically-weighted Gradient Boosting Machine. & \cite{zhan_spatiotemporal_2017} \\ \cline{2-3}
& Hybrid Convolution Neural Network. & \cite{di_assessing_2016} \\ \hline

\end{tabular}
\end{table*}
\subsubsection{Infectious diseases and toxins}
In today's era of standardized health measures against different types of disease, the use of GeoAI techniques to combat these diseases and minimize their spread is proving to be of paramount importance. Venna et al. \cite{venna_novel_2019} developed a multi-stage LSTM model to predict influenza in the United States, using data from the Centers for Disease Control and Prevention (CDC) \cite{cdc_website}, illustrated by Figure 44, and Google Flu Trends (GFT) \cite{dugas2013influenza}, as well as climatic data including precipitation and temperature. The model's choice is prouved against Auto-Regression Integrated Moving Average (ARIMA) \cite{box2015time} method. LSTM model achieved the best prediction performance with an RMSE of 1.032 and a MAPE of 14 over 5 different weekly periods. In a related scenario, Hu et al. \cite{hu_prediction_2018} have aimed to predict the real-time spread of Inflenza-Like Illness (ILI), using twitter posts and other data from CDC \cite{cdc_website}, especially average percentages of medical visits for ILI-like symptoms. An Improved Artificial Tree BackPropagation Neural Network (IAT-BPNN) is proposed as the base model, while reaching an MSE of 0.049, an RMSE of 0.038 and a MAPE of 0.147. Lu et al. \cite{lu_improved_2019} propose AGRONet, a set-based approach to estimate the spread of influenza in the United States. The presented model employes a statistical autoregressive approach, used to implement predictions based on historical trends, Google search frequencies and electronic health records. A network-based approach is also utilized, using regularized multivariable regression to take advantage of synchronicities in the spatio-temporal history. Predictions from 37 states have reached an RMSE of around 1.2 and a MAPE of around 0.7.
\begin{figure}[H]
    \centering
    \includegraphics[width=1\linewidth]{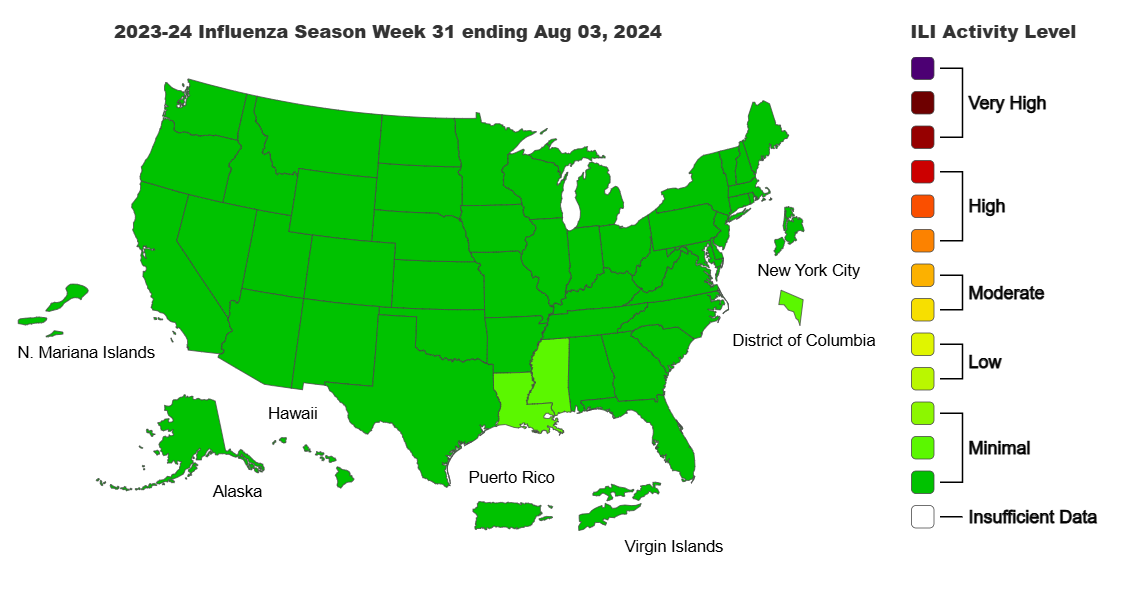}
    \captionsetup{font=scriptsize}
    \caption
    {Weekly U.S map of CDC data, available at the link \cite{CDC_Influenza_Surveillance}.}
    \label{fig:label-de-votre-figure2}
\end{figure}
\noindent In addition to flu forecasting, Guo et al. \cite{guo_developing_2017} compared a multitude of ML models, specifically Support Vector Regression (SVR) \cite{drucker1996support}, step-down linear regression, Gradient Boosted Machine (GBM) \cite{friedman2001greedy}, and Least Absolute Shrinkage and Selection Operator (LASSO) \cite{tibshirani1996regression}, for predicting dengue fever in Guangdong province from China. Weekly dengue case records and climatic data including temperature, humidity and precipitation from 2011 to 2014 are used. Comparison results favoured the SVR model, with an RMSE of 13.52 and an R² coefficient of 98\% during the total period.
\\
In addition to infectious diseases, diseases caused by poor eating habits and frequent poisoning are also a major focus of study. Sadilek et al. \cite{sadilek_machine-learned_2018} developed Foodborne IllNess DEtector in Real time (FINDER), an ML model for automatic detection of foodborne disease sources. Web search data are employed to train the model, based on two main components, a web search model (WSM) to calculate the probability linking web search and food-borne illnesses, besides a Location Model (LM) mapping the searches identified and analyzed by the WSM. Identified restaurants are found to be three times more likely to be hazardous than those identified by traditional methods, i.e. periodic inspections and those based on user complaints. 
\\
In summary, clinical observations and navigational research taking into account geographical locations are exploited to predict trends of infectious diseases and toxins, in order to speed up the response of the relevant authorities in the event of any crises. The next subsection looks at the case of pandemics, more prone to spread and therefore more pernicious.
\subsubsection{Pandemics : COVID-19}
GeoAI's analysis capacity have enabled interested parties to accurately model epidemics of various types, and to monitor their evolution over time and space. More specifically, and given the major impact of the COVID-19 pandemic on the whole world, a great deal of work in recent years focused on the contribution of GeoAI to predic the impact of this pandemic. In fact, it is decided to concentrate mainly on work relating to this pandemic, since it has attracted the greatest interest.
\\
Indeed, Zhao et al. \cite{zhao_predicting_2023} proposed COVID-19 high-resolution geographical convolutional network (CHGCN), a novel model for predicting the spread of human-to-human respiratory infectious disease (H-HRID). An innovative feature of the model incorporates the notion of spatial neighborhood through attention mechanisms, resulted in an average accuracy of 75\%. In the same logic, Da silva et al. \cite{da_silva_covid-19_2021} carried out a spatio-temporal predictive research of COVID-19 transmission in Brazil. Various models are used in this analysis, namely MLP, Linear Regression (LR), SVM and RF. In conclusion, MLP and LR give the best prediction efficiency, obtaining RMSE values of 4\% and 2\% respectively.
\\
Another field of application is the monitoring of COVID-19 pandemic clusters in order to ensure an in-depth understanding of transmissions. Atek et al. \cite{atek_geospatial_2022} present Earth Cognitive System for COVID-19 (ECO4CO), a comprehensive system for managing COVID-19. As shown in Figure 45, ECO4CO is utilized by two groups of users, end users who complete the monitoring and control tasks, and admins who configure these tasks and manage input data. Images from Pleiades satellite are used to train a DL network for identifying gatherings of people and vehicles, while giving reliable results. 
\begin{figure}[H]
    \centering
    \includegraphics[width=1\linewidth]{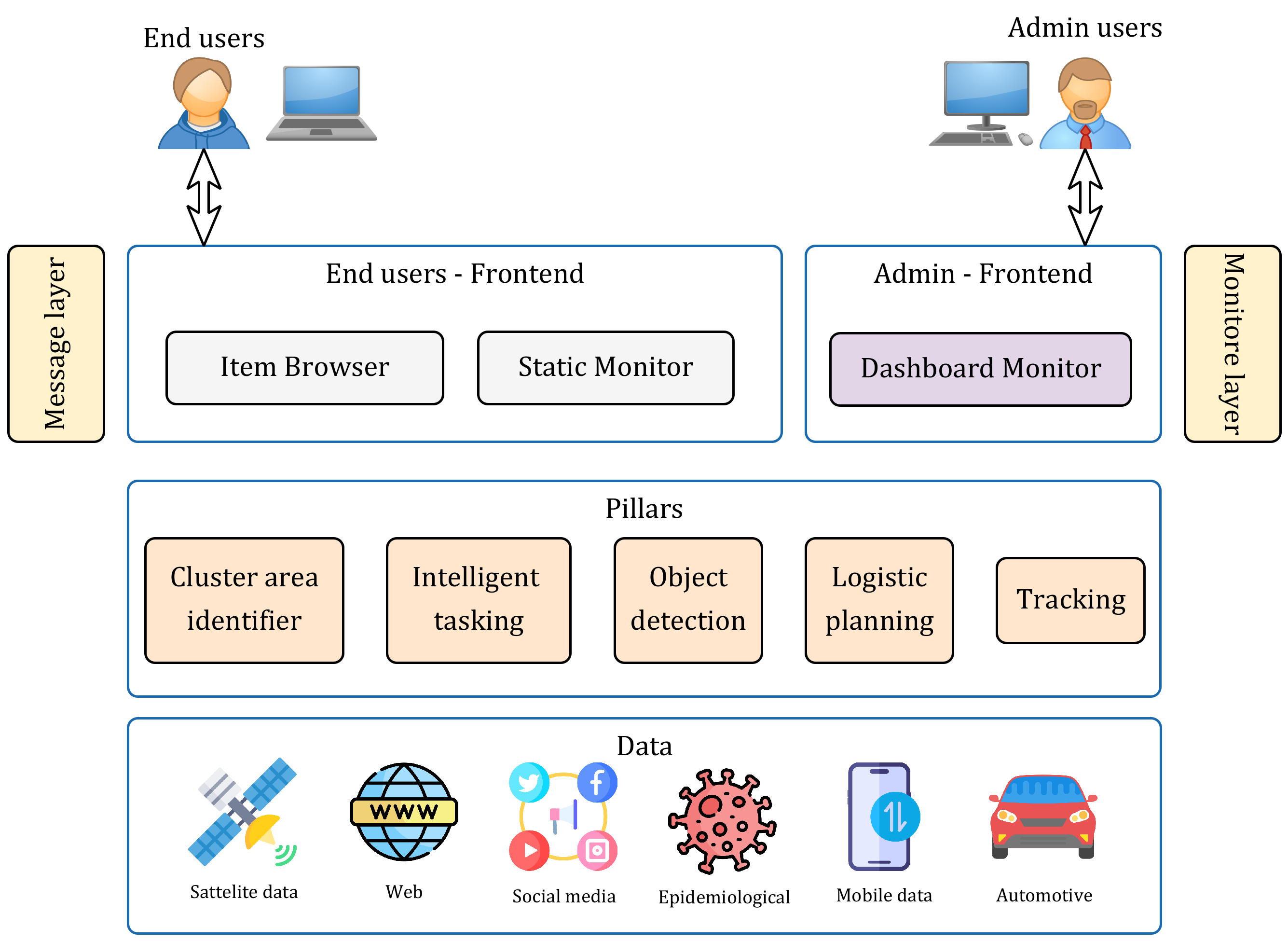}
    \captionsetup{font=scriptsize}
    \caption{Architecture of ECO4CO (\cite{atek_geospatial_2022}).}
    \label{fig:label-de-votre-figure2}
\end{figure}
\noindent For monitoring the so-called pandemic, Zhang et al. \cite{zhang_covid-19_2023} presented a multi-module DL approach, consisting of a CNN with batch normalization, an improved spatial transformer and a multi-grid module in order to identify individuals using video surveillance. As illustrated in Figure 46, this approach is implemented via a 3D visualisation using the city engine tool, providing a better understanding of the proximity and interactions between individuals. It is noted that this model achieved a high detection efficiency, while obtaining a mAP of 78\%.
\begin{figure}[H]
    \centering
    \includegraphics[width=1\linewidth]{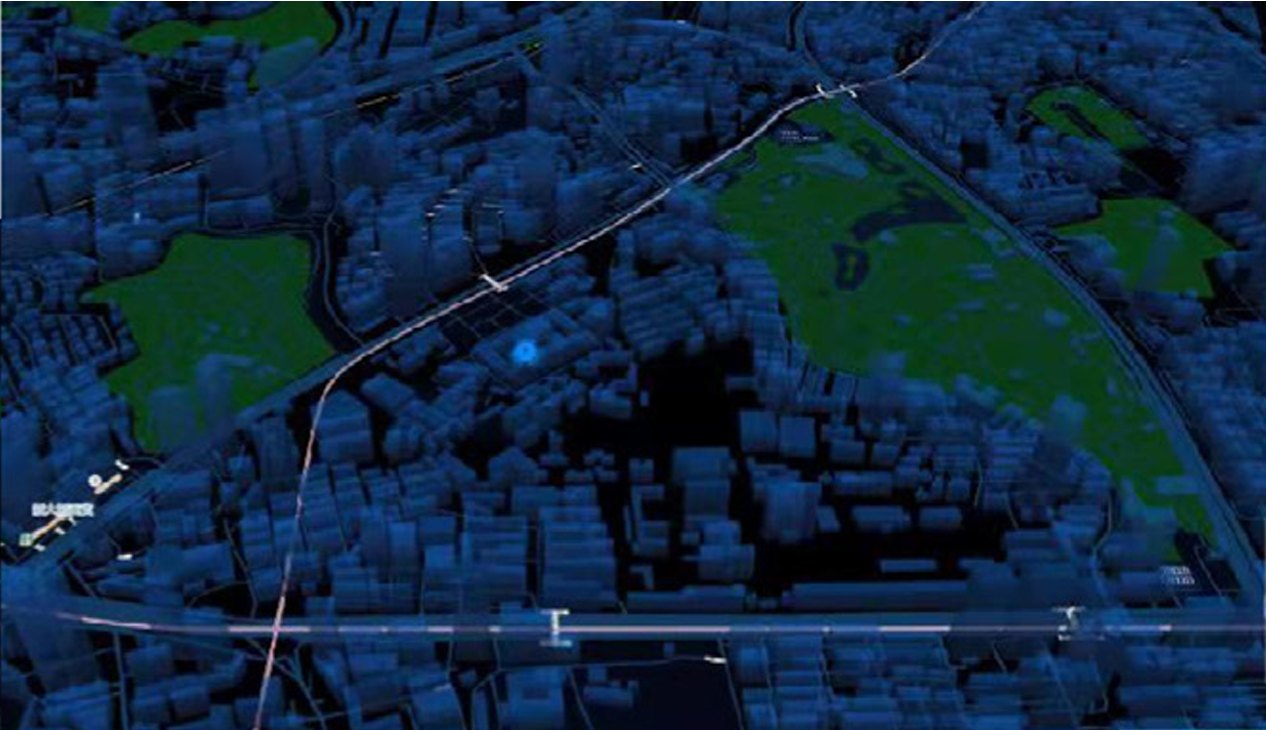}
    \captionsetup{font=scriptsize}
    \caption{Sample 3D model generated in \cite{zhang_covid-19_2023} for individuals proximity.}
    \label{fig:label-de-votre-figure2}
\end{figure}
\noindent Furthermore, Tan et al. \cite{tan_environmental_2022} studied the effect of COVID-19 on the concentration of PM2.5 in China. The authors use Aerosol Optical Depth (AOD) \cite{wei2020satellite} from MODIS along with UltraViolet Aerosol Index (UVAI) \cite{torres1998derivation} from Sentinel 5P sattelite. The approach consists of applying a DNN to correlate image data with PM2.5 concentrations and a Deep Regression model for predictions. This study have demonstrated high predictive capability, resulting in an MAE of around 17 µg/m³ and an R² of 53\%.
\\
From this census, it is clear that GeoAI constitutes a key tool in the management of pandemics. Taking COVID-19 as an illustrative example, GeoAI enables real-time monitoring of of this pandemic while tracking its impact on the population. Spatio-temporal data and updated statistics are used to this end, providing unrivalled support in the fight against and control of the pandemic, optimizing the human and material resources allocated for this purpose. Dealing with this subject, closely linked to the health and the safety of communities, obviously leads to take a particular interest in a broader challenge, that of public health.
\subsubsection{Public health}
In this subsection, the use of GeoAI models in public health is explored, with a few examples mainly concerning mobile health and air quality, as a parameter measuring the pollution of urban environments. According to Kahn et al. \cite{kahn2010mobile}, mobile health concerns the use of wireless communication devices for public health and clinical practice. Based on this, Kumar et al. \cite{kumar_center_2017} presented the Center of Excellence for Mobile Sensor Data-to-Knowledge (MD2K), effectively created in October 2014 and funded by the National Institutes of Health in the United States, to the end of interpreting a large mass of data from portable devices, mainly GPS location data and user behaviour. In addition, creative IA models such as puffMarker, mCrave and cStress are employed to train this data. These models show significant potential for addressing public health issues and ensuring rapid, accurate medical intervention when needed.
\\
Pollution is one of the main problems affecting the health of individuals and the purity of the environment, making it one of the main safety factors guaranteeing public and collective health. This is why many researchers are focused on this issue and its impact on collective health. Indeed, Zhan et al. \cite{zhan_spatiotemporal_2017} have implemented Geographically-Weighted Gradient Boosting Machine (GW-GBM) to accurately predict PM2.5 levels in China. The principle of this approach is to integrate geographical weights, i.e. spatial smoothing kernels, into a GBM model \cite{friedman2001greedy}. Aerosol data from MODIS instrument, ground-based PM2.5 measurements, demographic and climatological records are used to feed the model, achieving an RMSE of 23.0 µg/m³ and an R² coefficient of 76\%. In addition, Di et al. \cite{di_assessing_2016} presented a hybrid CNN to predict PM2.5 concentrations in the United States. Trained and tested datasets comprises Aerosol Optical Depth (AOD) images from MODIS instrument, Absorbing Aerosol Index (AAI) data from OMI instrument, illustrated in Figure 47 as an example, along with meteorological data and Land-Use terms. The deployed CNN performed very well, achieving an average coefficient R² of 84\%.
\begin{figure}[H]
    \centering
    \includegraphics[width=1\linewidth]{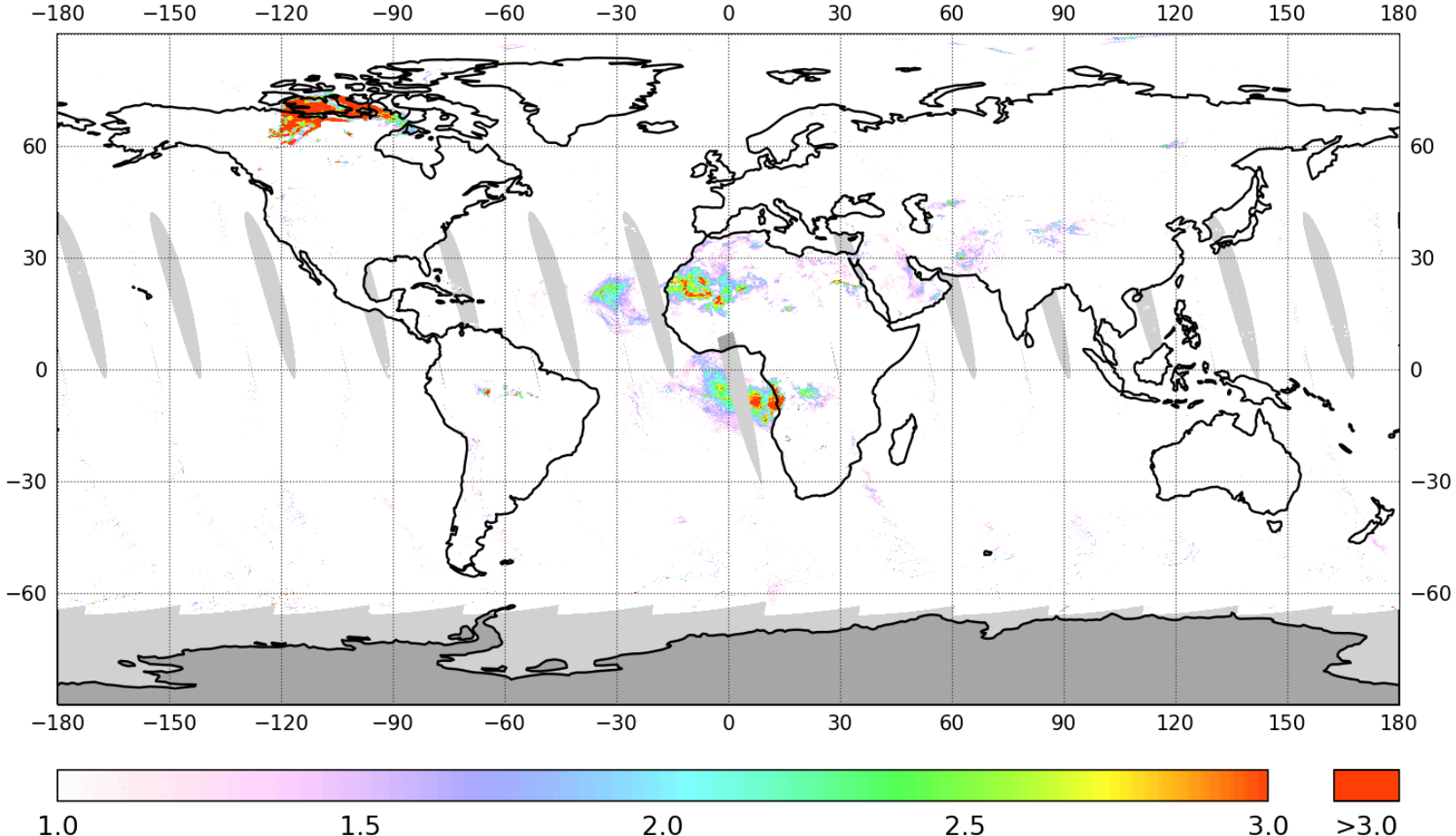}
    \captionsetup{font=scriptsize}
    \caption{Example of AAI data from OMI instrument, dated 08/11/2024 (\cite{temis_absaai}).}
    \label{fig:label-de-votre-figure2}
\end{figure}
\noindent Ultimately, numerous works have used original GeoAI models for public health, explored via remote sensing images and ground measurement data, and exploited for monitoring citizens behavior, for predicting pollution measurements and for analyzing their correlation with human activities. Indeed, implementing these models in decision-making systems is mandatory to ensure more effective governance in the face of today's challenges.
\section{Discussions and outlooks}
\subsection{Results}
Thanks to the review carried out, a body of research focusing on GeoAI is identified, while integrating the spatial dimension into the algorithms developed in this context. As shown in Figure 48, the majority of the research papers cited are fairly recent, i.e. 2022 and beyond.

\begin{figure}[H]
  \centering
  \includegraphics[width=1\linewidth]{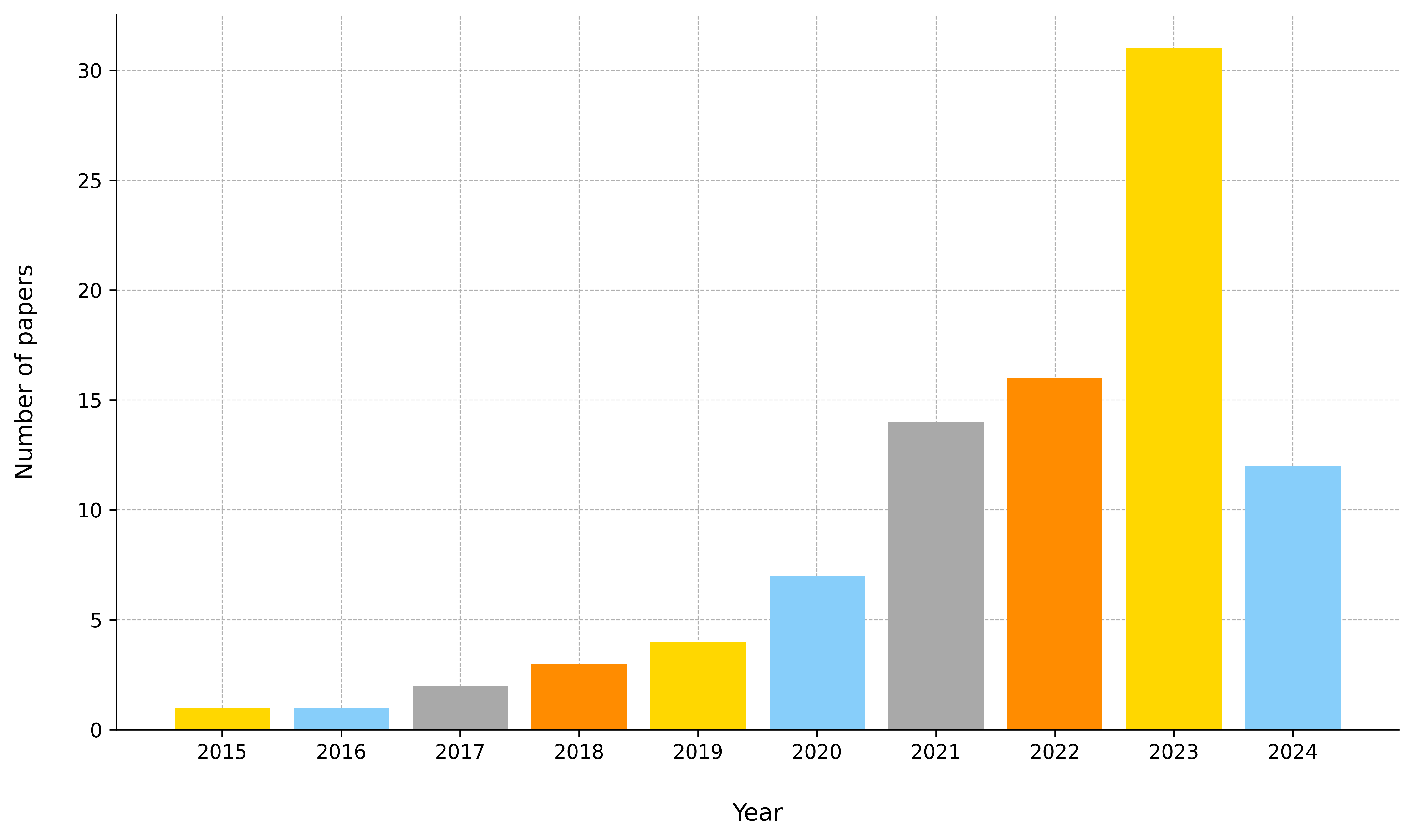}
  \captionsetup{font=scriptsize}
  \caption{Yearly papers numbers.}
  \label{fig:bar_chart2}
\end{figure}

\noindent Figure 49 presents a cloud of the most frequently used keywords in surveyed papers, offering a global perspective on the themes adressed. Terms such as \textbf{learning}, \textbf{data}, \textbf{image}, \textbf{neural network}, \textbf{detection}, \textbf{regression} and \textbf{classification} are used frequently, proving the interest of this review for learning methods, especially neural networks. In addition, several categories of data are employed to serve multiple tasks, particularly detection, classification and regression. Without forgetting to emphasize specific methods such as \textbf{spatio-temporal series}, \textbf{feature extraction} and \textbf{attention} calculation.
 \begin{figure}[H]
  \centering
  \includegraphics[width=0.8\linewidth]{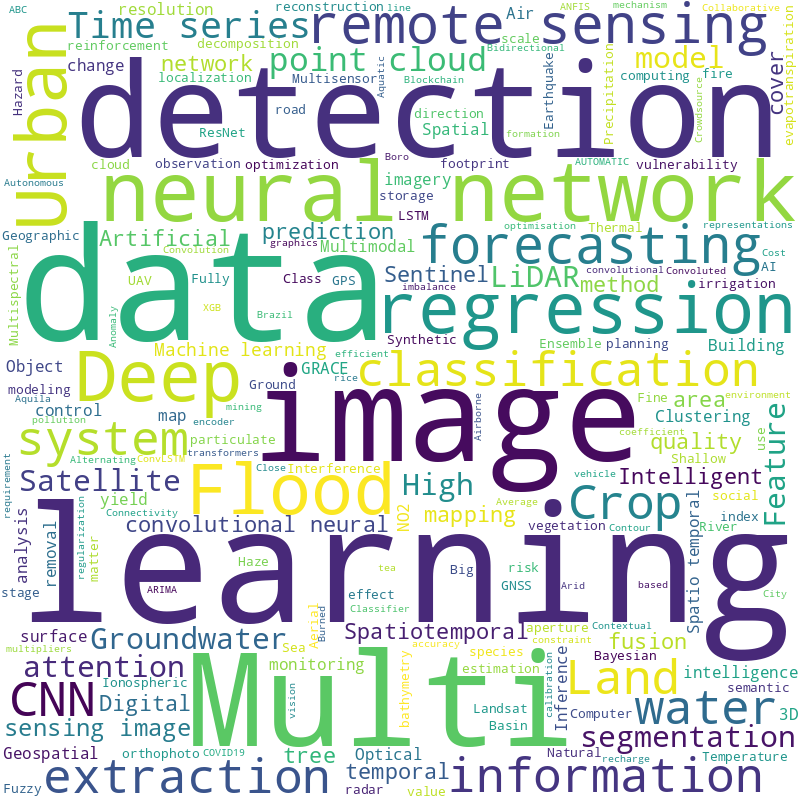}
  \captionsetup{font=scriptsize}
  \caption{Most used keywords.}
  \label{fig:Year_chart}
\end{figure}
\noindent Besides, Figure 50 shows the number of papers cited for each research area, while demonstrating, in general, an equilibrium balance in the interest accorded to each chosen application. Despite this, a relative attention is given to the fields of urban planning and water resource management. 
\begin{figure}[H]
  \centering
  \includegraphics[width=1\linewidth]{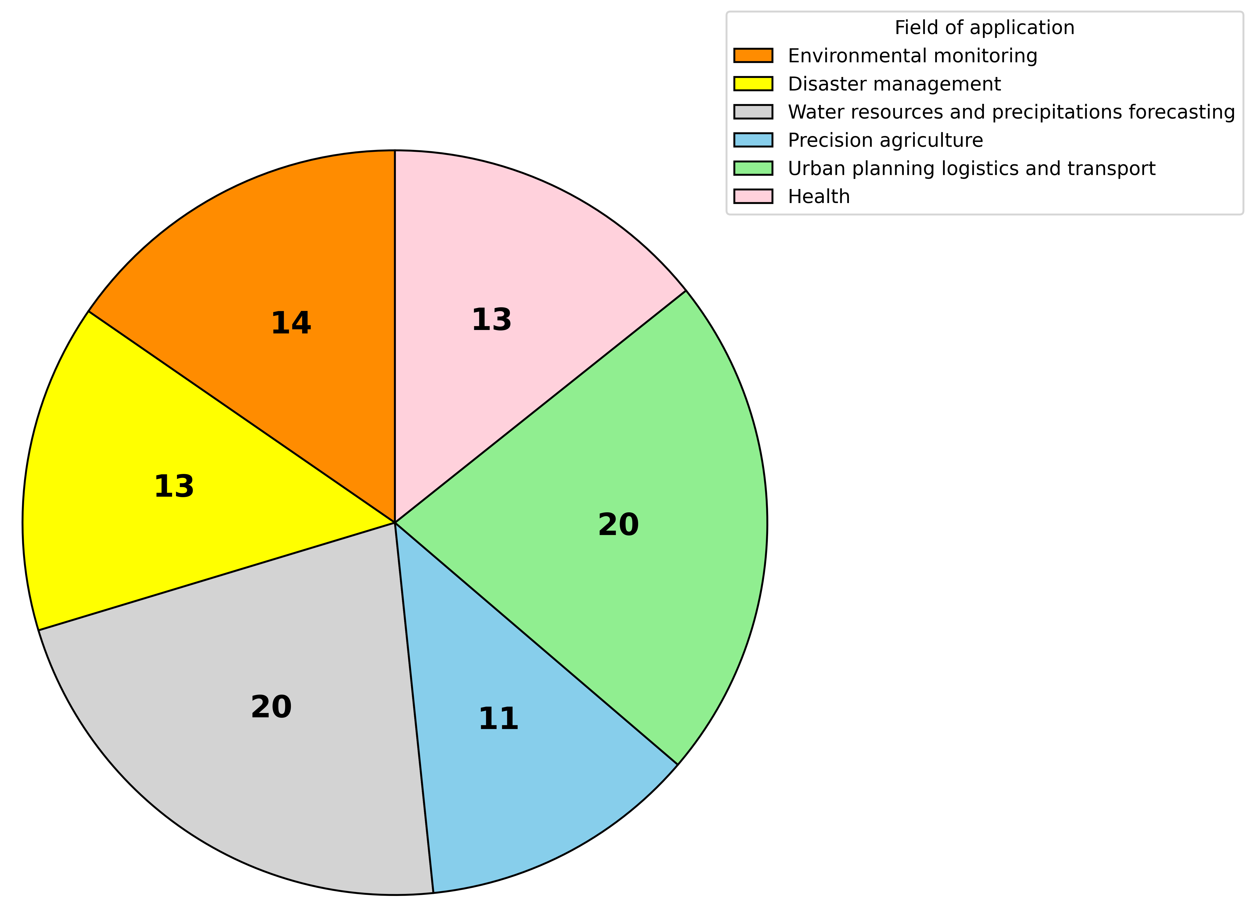}
  \captionsetup{font=scriptsize}
  \caption{Number of papers by field application.}
  \label{fig:Year_chart}
\end{figure}
\noindent Taking this analysis a step further, Figure 52 illustrates the number of papers for each chosen research sub-direction. Furthermore, the papers highlighting each GeoAI application area are balanced in terms of coverage, indicating the maturity of the axes selected. Moreover, the map in Figure 51 display the geographical distribution of all selected papers according to the university of the corresponding author, This map clearly demonstrates the global coverage of this study, citing research works from all five continents. Moreover, Chinese universities dominate this exploration, followed by the American universities.
\begin{figure*}[ht]
    \centering
    \includegraphics[width=0.86\textwidth]{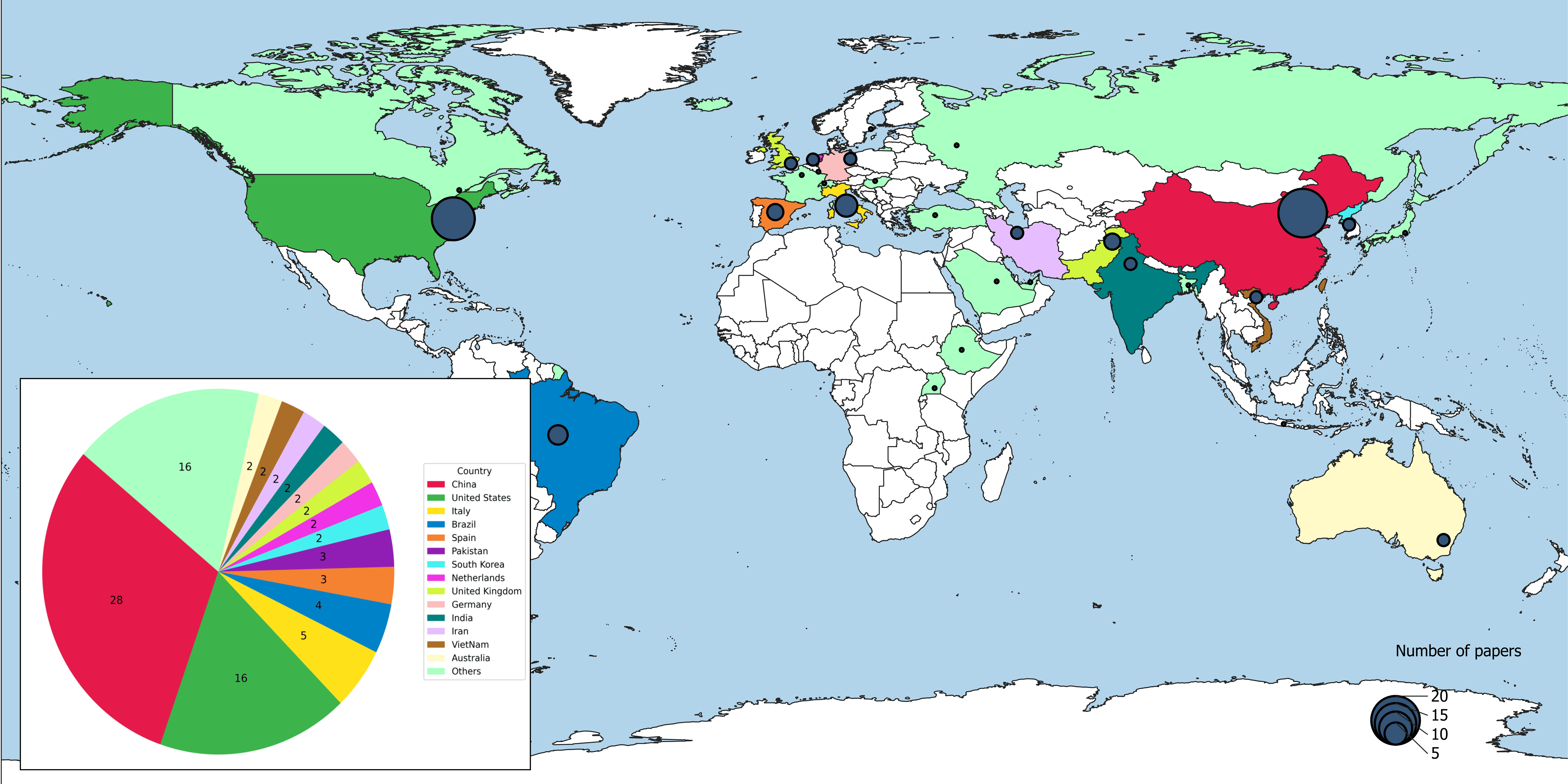}
    \captionsetup{font=scriptsize}
    \caption{Number of papers by country.}
    \label{fig:label-de-votre-figure2}
\end{figure*}
\subsection{Discussions about RQ1}
In Section 4, a range of GeoAI applications are explored, through interesting research in areas ranging from Precision agriculture to healthcare. Starting with the first area, innovative AI models are revolutionizing the agricultural sector, including models for agricultural crop mapping, yield prediction and creative methods for precision irrigation in order to optimize the agricultural productivity.
\\
A very interesting field of attack is explored next, that of urban planning, logistics and transport, covering information extraction, urban dynamics, smart cities, transportation, green infrastructure management and biodiversity. The models involved enabled automated monitoring and highly optimized urban development. Afterwards, the topic of environmental management is discussed. Then, GeoAI is widely used to assess climate change, quantify atmospheric chemistry and measure the impact of human activity on environment components, especially fauna, flora and abiotic environments.
\\
Subsequently, research linking water resource management to GeoAI is covered in this work, with a focus on hydrological modelling, groundwater monitoring and water quality. Disasters management is addressed in a set of research papers, including earthquakes, floods and forest fires.
\\
The healthcare is also explored, with a list of research projects, covering forecasting of infectious diseases using geospatial data and methods, measures to strengthen public health as well as pandemic control, highlighting COVID-19 as a major example.
\\
Altogether, this review gives a clear understanding of the contribution of GeoAI in different fields of application, guaranteeing:
\begin{itemize}[leftmargin=*]
    \item More reliable results: the complexity of used algorithms, the consistency of data and the ability to control results all mean improved accuracy and reduced operating costs.  
    \item Interoperability: data standardisation and continuous learning processes ensure the compatibility and consistency of geospatial methods.
    \item Performing tasks using new approaches that are totally different from those adopted by conventional methods, while ensuring the integration of spatial dimension into the analyses implemented.
    \item Trend forecasting: analysis of dynamics and changes to better predict events and facts.
    \item Analysis-based decision-making: GeoAI facilitates strategic decision-making based on big geodata, enabling on-the-spot interaction with alerts and emergencies.
\end{itemize}
\subsection{Fundamental aspects}
Several key aspects emerge from the discussion, notably the relevance of the methods discussed in this paper for exploiting geospatial data even with their complexity, in order to implement high-performance AI models despite, in many cases, the unavailability of input data. Furthermore, the versatility of GeoAI models makes this field broader than ever, particularly in front of the global challenges facing organizations and individuals. In spite of these challenges, the integration of high-tech AI in the service of geospatial data leads to unavoidable results. In fact, this analysis lays the foundations for a complete answer to RQ2 and 3, addressed in the next two subsections.
\begin{figure*}[ht]
    \centering
    \includegraphics[width=0.9\textwidth]{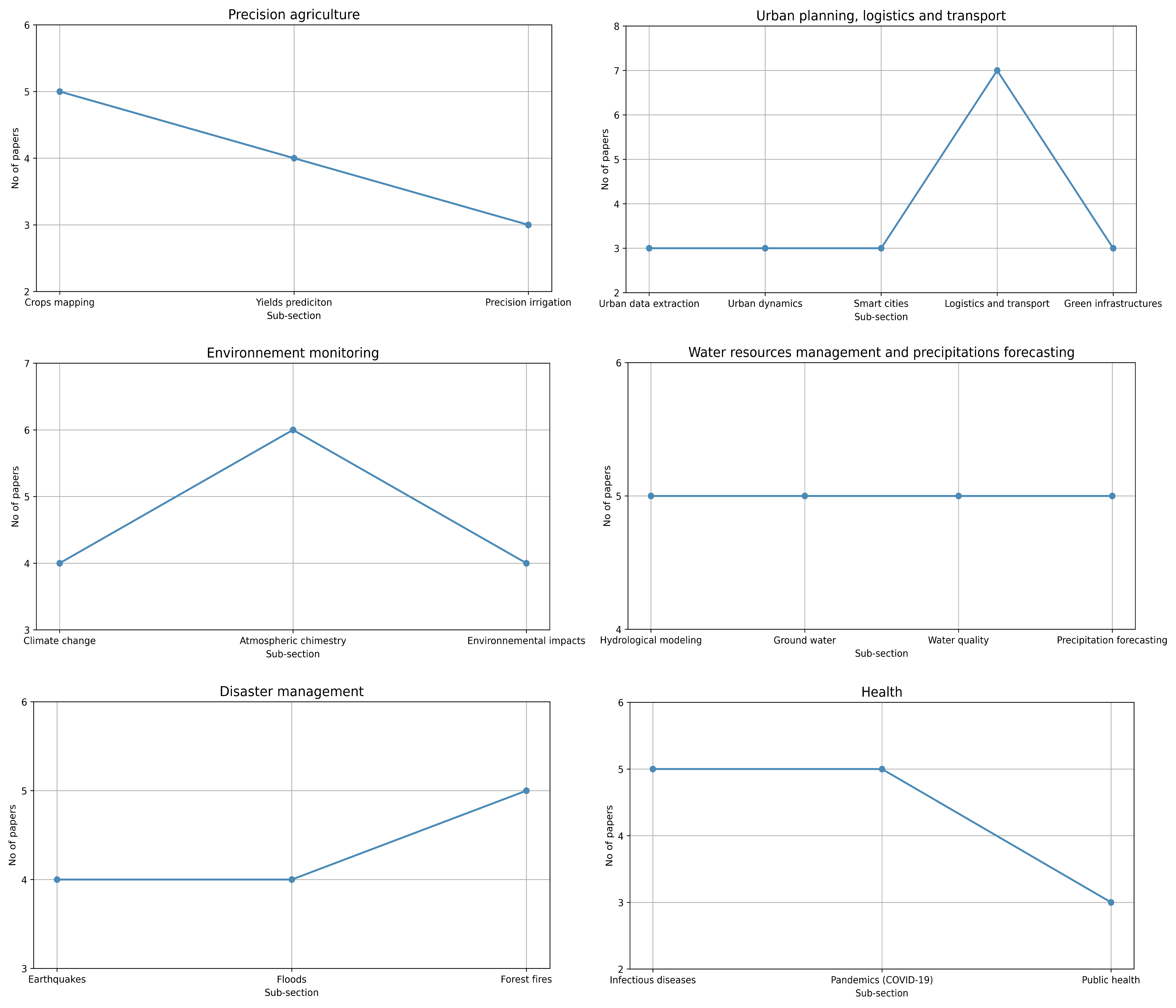}
    \captionsetup{font=scriptsize}
    \caption{Detailed statistics.}
    \label{fig:label-de-votre-figure2}
\end{figure*}
\subsection{GeoAI challenges (RQ2)}
Based on the foregoing, a number of challenges are identified. First of all, data heterogeneity poses enormous problems, especially when dealing with complex problems requiring several types of data, often unstructured. This issue being highly exposed in the case of data from multiple sources. Additionally, and given the volume and variety of geospatial data, managing and exploiting big geodata remains a significant problem, despite advances in research in this field, notably \cite{seng_spatiotemporal_2021, puttinaovarat_flood_2020, moishin_designing_2021}. 
\\
Moreover, there is an exponential need to obtain increasingly accurate models, especially when dealing with simulated or a limited volume of data. Also, data quality control remains a subject for discussion, in particular with regard to open-access data, given the need to verify its veracity and correct it where necessary. Furthermore, geospatial data sovereignty is one of the most frequent problems, as this data is more vulnerable to cyber-attacks, given its richness and implications. Besides, real-time data processing requires careful thought in terms of training, updating and feeding in newly introduced data.
\\
In contrast, numerous challenges related to the applicative aspect of GeoAI are mentioned. First of all, optimizing the training and inference times of models is a key objective, particularly in view of geospatial data complexity, requiring large amounts of memory and more optimized hardware architectures. The particular aspect of geospatiality leads to explore another confrontation, explicitly the development of specific model architectures for processing geospatial data, especially regarding elementary tasks as feature extraction and score calculation for attention mechanisms. As well, the development of models capable of handling updated geospatial data is an urgent need, particularly in relation to remote sensing data, improved each time the sattelite is emptied. This need can be met by a periodic fine-tuning, improving the accuracy of these models over time. 
\\
Security of GeoAI models against adversarial attacks requires specific adaptations so as to cope with the sensitivity and vulnerability of data. Although a few research works address this topic including \cite{xu2020assessing, qiu2021security, chen_lie_2022}, progress in terms of scientific literature is insufficiently explored so far in this context. Another challenge worth mentioning is scalability, referring to the ability of a model to handle a continuous growth in the number of tasks or data to be processed. This matter is due to the great evolution of geospatial sector in recent years, whether in terms of data accessibility and the growing number of requests. Moreover, many situations require the use of multiple GeoAI models in a single task. This configuration brings a multitude of benefits in terms of results and interoperability, but also presents many issues, often related to model validation and implementation costs, necessitating the implementation of distributed architectures \cite{jhummarwala2014parallel}, while guaranteeing modularity of deployment, secure data sharing, etc. 
\subsection{Perspectives and conclusions (RQ3)}
Faced with the challenges described above, new research horizons need to be opened up, notably the conceptualisation of new model architectures specifically adapted to the nature of geospatial data. As well as the integration of AI explainability in geospatial data processing. It is noted that little research works used this concept for geospatial tasks, such as \cite{adsuara_discovering_2020, levering_interpretable_2020, hsu2023explainable, cheng2024explainability}. Indeed, making model behavior more comprehensible and plausible with regard to the standards in question remains a promising prospect.
\\
According to van Kranenburg \cite{van2007internet}, Internet of Things (IoT) refers to an infrastructure of physical sensors networks, or virtual objects based on standard communication protocols, to exchange data between themselves, or with other external systems. Furthermore, the use of IoT devices to collect geospatial data and then employ it in AI models is an existing technique, as shown by \cite{lee2006mobeyes, siddique_sustainable_2024, puiu2016citypulse, mirdula2023mud}, but it is not sufficiently explored to take advantage of the efficiency, automation and optimization offered by IoT.
\\
In addition, the definition of standards for the training and inference of GeoAI models, similar to those governing the access and use of geospatial data, such as Spatio-Temporal Asset Catalog (STAC) \cite{STAC2023}, Geographically Encoded Objects for Really Simple Syndication (GeoRSS) \cite{GeoRSS2023}, and the standards of Open Geospatial Consortium (OGC) \cite{OGC2023}, constitutes an interesting outlook. The aim of defining such standards is to ensure the quality of the methods implemented and facilitate the integration and accessibility of models. Moreover, interpreting geospatial data, particularly remotely sensed images, is an affordable target area by coupling computer vision and Natural Language Processing (NLP) models \cite{parsing2009speech}. The aim of this combination is to replace the enormous efforts of specialists while completing the required tasks in a relatively short time. The work of \cite{tartini2023multi, guo_remote_2024} combining NLP models, more precisely Large Language Model (LLM) with CNNs, represent a good start in spite of the limitation of results. 
\\
In closing, it is worth mentioning that GeoAI realizes significant progress, not only in terms of the evolution of methods and algorithms, but also in terms of the broad scope of coverage of the themes in question. The diversity of GeoAI's applications demonstrates the breadth of its advances, a breadth extending to other areas of geospatial data analysis and processing. For example, a wealth of research cited demonstrates the importance of AI-GIS synergy, notably \cite{zhang_covid-19_2023, zafar_traffic_2022}.
\\
As far as the application side of GeoAI is concerned, an attractive societal and environmental impact is emerging, underlining the potential offered by GeoAI to help solve problems of different dimensions. In this respect, GeoAI provides solutions insofar as novel models are proposed to optimize the use of natural resources and agricultural land, ensure state-of-the-art urban planning with certainty. Also, GeoAI solemnly undertakes to respect social integrity in terms of availability of commercial services, protection against natural disasters and healthcare, etc.
\\
A crucial point to raise is the need to guarantee a professional framework for collaboration between academics and professionals. This collaboration manifests itself into a mutual and effective transfer of knowledge, and an enhancement of GeoAI's learning methods through the inclusion of a practical dimension. This interdisciplinary approach irrevocably certifies irreproachable results by offering unexpected solutions to the most intractable problems.
\\
Hence, it is concluded that GeoAI has a promising future. Given the exponential progress of AI, it is difficult to predict the next degree of evolution related to this theme. Furthermore, this work will certainly help researchers to better understand the current state of GeoAI, as well as the problems it poses, in order to continue making progress in such a hopeful subject.
\noindent
\uline{\textbf{Declaration of Competing Interest}}
\\
\\
The writing team of this paper mentions that no financial or lucrative interest is behind this work.
\\
\\
\uline{\textbf{Aknowledgments}}
\\
\\
The authors would like to thank the Artificial Intelligence Geo-decision Networking Optimisation And Cybersecurity (AGNOX) research team at the National Institute of Posts and Telecommunications (INPT) - Morocco for their unconditional support.
\printbibliography
\end{multicols}
\end{document}